\pdfoutput=1

\documentclass{article}
\usepackage{iclr2027_conference,times}

\usepackage[T1]{fontenc}
\usepackage[utf8]{inputenc}
\usepackage{microtype}
\usepackage{inconsolata}

\usepackage{amsmath}
\usepackage{amssymb}
\usepackage{graphicx}
\usepackage{algorithm}
\usepackage{algpseudocode}
\usepackage{multirow}
\usepackage[table]{xcolor}
\usepackage{colortbl}
\usepackage{tcolorbox}
\usepackage{float}
\usepackage{enumitem}
\usepackage{placeins}
\usepackage{booktabs}
\extrafloats{100}
\usepackage{tabularx}
\usepackage{tikz}
\usetikzlibrary{arrows.meta, positioning}
\tcbuselibrary{breakable,skins}
\usepackage{caption}
\usepackage{subcaption}
\usepackage{rotating}
\usepackage{needspace}
\usepackage{wrapfig}
\usepackage{hyperref}
\usepackage{url}

\setlist{nosep, leftmargin=*}

\title{MedPRMBench: A Fine-grained Benchmark for Process Reward Models in Medical Reasoning}

\author{Xiang Zheng$^{*}$\\
  {\normalfont\small wangyou.zx@alibaba-inc.com}\\
  {\normalfont\small Alibaba Group}\\
  {\normalfont\small China}
\And Lingyan Wu$^{*}$\\
  {\normalfont\small wulingyan.wly@alibaba-inc.com}\\
  {\normalfont\small lingyanwu666@foxmail.com}\\
  {\normalfont\small Alibaba Group}\\
  {\normalfont\small China}
\And Weiqi Zhai$^{*}$\\
  {\normalfont\small zhaiweiqi.zwq@alibaba-inc.com}\\
  {\normalfont\small Alibaba Group}\\
  {\normalfont\small China}
\AND Wei Wang\\
  {\normalfont\small weixiong.ww@alibaba-inc.com}\\
  {\normalfont\small Alibaba Group}\\
  {\normalfont\small China}
\And Xuan Ren\\
  {\normalfont\small renxuan.ren@alibaba-inc.com}\\
  {\normalfont\small Alibaba Group}\\
  {\normalfont\small China}
\And Zifan Zhang\\
  {\normalfont\small zhangzifan.zzf@alibaba-inc.com}\\
  {\normalfont\small Alibaba Group}\\
  {\normalfont\small China}
\AND \makebox[\dimexpr\textwidth-2\tabcolsep\relax][c]{%
  \begin{tabular}[t]{l}
  Bing Zhao$^{\dagger}$\\
  {\normalfont\small xiongdao@alibaba-inc.com}\\
  {\normalfont\small Alibaba Group}\\
  {\normalfont\small China}
  \end{tabular}}}

\iclrfinalcopy

\begin{document}
\begingroup
\renewcommand{\thefootnote}{\fnsymbol{footnote}}
\maketitle
\footnotetext[1]{These authors contributed equally to this work.}
\footnotetext[2]{Corresponding author.}
\endgroup
\begin{abstract}
Medical reasoning can reach a plausible answer while overlooking an unsafe or unsupported intermediate step. Process-Level Reward Models (PRMs) can verify such steps, yet existing PRM benchmarks largely target mathematics rather than patient-specific clinical errors. We introduce \textbf{MedPRMBench}, a step-level medical reasoning benchmark with 14 clinical error types, including five new and two clinically re-scoped types, and rule-defined severity grades. Built from seven medical QA sources via structured error construction and review by 86 physicians from Grade-A tertiary hospitals, it provides 6{,}500 evaluation questions, 13{,}000 reasoning chains, and 113{,}910 step labels, plus a leakage-clean 6{,}870-question training split. The best of 20 prompted critics reaches only 75.4\% PRMScore, and critics differ sharply in error-type profiles and operating points. A Qwen3-8B PRM trained on the split reaches 86.6\% PRMScore and 89.0\% strict first-error localization; in matched ablations, physician review raises error recall by 6.55 points and cuts clean-chain false alarms from 5.6\% to 2.1\%. The training signal replicates across model families; without further training, the PRM transfers to ProcessBench and PRMBench, exceeding six of nine math-trained PRMs on each.
\par\smallskip\noindent\href{https://huggingface.co/datasets/skylenage-ai/MedPRMBench-Test}{\raisebox{-0.3\height}{\includegraphics[height=1.1em]{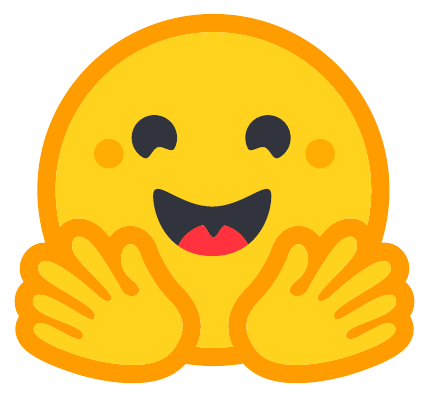}}}\,\textbf{Public test set:}
\par\noindent\url{https://huggingface.co/datasets/skylenage-ai/MedPRMBench-Test}
\end{abstract}


\section{Introduction}
\label{sec:introduction}

Large language models (LLMs) have made progress on complex reasoning~\citep{jaech2024openai,guo2025deepseek,yang2025qwen3}, including medical question answering. A correct final answer, however, does not establish that each intermediate inference is sound: a reasoning chain may overlook an allergy, misapply evidence to a patient, or omit a necessary differential. Medical reasoning is safety-sensitive, knowledge-intensive, and often made under incomplete information, so such failures can matter even when the answer is right. They motivate checking the reasoning process, not only its outcome. Process-Level Reward Models (PRMs) provide one way to do so by scoring individual steps; their scores can supervise model training~\citep{lightman2023lets,ouyang2022training} or rank candidate reasoning trajectories during test-time search~\citep{wang2023mathshepherd,snell2024scaling}.

How, then, should we evaluate a PRM used for medical reasoning? ProcessBench~\citep{zheng2024processbench} tests first-error localization in mathematical solutions, while PRMBench~\citep{song2025prmbench} evaluates fine-grained errors using a taxonomy developed for mathematics. Both provide useful process-evaluation protocols, but neither is designed to test whether a verifier notices a missed safety check, patient-context mismatch, or incomplete differential diagnosis, nor how clinically consequential an error is. Medical PRM and reasoning methods such as Med-PRM~\citep{yun2025medprm}, MedCEG~\citep{mu2025medceg}, and MedS$^3$~\citep{jiang2025meds3} further motivate this evaluation need: developing a reward signal is distinct from systematically measuring which clinical reasoning failures it detects.

\begin{figure}[!t]
\centering
\includegraphics[width=0.98\columnwidth]{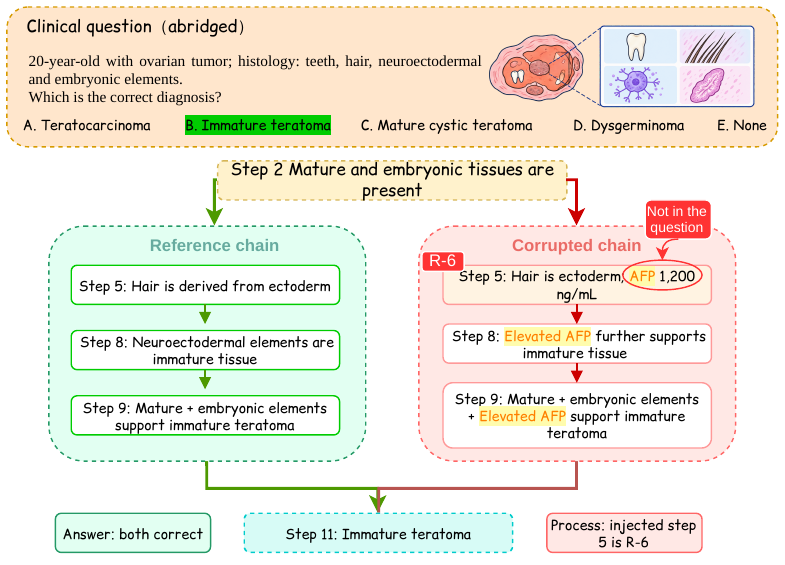}
\caption{A correct final answer can coexist with unsupported clinical evidence. In this MedPRMBench test example, the corrupted chain invents an AFP result at Step~5 (gold R-6 label), carries it through Steps~8--9, and still selects the correct diagnosis, immature teratoma.}
\label{fig:motivation}
\end{figure}

We introduce \textbf{MedPRMBench} to fill this evaluation gap (Figure~\ref{fig:motivation}). Extending PRMBench's three-category scaffold to clinical reasoning, we define 14 error types and record severity alongside step labels. The benchmark draws on seven medical QA sources whose reasoning is heterogeneous: some provide expert-authored step-by-step explanations (Class~A), while others provide only answers and require generated traces (Class~B). To construct comparable, controlled errors across both, we distill each trace into a Clinical Reasoning Blueprint (CRB) that exposes evidence, prerequisites, and safety-critical steps, then combine blueprint-guided candidate construction with deterministic verification and physician review. The release comprises 6{,}500 evaluation questions, 13{,}000 paired correct and erroneous chains, and 113{,}910 step-level labels, plus 6{,}870 training questions. We evaluate prompted critics and medical PRMs, train a benchmark-specific Qwen3-8B PRM following the Med-PRM objective~\citep{yun2025medprm}, and test its no-further-training transfer to ProcessBench and PRMBench. The best prompted critic, GPT-5.4, reaches 75.4\% PRMScore, while the two open-source critic groups average 53.8--54.8\%; Non-Redundancy and Deception Resistance are the hardest slices for frontier API critics. The trained PRM reaches 86.6\%, yet still misses nearly half of prerequisite errors and about a third of safety-awareness errors. A full annotated example appears in Appendix~\ref{app:error_types}.

Our main contributions are as follows:
\begin{itemize}
    \item We introduce \textbf{MedPRMBench}, a step-level medical process-verification benchmark with 6{,}500 evaluation questions, 113{,}910 step labels, and 14 clinical error types (five new and two clinically re-scoped relative to PRMBench) with severity metadata.
    \item We propose a \textbf{blueprint-guided construction pipeline} that unifies expert-authored and generated reasoning through CRBs and combines controlled candidate construction, deterministic checks, and physician validation, together with a 6{,}870-question training split free of test overlap.
    \item We \textbf{evaluate medical error detection} across PRMs and prompted critics, with three main findings: (i) models with similar PRMScores can sit at very different operating points, so error recall and clean-chain false alarms must accompany the aggregate; (ii) after in-distribution training, missed prerequisites and safety checks remain the weakest error types, but training rows carrying them raise their recall by 7.4 and 20.2 points; (iii) transfer is asymmetric: without further training, the medical PRM stays competitive with math-trained PRMs on ProcessBench and PRMBench, whereas math-trained PRMs miss most clinical errors.
\end{itemize}


\section{Related Work}
\label{sec:related_work}

\noindent\textbf{Medical process reward modeling.}
Recent medical PRM methods improve medical reasoning or reward estimation through retrieval-augmented verification, graph-based clinical evidence alignment, and search-based supervision~\citep{yun2025medprm,mu2025medceg,jiang2025meds3}. These efforts provide useful training paradigms, but do not provide a dedicated evaluation resource for fine-grained, step-level medical reasoning errors.

\smallskip
\noindent\textbf{Process-evaluation benchmarks.}
ProcessBench~\citep{zheng2024processbench} and PRMBench~\citep{song2025prmbench} establish step-level evaluation protocols for process reward models. Their tasks and taxonomies, however, are principally grounded in mathematical reasoning and do not represent medical failures such as missed safety checks, insufficient differential diagnosis, contextual inapplicability, or skipped prerequisites.

\smallskip
\noindent\textbf{Positioning of MedPRMBench.}
MedPRMBench complements these lines of work as an evaluation benchmark for medical reasoning verification. It combines a 14-type clinical error taxonomy, severity metadata, and documented construction and review procedures, enabling diagnostic evaluation of medical PRMs and prompted critics. Appendix~\ref{app:detailed_related_work} provides a detailed comparison with prior work.

\section{MedPRMBench Construction}

\label{sec:method_overview}

\begin{figure*}[t]
    \centering
    \includegraphics[width=0.88\textwidth]{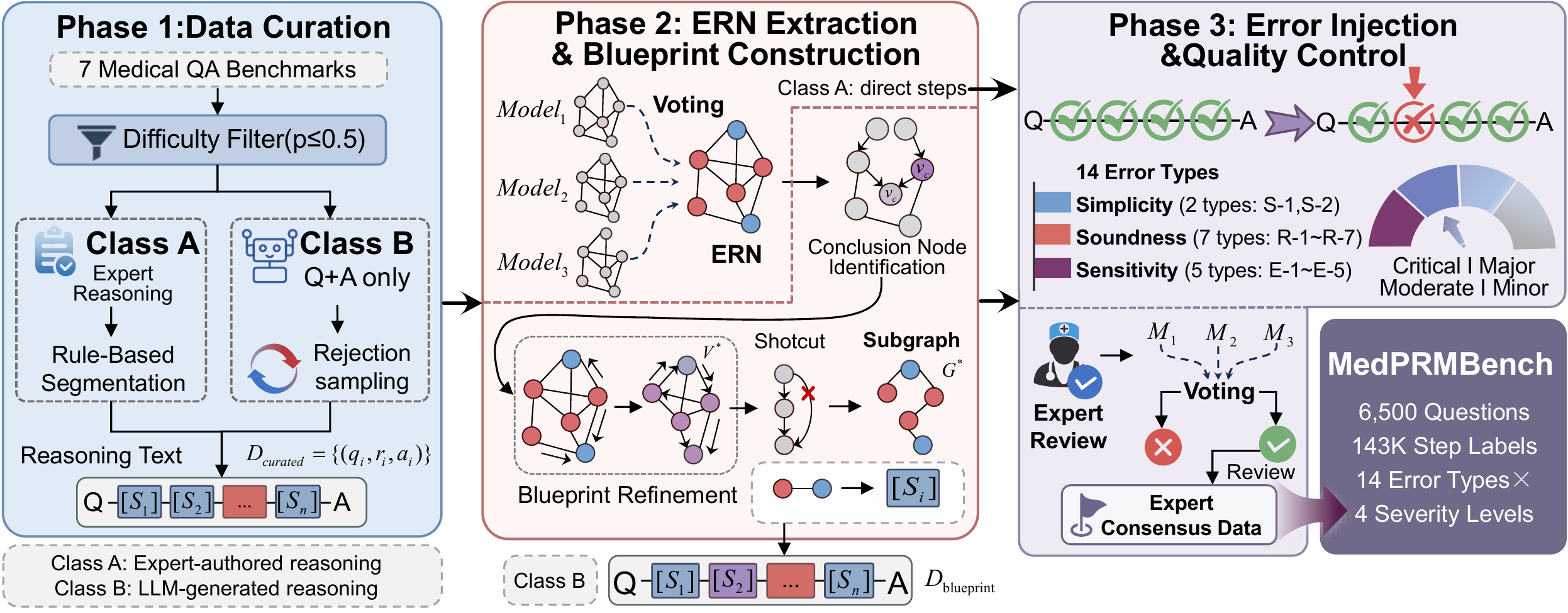}
    \caption{Overview of the MedPRMBench construction pipeline: Phase~1 aggregates seven medical QA sources, filters by difficulty, and generates reasoning traces by rejection sampling; Phase~2 extracts ERNs, distills CRBs, and adds safety-critical, criticality, and prerequisite annotations; Phase~3 constructs controlled candidate errors, followed by quality control and expert revision.}
    \label{fig:framework}
\end{figure*}

This section presents the construction pipeline of MedPRMBench. As illustrated in Figure~\ref{fig:framework}, the construction process comprises three phases: \textbf{Phase~1: Data Curation} (\S\ref{sec:data_curation}), \textbf{Phase~2: Clinical Reasoning Blueprints} (\S\ref{sec:blueprint_construction}), and \textbf{Phase~3: Error Taxonomy and Candidate Construction} (\S\ref{sec:error_injection}). Appendix~\ref{app:error_types} (Figure~\ref{fig:medprmbench_example}) illustrates a complete corrupted chain with step-level clinical error annotations.

\subsection{Data Curation}
\label{sec:data_curation}
We aggregate seven medical QA benchmarks---MedCaseReasoning~\citep{wu2025medcasereasoning}, MedQA~\citep{jin2021disease}, MedMCQA~\citep{pal2022medmcqa}, the medical subset of MMLU~\citep{hendrycks2021measuring}, PubMedQA~\citep{jin2019pubmedqa}, MedXpertQA~\citep{zuo2025medxpertqa}, and MedExpQA~\citep{medexpqa2024}---and apply difficulty-aware filtering. Sources with expert-authored reasoning (Class~A, e.g., MedCaseReasoning) are segmented directly; sources without reasoning text (Class~B) receive rejection-sampled reasoning chains verified against the gold answer. This yields the curated dataset $\mathcal{D}_{\text{curated}} = \{(q_i, r_i, a_i)\}_{i=1}^{N}$ of questions, reasoning traces, and answers for blueprint construction. Appendix~\ref{app:data_curation_details} gives the curation details.

\subsection{Clinical Reasoning Blueprints}
\label{sec:blueprint_construction}

Free-text reasoning chains obscure which evidence supports which conclusions and which steps are prerequisites, making controlled modification difficult. We therefore construct \emph{Clinical Reasoning Blueprints} (CRBs): directed knowledge subgraphs distilled from reasoning text that trace a core path from clinical evidence to the final conclusion~\citep{mu2025medceg}. We attach safety, prerequisite, and structural annotations so that the construction pipeline can select applicable operations and record their targets. This phase segments reasoning, extracts Evidence Reasoning Networks, distills and verifies CRBs, adds the annotations, and linearizes the resulting structures. Appendix~\ref{app:blueprint_example} gives an end-to-end example.

As described in Section~\ref{sec:data_curation}, datasets are categorized into Class~A and Class~B based on reasoning text provenance. The two classes follow different processing paths in this phase but converge to a unified output format.

\paragraph{Reasoning step segmentation.}
We segment each reasoning trace into independent steps $\mathbf{s} = (s_1, s_2, \ldots, s_L)$. Class~A traces use their explicit numbered format; Class~B traces are segmented during ERN extraction and blueprint linearization.

\paragraph{Evidence reasoning network extraction.}
For Class~B datasets, we extract an \emph{Evidence Reasoning Network} (ERN), a directed graph $G_{\text{ERN}} = (V, E)$ whose edges encode clinical relations among entities in the reasoning trace. A three-model semantic-voting procedure retains triplets supported by at least two models and encourages connected evidence paths. Class~A datasets use their official reasoning steps directly. The resulting ERN provides the structured representation used for blueprint distillation; model identities, matching rules, and fixed thresholds are given in Appendix~\ref{app:semantic_voting}.

\paragraph{Blueprint distillation, annotation, and linearization.}
Blueprint distillation removes redundant information from $G_{\text{ERN}}$ to obtain the CRB $G^* = (V^*, E^*)$. We identify the conclusion node, bridge semantically equivalent nodes when needed, traverse the connected evidence graph in both directions, and apply transitive reduction~\citep{aho1972transitive} to remove shortcut edges. An LLM-based sufficiency check adds connected supporting triplets when the resulting blueprint lacks enough information for the gold answer. Class~A traces bypass ERN distillation but undergo the same sufficiency check. Formal definitions and the construction algorithm appear in Appendices~\ref{app:blueprint_details} and \ref{app:crb_algorithm}.

Finally, we add safety-critical, node-criticality, and prerequisite annotations and linearize each blueprint into a natural-language step sequence. Both source classes then share the blueprint dataset $\mathcal{D}_{\text{blueprint}}$, which supplies the reasoning sequence, CRB, and annotations for Phase~3 error construction.

\subsection{Error Taxonomy and Candidate Construction}
\label{sec:error_injection}

Building on the blueprint dataset from Section~\ref{sec:blueprint_construction}, this phase constructs candidate erroneous reasoning chains from the correct chains and then filters them through verification and expert review. Rather than treating errors as arbitrary text perturbations, the construction is guided by a 14-type error taxonomy spanning three categories---\emph{Simplicity}, \emph{Soundness}, and \emph{Sensitivity}---and proceeds through three core stages: error type selection and target localization, candidate error construction with severity assessment, and composite error synthesis with deterministic verification.

\paragraph{Error taxonomy.}
\label{sec:error_selection}
We adapt PRMBench's nine-type taxonomy to clinical reasoning and define 14 error types organized into the same three high-level categories. Table~\ref{tab:error_taxonomy} makes the provenance explicit: seven types retain the earlier labels as an organizing scaffold, two are clinically re-scoped, and five have no PRMBench counterpart. Each type has an inherent severity weight $w_{\text{type}}(c) \in [0, 1]$ (reflecting its potential clinical impact) and a corresponding construction operation; Appendix~\ref{app:error_types} gives detailed definitions and examples.

\begin{table*}[t]
\centering
\caption{Taxonomy provenance and construction operations. MedPRMBench retains seven labels as clinical instances, re-scopes two types, and adds five types without PRMBench counterparts. ``Re-scoped'' denotes a clinically specialized definition and operation.}
\label{tab:error_taxonomy}
\scriptsize
\resizebox{0.98\textwidth}{!}{%
\begin{tabular}{lllll}
\toprule
\textbf{Category} & \textbf{Code} & \textbf{MedPRMBench type} & \textbf{PRMBench provenance} & \textbf{Construction operation} \\
\midrule
\multirow{2}{*}{Simplicity} & S-1 & Non-Redundancy (NR) & Same label; clinically instantiated & Insert redundant step \\
 & S-2 & Non-Circular Logic (NCL) & Same label; clinically instantiated & Inject circular argument \\
\midrule
\multirow{7}{*}{Soundness} & R-1 & Evidence-Based Soundness (EBS) & Re-scoped from Empirical Soundness & Replace medical fact \\
 & R-2 & Step Consistency (SC) & Same label; clinically instantiated & Introduce contradiction \\
 & R-3 & Contextual Applicability (CA) & Re-scoped from Domain Consistency & Ignore patient context \\
 & R-4 & Confidence Invariance (CI) & Same label; clinically instantiated & Insert overconfident claim \\
 & R-5 & Safety Awareness (SA) & New: no PRMBench counterpart & Remove safety check \\
 & R-6 & Information Grounding Compliance (IGC) & New: no PRMBench counterpart & Fabricate entity \\
 & R-7 & Trajectory Reasoning (TR) & New: no PRMBench counterpart & Reverse causal/temporal order \\
\midrule
\multirow{5}{*}{Sensitivity} & E-1 & Prerequisite Sensitivity (PS) & Same label; clinically instantiated & Delete prerequisite step \\
 & E-2 & Deception Resistance (DR) & Same label; clinically instantiated & Insert distractor \\
 & E-3 & Multi-Solution Consistency (MSC) & Same label; clinically instantiated & Dismiss alternatives \\
 & E-4 & Quantitative Correctness (QC) & New: no PRMBench counterpart & Alter numerical value \\
 & E-5 & Differential Diagnosis Coverage (DDC) & New: no PRMBench counterpart & Narrow differential \\
\bottomrule
\end{tabular}%
}
\end{table*}

Not all error types apply to every question. We use the question, reasoning chain, and blueprint annotations to tag applicability, then select one to three target types per instance by distribution-aware sampling. Appendix~\ref{app:applicability_sampling} gives the sampling and target-step rules.

\paragraph{Candidate error construction.}
\label{sec:injection_execution}

LLMs apply type-specific few-shot prompts to the original chain $\mathbf{s}$, a target error type $c$, and target step $k$ to produce candidate corrupted chains $\tilde{\mathbf{s}}$ with modified-step metadata. We use round-robin generation across three models for diversity, followed by deterministic verification and physician review.

For each instance, the operation is executed independently for each selected error type, producing up to three \emph{single-error variants}, each consisting of a corrupted step sequence $\tilde{\mathbf{s}}_i^{(c)}$, its error type $c$, and its verified error-step index set $\mathbf{e}_i^{(c)} \subseteq \{1, \ldots, L\}$.

Each corrupted variant receives a continuous severity score combining disrupted-step fraction, target-step safety criticality, and error-type weight. A separate rule assigns the four-level label (Critical/Major/Moderate/Minor) from disruption and target-step criticality. These reproducible construction grades place 6{,}326/6{,}500 evaluation variants (97.3\%) in Major or Critical; Appendix~\ref{app:severity_assessment} specifies both rules.

\paragraph{Composite errors and verification.}
\label{sec:composite_and_verification}

To represent interacting failures, we synthesize composite variants with two or three single-error types chosen for cross-category, step, and severity diversity. Deterministic text-diff verification reconciles reported error indices with actual textual changes before physician review. After verification and quality control, each evaluation question retains one corrupted chain paired with its original correct chain; multi-error chains receive multi-label step annotations. Appendix~\ref{app:composite_synthesis} gives the synthesis and verification details.

\subsection{Quality Control and Statistics}
\label{sec:quality_control}

\begin{wraptable}{r}{0.46\textwidth}
\vspace{-\intextsep}
\centering
\caption{Leakage-clean MedPRMBench split and chain provenance (Class~A expert-authored, Class~B generated; identical construction and physician-validation workflow).}
\label{tab:dataset_split_main}
\scriptsize
\setlength{\tabcolsep}{4pt}
\begin{tabular}{lrr}
\toprule
 & \textbf{Training} & \textbf{Evaluation} \\
\midrule
Questions & 6{,}870 & 6{,}500 \\
Correct chains & 6{,}870 & 6{,}500 \\
Corrupted chains & 6{,}870 & 6{,}500 \\
Class A / B & 3{,}244 / 3{,}626 & 3{,}711 / 2{,}789 \\
\bottomrule
\end{tabular}
\end{wraptable}

After automatic filtering, 14{,}982 candidate instances receive validation by 86 physicians from Grade-A tertiary hospitals in China: each is annotated and then checked by a different physician in 98.6\% of review records. A blinded subset of 1{,}000 Class~A candidates receives an additional independent annotation. Across its 5{,}763 corrupted-chain steps, the physicians' proposed binary error-step labels achieve 84.7\% observed agreement and Cohen's $\kappa=0.69$. Physician corrections are adopted only if a $\geq$2/3 three-model consistency vote supports them; we exclude the 1{,}600 candidates whose corrections fail this vote, one out-of-scope (veterinary) case, and two cases left without an error code, yielding 13{,}379 questions before overlap removal. Appendix~\ref{app:annotation_reliability} gives the full agreement protocol; the public artifact retains initial annotations, second-stage outcomes, model votes, and filtering reasons.

The resulting MedPRMBench contains 13{,}370 questions with diverse error types and clinical contexts. Table~\ref{tab:medprmbench_stats} summarizes error-type occurrence statistics; per-type counts are not mutually exclusive because a corrupted chain may contain multiple injected errors. Appendices~\ref{app:quality_control}, \ref{app:applicability_sampling}, and \ref{app:dataset_splitting} provide the detailed quality-control, sampling, and split protocols.

We preserve source-dataset partitions when forming the release
(Table~\ref{tab:dataset_split_main}): MedQA-USMLE and
MedMCQA retain their original training splits as protected datasets, and a
normalized-question audit removes nine train--test overlaps, yielding the
6{,}870-question training release. The complete split procedure
and integrity audit are provided in Appendix~\ref{app:dataset_splitting}.

Across all 13{,}370 training and evaluation questions, corrupted chains
contain 9.6 steps on average, including 3.3 erroneous steps, with the first
error at step 3.2 on average. Questions average 917 characters, compared with
153 in PRMBench, reflecting the richer clinical context required for medical
process verification.

\section{Evaluation}
\subsection{Setup}

\noindent\textbf{Task.}
Given a medical question $Q$ and a step-by-step reasoning chain
$S = \{s_1, \ldots, s_n\}$, models must assess whether each step $s_i$ is
correct; a step is erroneous if it contains factual medical mistakes, flawed
clinical logic, missing prerequisite reasoning, or contradictions with the
clinical vignette. Every step in the chain is evaluated, reflecting that
clinical reasoning errors can propagate and co-occur. Our primary measure is
balanced PRMScore; we additionally report erroneous-step detection and
first-error detection to characterize how models identify clinically important
failures.

\medskip
\noindent\textbf{Metrics and Evaluation Protocol.}
Models label each reasoning step ``+'' (medically correct and logically valid)
or ``-'' (erroneous). PRMs with accessible token logits use the normalized
step-boundary probability $P(+)$ as the correctness score, with a fixed 0.5
decision threshold. Open and frontier API critic models produce the same labels
generatively. We report these models as prompted critics and PRMs separately:
the shared task enables a broad capability comparison, while the two interfaces
provide different forms of confidence. Prompts, thresholds, and checkpoints are
fixed without test-set selection. Appendix~\ref{app:prompt_eval} gives the
prompts, and Appendix~\ref{app:api_protocol} reports repeated-sampling and
continuous-score diagnostics where available.

\medskip
\noindent\textbf{Models.}
Our benchmark-specific PRM fully fine-tunes Qwen3-8B on the 6{,}870-question
training split (13{,}740 correct and corrupted chains). Cross-entropy is applied
only to the ``+''/``-'' logits at step boundaries; we use one pre-specified epoch,
the epoch-end checkpoint, and no test-set checkpoint or threshold selection.
Appendix~\ref{app:prm_details} gives the training configuration. We also evaluate
the released Med-PRM-Reward v1.0, open-source medical and reasoning critics, and
frontier API critics (proprietary and open-weight); Appendix~\ref{app:model_details} lists their versions
and inference settings.


\subsection{Main Results}
\label{sec:main_results}

\noindent\textbf{Evaluation Metrics.}
Following PRMBench~\citep{song2025prmbench}, we adopt \textbf{PRMScore} as the primary evaluation metric, defined as:
\begin{equation}
\mathrm{PRMScore} = \tfrac{1}{2}F1_{\mathrm{err}} + \tfrac{1}{2}F1_{\mathrm{cor}}.
\end{equation}
Here, $F1_{\mathrm{err}}$ treats erroneous steps as the positive class and
$F1_{\mathrm{cor}}$ treats correct steps as the positive class. This equal
weighting rewards error detection while penalizing false alarms on valid clinical
reasoning. We additionally report \textbf{Acc}, the erroneous-step detection
rate (error recall), and \textbf{First}, the rate at which the gold first-error step is
identified. PRMBench uses the same definition.

\begin{table*}[!t]
\centering
\caption{PRMScore (\%) on MedPRMBench. Taxonomy columns are overlapping chain slices (code-attributed recall: Appendix~\ref{app:attributed_diagnostics}). \textbf{Bold}/\underline{underline}: best/second-best zero-shot or released model; shaded row: in-distribution reference. Six open-model rows are full-coverage reruns (Appendix~\ref{app:parse_coverage}).}
\label{tab:error_detection_by_type_en}
\centering
\resizebox{\textwidth}{!}{
\large
\setlength{\tabcolsep}{4.5pt}
\begin{tabular}{l ccc cc ccccccc ccccc}
\toprule
\multirow{2}{*}{\textbf{Model}} & \multirow{2}{*}{\textbf{PRMScore}} & \multirow{2}{*}{\textbf{Acc}} & \multirow{2}{*}{\textbf{First}} & \multicolumn{2}{c|}{\textbf{Simplicity}} & \multicolumn{7}{c|}{\textbf{Soundness}} & \multicolumn{5}{c}{\textbf{Sensitivity}} \\
\cline{5-6} \cline{7-13} \cline{14-18}
 & & & & \textbf{NR} & \textbf{NCL} & \textbf{EBS} & \textbf{SC} & \textbf{CA} & \textbf{CI} & \textbf{SA} & \textbf{IGC} & \textbf{TR} & \textbf{PS} & \textbf{DR} & \textbf{MSC} & \textbf{QC} & \textbf{DDC} \\
\midrule
\rowcolor{gray!15} \multicolumn{18}{c}{\textbf{\emph{Open-source Medical Models, Prompted as Critic Models}}} \\
txgemma-9b-chat & 57.7 & 45.5 & 47.0 & 55.6 & 57.9 & 57.2 & 58.0 & 57.8 & 58.6 & 58.7 & 57.1 & 59.0 & 58.3 & 55.6 & 59.2 & 58.4 & 57.9 \\
Llama-3.1-8B-UltraMedical & 56.6 & 43.2 & 28.7 & 58.1 & 55.5 & 56.1 & 56.2 & 55.7 & 56.5 & 57.2 & 56.1 & 57.1 & 56.5 & 55.3 & 57.6 & 57.0 & 56.3 \\
llama-3-meerkat-8b & 55.1 & 38.1 & 23.0 & 55.9 & 53.9 & 54.8 & 55.5 & 55.2 & 54.4 & 55.5 & 55.0 & 55.4 & 55.4 & 54.0 & 55.0 & 56.2 & 54.1 \\
HuatuoGPT-o1-8B & 57.9 & 39.9 & 26.0 & 57.7 & 57.0 & 57.8 & 58.0 & 58.3 & 57.9 & 58.4 & 57.1 & 58.8 & 58.6 & 56.5 & 59.5 & 59.1 & 57.7 \\
Meditron3-8B & 46.8 & 9.7 & 4.4 & 43.6 & 47.0 & 45.9 & 46.2 & 48.4 & 50.7 & 46.4 & 45.9 & 47.9 & 48.6 & 45.3 & 48.8 & 48.7 & 45.7 \\
Avg. & 54.8 & 35.3 & 25.8 & 54.2 & 54.3 & 54.4 & 54.8 & 55.1 & 55.6 & 55.2 & 54.2 & 55.6 & 55.5 & 53.3 & 56.0 & 55.9 & 54.3 \\
\midrule
\rowcolor{gray!15} \multicolumn{18}{c}{\textbf{\emph{Open-source Reasoning Models, Prompted as Critic Models}}} \\
Sky-T1-32B-Flash & 65.4 & 71.2 & 68.9 & 61.7 & 65.8 & 65.5 & 65.4 & 66.7 & 66.0 & 67.2 & 64.1 & 66.2 & 66.1 & 63.1 & 69.4 & 66.6 & 64.9 \\
QwQ-32B & 39.0 & \textbf{85.7} & \textbf{89.9} & 40.1 & 40.6 & 40.1 & 39.4 & 38.7 & 40.2 & 38.7 & 41.3 & 38.5 & 39.6 & 38.8 & 36.6 & 36.8 & 38.3 \\
Marco-o1 & 54.3 & 24.3 & 25.2 & 49.3 & 55.1 & 53.8 & 55.7 & 55.2 & 57.2 & 54.1 & 53.5 & 55.6 & 55.9 & 51.1 & 56.0 & 55.8 & 53.9 \\
DeepSeek-R1-Distill-Llama-70B & 59.3 & 39.4 & 14.2 & 57.2 & 58.8 & 59.8 & 60.7 & 59.4 & 59.9 & 60.2 & 58.5 & 60.0 & 59.9 & 55.1 & 60.2 & 60.3 & 58.8 \\
Qwen3-8B & 57.5 & 60.4 & 57.9 & 54.4 & 57.8 & 58.1 & 59.0 & 57.5 & 56.6 & 58.5 & 56.3 & 59.1 & 59.3 & 54.8 & 58.3 & 57.6 & 57.0 \\
DeepSeek-R1-Distill-Qwen-32B & 47.4 & 75.0 & 72.9 & 48.9 & 48.4 & 48.9 & 46.9 & 45.6 & 43.6 & 47.9 & 47.4 & 47.3 & 47.3 & 49.6 & 44.9 & 48.5 & 44.1 \\
Avg. & 53.8 & 59.3 & 54.8 & 51.9 & 54.4 & 54.4 & 54.5 & 53.9 & 53.9 & 54.4 & 53.5 & 54.5 & 54.7 & 52.1 & 54.2 & 54.3 & 52.8 \\
\midrule
\rowcolor{gray!15} \multicolumn{18}{c}{\textbf{\emph{Frontier API Models, Prompted as Critic Models}}} \\
Claude-Opus-4.5 & 73.7 & 82.5 & \underline{83.4} & \textbf{69.0} & 73.6 & 73.5 & 73.8 & 75.2 & 72.7 & 76.5 & 73.9 & 73.8 & 74.9 & 71.4 & 78.5 & 76.0 & 71.2 \\
DeepSeek-V3.2 & 65.4 & 81.0 & 82.6 & 63.4 & 65.1 & 65.1 & 65.0 & 66.1 & 64.1 & 67.7 & 64.0 & 66.1 & 67.0 & 62.3 & 69.0 & 66.4 & 64.2 \\
GPT-5.4 & \textbf{75.4} & 79.0 & 83.0 & \underline{68.0} & \textbf{75.4} & \textbf{75.9} & \textbf{76.1} & \textbf{76.8} & \underline{75.1} & \textbf{78.0} & \textbf{75.2} & \textbf{75.9} & \textbf{75.8} & \textbf{74.0} & \textbf{79.6} & \textbf{77.8} & \textbf{74.0} \\
GPT-5.2 & \underline{74.6} & 76.9 & 81.4 & 66.3 & \underline{75.0} & \underline{75.3} & \underline{75.5} & \underline{76.3} & \textbf{75.2} & \underline{77.3} & \underline{74.3} & \underline{74.9} & \underline{75.1} & \underline{72.9} & \underline{79.1} & \underline{77.0} & \underline{73.4} \\
GLM-4.7 & 72.4 & 76.0 & 77.3 & 66.6 & 72.5 & 72.6 & 72.7 & 73.9 & 72.7 & 74.6 & 72.6 & 73.5 & 73.4 & 69.8 & 76.3 & 74.5 & 70.7 \\
Qwen3-Max & 70.4 & 70.4 & 74.6 & 64.3 & 69.5 & 70.3 & 71.1 & 71.9 & 71.8 & 73.6 & 69.2 & 71.5 & 71.9 & 65.3 & 76.5 & 72.4 & 70.0 \\
DeepSeek-V3.1 & 70.7 & 69.0 & 70.4 & 64.0 & 70.5 & 71.0 & 71.6 & 71.9 & 71.2 & 74.1 & 70.1 & 71.8 & 72.0 & 64.7 & 75.8 & 74.1 & 70.0 \\
DeepSeek-R1 & 67.8 & 64.0 & 64.5 & 63.7 & 66.8 & 67.5 & 67.9 & 69.6 & 68.6 & 70.4 & 66.1 & 68.9 & 68.9 & 61.6 & 72.8 & 70.3 & 67.4 \\
Gemini-3.1-Pro & 63.6 & 41.5 & 67.8 & 55.4 & 64.8 & 64.0 & 64.9 & 65.6 & 65.5 & 63.8 & 64.6 & 64.3 & 63.7 & 61.3 & 63.5 & 65.7 & 64.8 \\
Avg.                      & 70.4 & 71.1 & 76.1 & 64.5 & 70.4 & 70.6 & 71.0 & 71.9 & 70.8 & 72.9 & 70.0 & 71.2 & 71.4 & 67.0 & 74.6 & 72.7 & 69.5 \\
\midrule
\rowcolor{gray!15} \multicolumn{18}{c}{\textbf{\emph{Medical Process Reward Models}}} \\
Med-PRM-Reward v1.0 & 56.1 & \underline{83.8} & 65.8 & 57.1 & 55.6 & 58.5 & 58.5 & 55.7 & 56.1 & 57.1 & 59.4 & 58.7 & 55.7 & 55.0 & 53.2 & 57.9 & 53.8 \\
\rowcolor{gray!15}
Ours (in-distribution)    & 86.6 & 75.7 & 92.3 & 87.5 & 88.1 & 85.2 & 84.8 & 86.8 & 86.2 & 85.1 & 86.4 & 85.3 & 84.1 & 87.5 & 89.3 & 88.3 & 86.2 \\
\bottomrule
\end{tabular}
}
\end{table*}

Table~\ref{tab:error_detection_by_type_en} compares zero-shot medical error
detectors with the PRM trained on MedPRMBench.

\smallskip
\noindent\textbf{Frontier critics provide a strong zero-shot reference.}
GPT-5.4 leads the prompted critics at 75.4\% PRMScore, followed by GPT-5.2
(74.6\%) and Claude-Opus-4.5 (73.7\%); the two open-source critic groups average only
53.8--54.8\%. The released Med-PRM-Reward v1.0 reaches 56.1\%: its 83.8\%
erroneous-step detection coincides with frequent false alarms on correct steps.
Across frontier API critics, chains containing Non-Redundancy (64.5\%) and
Deception Resistance (67.0\%) errors are the hardest slices, while
Multi-Solution Consistency is the easiest (74.6\%).

\smallskip
\noindent\textbf{MedPRMBench supervision yields a strong medical PRM baseline.}
The Qwen3-8B PRM trained on the 6{,}870-question split reaches
86.6\% PRMScore, with $F1_{\mathrm{err}}{=}82.5$ and
$F1_{\mathrm{cor}}{=}90.8$. It detects 75.7\% of erroneous steps and flags the
gold first error in 92.3\% of corrupted chains. Under the stricter criterion
that no earlier step is falsely flagged, first-error localization is 89.0\%; on
clean chains, 2.0\% receive at least one false alarm
(Appendix~\ref{app:backbone_replication}). The released split thus provides a strong
supervised baseline for future medical PRMs.

\smallskip
\noindent\textbf{Missed prerequisites and safety checks persist after training.}
The overlapping taxonomy columns in Table~\ref{tab:error_detection_by_type_en}
describe chains containing each error type. Code-attributed recall, which uses
physician-finalized step--code assignments, provides a more targeted diagnostic
(Appendix~\ref{app:attributed_diagnostics}). The trained PRM detects 97.5\% of
Quantitative Correctness (E-4), 96.8\% of Multi-Solution Consistency (E-3), and
95.5\% of Deception Resistance (E-2) steps. The corresponding rates are 53.4\%
for Prerequisite Sensitivity (E-1), 67.2\% for Evidence-Based Soundness (R-1),
and 68.8\% for Safety Awareness (R-5): even after in-distribution training,
missed prerequisites and missed safety checks remain the main failure modes.


\subsection{Additional Analyses}
\label{sec:additional_analyses}

We probe how construction choices affect the PRM and where different verifiers
make errors.

\begin{wraptable}{r}{0.58\textwidth}
\vspace{-12pt}
\centering
\scriptsize
\setlength{\tabcolsep}{3pt}
\caption{Key data ablations (\%): mean $\pm$ SD over three seeds ($^{\dagger}$one seed); Table~\ref{tab:error_detection_by_type_en} reports seed 42 (86.6). Error rec./correct acc.\ use corrupted-chain steps; Clean FA is the share of clean chains with any false alarm. Details: Appendix~\ref{sec:ablation}.}
\label{tab:main_ablation}
\begin{tabular}{lcccc}
\toprule
\textbf{Configuration} & \textbf{PRMScore} & \textbf{Error rec.} & \textbf{Correct acc.} & \textbf{Clean FA}$\downarrow$ \\
\midrule
Full Model & $86.5{\pm}0.4$ & $76.7{\pm}2.6$ & $94.3{\pm}1.3$ & $2.1{\pm}0.6$ \\
w/o Physician Review & $84.8{\pm}0.1$ & $70.1{\pm}0.5$ & $96.3{\pm}0.2$ & $5.6{\pm}0.5$ \\
w/o Error Construction$^{\dagger}$ & 38.5 & 0.0 & 100.0 & 0.0 \\
\bottomrule
\end{tabular}
\end{wraptable}
\noindent\textbf{Physician review improves the training signal.}
Removing constructed errors eliminates erroneous-step supervision and produces
an all-positive classifier (Table~\ref{tab:main_ablation}). Across three matched
seeds, removing physician review lowers PRMScore by $1.72{\pm}0.33$ points and
error recall by $6.55{\pm}2.45$ points, while increasing clean-chain false alarms
by $3.54{\pm}0.91$ points. Without review, the PRM accepts slightly more correct
steps in corrupted chains but raises more clean-chain false alarms.

\noindent\textbf{Aggregate scores hide operating points.}
Within corrupted chains, QwQ-32B detects 85.7\% of erroneous steps but accepts
only 15.9\% of correct steps; Meditron3-8B shows the opposite pattern (9.7\%
and 96.4\%, respectively). Our PRM accepts 95.2\% of correct steps and detects
75.7\% of erroneous steps. Neither error recall nor PRMScore alone reveals this trade-off, so we report
both with clean-chain false alarms (Appendix~\ref{sec:inference_bias}).
On a 700-example Hard Subset with unchanged answers and low severity scores,
eight of nine frontier API critics score lower than on the full test set, by
1.5 PRMScore points on average and by up to 3.5 (Gemini-3.1-Pro) and 2.6 (GPT-5.4)
points (Appendix~\ref{sec:hard_subset}).

\noindent\textbf{Targeted data narrows the prerequisite and safety gaps.}
To measure type-specific training signal beyond training
set size, we compare leave-one-type-out runs with controls that remove the same
number of rows matched by source, severity, chain length, and error count.
Retaining rows containing E-1 raises attributed prerequisite-error recall by
7.38 points; retaining R-5-bearing rows raises safety-error recall by 20.21
points. The corresponding clean-chain false-alarm rates also rise by 1.29 and
1.57 points. At the aggregate level, only R-5 rows significantly raise PRMScore ($+0.57$):
type-level gains again remain largely invisible in the single score. All five
matched controls and ranking diagnostics are in
Appendix~\ref{app:attributed_diagnostics}.

\begin{wraptable}{r}{0.50\textwidth}
\vspace{-12pt}
\centering
\scriptsize
\setlength{\tabcolsep}{3pt}
\caption{Error-step recall (\%): expert-authored Class~A vs.\ generated Class~B; full-coverage critic protocol in Appendix~\ref{app:parse_coverage}.}
\label{tab:main_classab}
\vspace{-1pt}
\begin{tabular*}{\linewidth}{@{\extracolsep{\fill}}lccc}
\toprule
\textbf{Model} & \textbf{Class A} & \textbf{Class B} & \textbf{Gap (A$-$B)} \\
\midrule
Our PRM & 81.29 & 69.07 & $+$12.22 \\
DS-R1-Distill-Llama-70B & 47.29 & 30.00 & $+$17.29 \\
DS-R1-Distill-Qwen-32B & 73.07 & 77.23 & $-$4.16 \\
HuatuoGPT-o1-8B & 43.61 & 35.42 & $+$8.20 \\
Meerkat-8B & 39.99 & 35.87 & $+$4.12 \\
QwQ-32B & 83.85 & 87.90 & $-$4.04 \\
Qwen3-8B & 65.53 & 54.29 & $+$11.24 \\
\bottomrule
\end{tabular*}
\end{wraptable}
\noindent\textbf{Generated traces are not easier.}
The trained PRM detects 81.29\% of error steps in expert-authored Class~A traces
versus 69.07\% in generated Class~B traces (Table~\ref{tab:main_classab}).
Four of six exact-output critics show the same direction, so detection does not
reduce to recognizing corrupted generated text; see Appendix~\ref{app:classab}.

\noindent\textbf{Step scores help weight answer consensus.}
We also test whether the frozen PRM's step scores help select answers from
$N{=}64$ candidate chains on MedQA and MedMCQA. For three open-source policy
models, reward-weighted self-consistency improves accuracy over unweighted
majority voting in all six policy--dataset settings, by 0.24--1.10 points.
The largest gain, on UltraMedical with MedQA, remains significant after
correction for six comparisons ($+1.10$ points). In contrast, using the PRM
alone as a best-of-$N$ ranker underperforms self-consistency in five of six
settings, so the step scores are more useful for weighting consensus than for
standalone selection. Appendix~\ref{sec:prm_verifier} reports the full sampling
protocol and comparisons.


\section{Benchmark Utility and Generalization}
\label{sec:generalization}

We test whether the benchmark's training signal reproduces across backbones,
how it transfers to other benchmarks, and whether evaluation readouts remain
stable under complete-coverage and repeated-sampling checks.

\subsection{Scale and Cross-Family Replication}
\label{sec:backbone_replication}

We train Qwen3-4B and Meta-Llama-3.1-8B-Instruct with the same
split and fixed protocol as Qwen3-8B: one pre-specified epoch, with no checkpoint,
threshold, or architecture selected on test performance
(Appendix~\ref{app:backbone_replication}). Evaluated on all 13{,}000 test chains
(113{,}910 steps) without truncation, the three models obtain consistent
in-distribution PRMScores (86.6, 86.4, and 87.2, respectively;
Table~\ref{tab:main_backbone_replication}), so the training signal reproduces across
scale and model family. External transfer depends on the backbone: Qwen3-4B remains
close to Qwen3-8B, whereas Llama reaches 25.2 ProcessBench macro F1 and 56.1
PRMBench PRMScore. A higher in-distribution PRMScore therefore does not imply better
transfer: Llama has the highest MedPRMScore but the lowest external scores.

\begin{table}[H]
\centering
\footnotesize
\setlength{\tabcolsep}{3pt}
\caption{Scale and cross-family replication (\%). MedPRMBench metrics use the 6{,}500 corrupted chains, except Clean FA, the share of the 6{,}500 clean chains with any false alarm; Strict first requires the earliest predicted error to equal the gold first error. PB and PRMB are no-further-training ProcessBench macro F1 and official PRMBench PRMScore. One run per backbone (seed 42).}
\label{tab:main_backbone_replication}
\begin{tabular}{lccccccc}
\toprule
\textbf{Backbone} & \textbf{MedPRMScore} & \textbf{Error rec.} & \textbf{Correct acc.} & \textbf{Strict first} & \textbf{Clean FA}$\downarrow$ & \textbf{PB} & \textbf{PRMB} \\
\midrule
Qwen3-8B & 86.6 & 75.7 & 95.2 & 89.0 & 2.0 & 55.3 & 64.3 \\
Qwen3-4B & 86.4 & 83.0 & 89.8 & 87.4 & 6.0 & 50.7 & 64.4 \\
Llama-3.1-8B-Instruct & 87.2 & 81.1 & 92.3 & 88.7 & 2.5 & 25.2 & 56.1 \\
\bottomrule
\end{tabular}
\end{table}

The benchmark also distinguishes operating points behind similar PRMScores:
Qwen3-4B detects 83.0\% of erroneous steps versus Qwen3-8B's 75.7\%, but
recognizes fewer correct steps (89.8\% versus 95.2\%), raises more false
alarms on clean chains (6.0\% versus 2.0\%), and has lower strict first-error
localization (87.4\% versus 89.0\%). Reporting error recall alongside
clean-chain false alarms exposes this trade-off, which PRMScore alone hides.
Llama lies between the two Qwen models on all four of these readouts, none of
which anticipates its weaker external transfer; cross-domain behavior must
therefore be measured directly.

\needspace{6\baselineskip}
\subsection{External Benchmark Stress Tests}
\label{sec:transfer}

\begin{wraptable}{r}{0.50\textwidth}
\vspace{-12pt}
\centering
\scriptsize
\setlength{\tabcolsep}{3pt}
\caption{No-further-training transfer (PB: ProcessBench macro F1; PRMB: PRMBench PRMScore). Published values~\citep{zheng2024processbench,song2025prmbench}; $^{*}$our reproduction, official harness; $^{\dagger}$before training (App.~\ref{app:reverse_transfer}, \ref{app:matched_base_control}).}
\label{tab:transfer}
\vspace{-1pt}
\begin{tabular}{lcc}
\toprule
\textbf{Model} & \textbf{PB} & \textbf{PRMB} \\
\midrule
Qwen2.5-Math-PRM-7B & 73.1$^{*}$ & 65.5 \\
Qwen2.5-Math-7B-PRM800K & 56.5 & 58.3$^{*}$ \\
Pure-PRM-7B & 56.7$^{*}$ & 65.3 \\
Skywork-PRM-7B & 42.1 & 65.1 \\
Skywork-PRM-1.5B & 36.4 & 61.1 \\
ReasonEval-7B & 35.3$^{*}$ & 60.1 \\
Math-Shepherd-PRM-7B & 31.5 & 47.0 \\
RLHFlow-PRM-Mistral-8B & 28.4 & 54.4 \\
RLHFlow-PRM-Deepseek-8B & 26.6 & 54.2 \\
\midrule
Qwen3-8B, untrained$^{\dagger}$ & 10.8 & 29.6 \\
MedPRM Qwen3-8B (ours) & \textbf{55.3} & \textbf{64.3} \\
\bottomrule
\end{tabular}
\vspace{-6pt}
\end{wraptable}

We evaluate the MedPRMBench-trained Qwen3-8B, \emph{with no further training},
on ProcessBench~\citep{zheng2024processbench} (macro F1 over four subsets) and
PRMBench~\citep{song2025prmbench} (official PRMScore using the authors'
\texttt{mr\_eval} harness). Table~\ref{tab:transfer} places these results beside
math-trained open PRMs, filling unreported cells with our reproductions under the official harnesses.

Without further training, the frozen Qwen3-8B reaches 55.3 ProcessBench macro
F1 and 64.3 PRMBench PRMScore, although it never sees mathematical process
labels. Training lifts the same initialization from 10.8 to 55.3 on ProcessBench and
from 29.6 to 64.3 on PRMBench. Among the ten PRMs in Table~\ref{tab:transfer}, it ranks
fourth on both benchmarks, is statistically tied on ProcessBench with the second- and
third-ranked PURE-PRM-7B and Qwen2.5-Math-7B-PRM800K (paired-bootstrap 95\% CIs
include zero), and exceeds the latter by 6.0 points on PRMBench, where it trails the three
higher-ranked PRMs by 0.8--1.2 points (CIs exclude zero). On ProcessBench it also exceeds far larger prompted critics such as Qwen2.5-32B-Instruct (49.3) and Llama-3.1-70B-Instruct (52.7)~\citep{zheng2024processbench}.
The external benchmarks expose distinct operating points: ProcessBench correct-step
accuracy and error recall are 95.4/53.1\%; on PRMBench they are 97.8/24.7\%. A
same-backbone diagnostic at approximately matched correct-step acceptance
(97.8--97.9\%) yields 24.7\% error recall for the trained model versus
7.6--8.5\% for the untrained Qwen3-8B initialization under two prompted scoring
interfaces, with base thresholds cross-fitted on held-out folds
(Appendix~\ref{app:matched_base_control}). In the reverse direction, the best
mathematical PRM that passes its native-interface positive control reaches 62.7 MedPRMScore and 31.7\% error recall on
MedPRMBench (Appendix~\ref{app:reverse_transfer}). Transfer is thus asymmetric: mathematical
PRMs miss most clinical errors, whereas the medical PRM stays competitive on
mathematical benchmarks.

\subsection{Evaluation Protocol Robustness}
\label{sec:protocol_robustness}

All reported models are scored on the same fixed population. The six open
generative critics use an exact single-label-per-step output protocol that
covers all 6{,}500 corrupted chains and 59{,}174 steps with zero parse failures,
and all trained PRMs score all 13{,}000 chains and 113{,}910 steps without
truncation (Appendices~\ref{app:parse_coverage} and~\ref{app:backbone_replication}).
Because frontier API critics yield generated labels rather than step
probabilities, we check their stability with five independent completions on an
800-question stratified subset: across four frontier critics, mean-score or
majority-vote aggregation changes PRMScore by at most $+0.87$ points, and
per-replicate ranges are 0.24--0.39 points (Appendices~\ref{app:api_protocol}
and~\ref{app:variance}). The fixed split, checkpoint, prompts, and decision
threshold form a reproducible protocol.

\section{Conclusion}
\label{sec:conclusion}

MedPRMBench provides physician-reviewed labels for 14 clinical error types
across 6{,}500 questions and 113{,}910 steps. The best prompted critic reaches
75.4\% PRMScore and the two open-source critic groups average 53.8--54.8\%, leaving
substantial headroom for zero-shot clinical step verification. A Qwen3-8B PRM trained
on the benchmark's split reaches 86.6\% PRMScore and 89.0\% strict first-error
localization, and physician review contributes measurably to this result. Missed
prerequisites (E-1) and safety checks (R-5) remain its weakest error types,
marking where medical verifiers most need improvement; without further training,
the same model remains competitive with math-trained PRMs on ProcessBench and
PRMBench.
These findings suggest that medical verifiers should be reported by error type
and operating point, not by a single aggregate score. Because the release
retains initial physician annotations, second-stage corrections, model votes,
and filtering reasons, future work can also re-derive labels under alternative
adjudication rules and study where clinicians disagree. The benchmark is
English-only, text-only, and built from constructed rather than naturally
occurring errors; extending it along these axes is the main direction for
future work (Appendix~\ref{app:limitations}).
\label{main:text:end}

\bibliographystyle{iclr2027_conference}
\bibliography{custom}

@inproceedings{lightman2023lets,
  title     = {Let's Verify Step by Step},
  author    = {Hunter Lightman and Vineet Kosaraju and Yura Burda and Harri Edwards and Bowen Baker and Teddy Lee and Jan Leike and John Schulman and Ilya Sutskever and Karl Cobbe},
  booktitle = {The Twelfth International Conference on Learning Representations (ICLR)},
  year      = {2024},
  url       = {https://openreview.net/forum?id=v8L0pN6EOi},
}

@article{uesato2022solving,
  title   = {Solving Math Word Problems with Process- and Outcome-Based Feedback},
  author  = {Jonathan Uesato and Nate Kushman and Ramana Kumar and Francis Song and Noah Siegel and Lisa Wang and Antonia Creswell and Geoffrey Irving and Irina Higgins},
  journal = {arXiv preprint arXiv:2211.14275},
  year    = {2022},
}

@inproceedings{ouyang2022training,
  title     = {Training Language Models to Follow Instructions with Human Feedback},
  author    = {Long Ouyang and Jeffrey Wu and Xu Jiang and Diogo Almeida and Carroll Wainwright and Pamela Mishkin and Chong Zhang and Sandhini Agarwal and Katarina Slama and Alex Ray and John Schulman and Jacob Hilton and Fraser Kelton and Luke Miller and Maddie Simens and Amanda Askell and Peter Welinder and Paul F. Christiano and Jan Leike and Ryan Lowe},
  booktitle = {Advances in Neural Information Processing Systems},
  volume    = {35},
  pages     = {27730--27744},
  year      = {2022},
}

@inproceedings{wang2023mathshepherd,
  title     = {{Math-Shepherd}: Verify and Reinforce {LLMs} Step-by-step without Human Annotations},
  author    = {Peiyi Wang and Lei Li and Zhihong Shao and Runxin Xu and Damai Dai and Yifei Li and Deli Chen and Yu Wu and Zhifang Sui},
  booktitle = {Proceedings of the 62nd Annual Meeting of the Association for Computational Linguistics (Volume 1: Long Papers)},
  pages     = {9426--9439},
  year      = {2024},
  publisher = {Association for Computational Linguistics},
  doi       = {10.18653/v1/2024.acl-long.510},
  url       = {https://aclanthology.org/2024.acl-long.510/},
}

@article{snell2024scaling,
  title   = {Scaling {LLM} Test-Time Compute Optimally can be More Effective than Scaling Model Parameters},
  author  = {Charlie Snell and Jaehoon Lee and Kelvin Xu and Aviral Kumar},
  journal = {arXiv preprint arXiv:2408.03314},
  year    = {2024},
}

@article{setlur2024rewarding,
  title   = {Rewarding Progress: Scaling Automated Process Verifiers for {LLM} Reasoning},
  author  = {Amrith Setlur and Chirag Nagpal and Adam Fisch and Xinyang Geng and Jacob Eisenstein and Rishabh Agarwal and Alekh Agarwal and Jonathan Berant and Aviral Kumar},
  journal = {arXiv preprint arXiv:2410.08146},
  year    = {2024},
}

@inproceedings{xia2024reasoneval,
  title     = {Evaluating Mathematical Reasoning Beyond Accuracy},
  author    = {Shijie Xia and Xuefeng Li and Yixin Liu and Tongshuang Wu and Pengfei Liu},
  booktitle = {Proceedings of the AAAI Conference on Artificial Intelligence},
  volume    = {39},
  pages     = {27723--27730},
  year      = {2025},
  doi       = {10.1609/aaai.v39i26.34987},
  url       = {https://doi.org/10.1609/aaai.v39i26.34987},
}

@inproceedings{yun2025medprm,
  title     = {{Med-PRM}: Medical Reasoning Models with Stepwise, Guideline-verified Process Rewards},
  author    = {Jaehoon Yun and Jiwoong Sohn and Jungwoo Park and Hyunjae Kim and Xiangru Tang and Daniel Shao and Yong Hoe Koo and {Ko Minhyeok} and Qingyu Chen and Mark Gerstein and Michael Moor and Jaewoo Kang},
  booktitle = {Proceedings of the 2025 Conference on Empirical Methods in Natural Language Processing},
  pages     = {16554--16571},
  year      = {2025},
  publisher = {Association for Computational Linguistics},
  doi       = {10.18653/v1/2025.emnlp-main.837},
  url       = {https://aclanthology.org/2025.emnlp-main.837/},
}

@article{mu2025medceg,
  title   = {{MedCEG}: Reinforcing Verifiable Medical Reasoning with Critical Evidence Graph},
  author  = {Linjie Mu and Yannian Gu and Zhongzhen Huang and Yakun Zhu and Shaoting Zhang and Xiaofan Zhang},
  journal = {arXiv preprint arXiv:2512.13510},
  year    = {2025},
}

@inproceedings{jiang2025meds3,
  title     = {{MedS}$^3$: Towards Medical Slow Thinking with Self-Evolved Soft Dual-sided Process Supervision},
  author    = {Shuyang Jiang and Yusheng Liao and Zhe Chen and Ya Zhang and Yanfeng Wang and Yu Wang},
  booktitle = {Proceedings of the AAAI Conference on Artificial Intelligence},
  volume    = {40},
  pages     = {31319--31327},
  year      = {2026},
  doi       = {10.1609/aaai.v40i37.40395},
  url       = {https://doi.org/10.1609/aaai.v40i37.40395},
}

@article{wu2025cot,
  title   = {Why Chain of Thought Fails in Clinical Text Understanding},
  author  = {Jiageng Wu and Kevin Xie and Bowen Gu and Nils Kr{\"u}ger and Kueiyu Joshua Lin and Jie Yang},
  journal = {arXiv preprint arXiv:2509.21933},
  year    = {2025},
}

@inproceedings{song2025prmbench,
  title     = {{PRMBench}: A Fine-grained and Challenging Benchmark for Process-Level Reward Models},
  author    = {Mingyang Song and Zhaochen Su and Xiaoye Qu and Jiawei Zhou and Yu Cheng},
  booktitle = {Proceedings of the 63rd Annual Meeting of the Association for Computational Linguistics (Volume 1: Long Papers)},
  pages     = {25299--25346},
  year      = {2025},
  publisher = {Association for Computational Linguistics},
  doi       = {10.18653/v1/2025.acl-long.1230},
  url       = {https://aclanthology.org/2025.acl-long.1230/},
}

@inproceedings{zheng2024processbench,
  title     = {{ProcessBench}: Identifying Process Errors in Mathematical Reasoning},
  author    = {Chujie Zheng and Zhenru Zhang and Beichen Zhang and Runji Lin and Keming Lu and Bowen Yu and Dayiheng Liu and Jingren Zhou and Junyang Lin},
  booktitle = {Proceedings of the 63rd Annual Meeting of the Association for Computational Linguistics (Volume 1: Long Papers)},
  pages     = {1009--1024},
  year      = {2025},
  publisher = {Association for Computational Linguistics},
  doi       = {10.18653/v1/2025.acl-long.50},
  url       = {https://aclanthology.org/2025.acl-long.50/},
}

@article{zeng2023mrgsm8k,
  title   = {{MR-GSM8K}: A Meta-Reasoning Benchmark for Large Language Model Evaluation},
  author  = {Zhongshen Zeng and Pengguang Chen and Shu Liu and Haiyun Jiang and Jiaya Jia},
  journal = {arXiv preprint arXiv:2312.17080},
  year    = {2023},
}

@article{liu2024rmbench,
  title   = {{RM-Bench}: Benchmarking Reward Models of Language Models with Subtlety and Style},
  author  = {Yantao Liu and Zijun Yao and Rui Min and Yixin Cao and Lei Hou and Juanzi Li},
  journal = {arXiv preprint arXiv:2410.16184},
  year    = {2024},
}

@inproceedings{lin2024criticbench,
  title     = {{CriticBench}: Benchmarking {LLMs} for Critique-Correct Reasoning},
  author    = {Zicheng Lin and Zhibin Gou and Tian Liang and Ruilin Luo and Haowei Liu and Yujiu Yang},
  booktitle = {Findings of the Association for Computational Linguistics: ACL 2024},
  pages     = {1552--1587},
  year      = {2024},
  publisher = {Association for Computational Linguistics},
  doi       = {10.18653/v1/2024.findings-acl.91},
  url       = {https://aclanthology.org/2024.findings-acl.91/},
}

@inproceedings{zeng2024mrben,
  title     = {{MR-Ben}: A Meta-Reasoning Benchmark for Evaluating System-2 Thinking in {LLMs}},
  author    = {Zhongshen Zeng and Yinhong Liu and Yingjia Wan and Jingyao Li and Pengguang Chen and Jianbo Dai and Yuxuan Yao and Rongwu Xu and Zehan Qi and Wanru Zhao and Linling Shen and Jianqiao Lu and Haochen Tan and Yukang Chen and Hao Zhang and Zhan Shi and Bailin Wang and Zhijiang Guo and Jiaya Jia},
  booktitle = {Advances in Neural Information Processing Systems 37},
  pages     = {119466--119546},
  year      = {2024},
  doi       = {10.52202/079017-3797},
  url       = {https://doi.org/10.52202/079017-3797},
}

@article{jaech2024openai,
  title   = {{OpenAI} o1 System Card},
  author  = {{OpenAI}},
  journal = {arXiv preprint arXiv:2412.16720},
  year    = {2024},
}

@article{guo2025deepseek,
  title   = {{DeepSeek-R1}: Incentivizing Reasoning Capability in {LLMs} via Reinforcement Learning},
  author  = {{DeepSeek-AI}},
  journal = {arXiv preprint arXiv:2501.12948},
  year    = {2025},
}

@article{yang2025qwen3,
  title   = {Qwen3 Technical Report},
  author  = {An Yang and Anfeng Li and Baosong Yang and Beichen Zhang and Binyuan Hui and Bo Zheng and Bowen Yu and Chang Gao and Chengen Huang and Chenxu Lv and others},
  journal = {arXiv preprint arXiv:2505.09388},
  year    = {2025},
}

@misc{alibabacloud2026qwen3max,
  title  = {{qwen3-max} Model Info},
  author = {{Alibaba Cloud}},
  year   = {2026},
  url    = {https://www.alibabacloud.com/help/en/model-studio/model-qwen3-max},
  note   = {Alibaba Cloud Model Studio documentation, updated September 11, 2026},
}

@article{shao2024deepseekmath,
  title   = {{DeepSeekMath}: Pushing the Limits of Mathematical Reasoning in Open Language Models},
  author  = {Zhihong Shao and Peiyi Wang and Qihao Zhu and Runxin Xu and Junxiao Song and Xiao Bi and Haowei Zhang and Mingchuan Zhang and Y. K. Li and Y. Wu and Daya Guo},
  journal = {arXiv preprint arXiv:2402.03300},
  year    = {2024},
}

@misc{qwq32b,
  title  = {{QwQ-32B}: Embracing the Power of Reinforcement Learning},
  author = {{Qwen Team}},
  year   = {2025},
  month  = mar,
  url    = {https://qwenlm.github.io/blog/qwq-32b/},
}

@misc{sky_t1,
  author = {{NovaSky Team}},
  title  = {{Sky-T1}: Train your own {O1} preview model within \$450},
  year   = {2025},
  month  = jan,
  url    = {https://novasky-ai.github.io/posts/sky-t1/},
}

@article{marco_o1,
  title   = {Marco-o1: Towards Open Reasoning Models for Open-Ended Solutions},
  author  = {Yu Zhao and Huifeng Yin and Bo Zeng and Hao Wang and Tianqi Shi and Chenyang Lyu and Longyue Wang and Weihua Luo and Kaifu Zhang},
  journal = {arXiv preprint arXiv:2411.14405},
  year    = {2024},
}

@article{txgemma,
  title   = {{TxGemma}: Efficient and Agentic {LLMs} for Therapeutics},
  author  = {Eric Wang and Samuel Schmidgall and Paul F. Jaeger and Fan Zhang and Rory Pilgrim and Yossi Matias and Joelle Barral and David Fleet and Shekoofeh Azizi},
  journal = {arXiv preprint arXiv:2504.06196},
  year    = {2025},
}

@article{meerkat,
  title     = {Small language models learn enhanced reasoning skills from medical textbooks},
  author    = {Hyunjae Kim and Hyeon Hwang and Jiwoo Lee and Sihyeon Park and Dain Kim and Taewhoo Lee and Chanwoong Yoon and Jiwoong Sohn and Jungwoo Park and Olga Reykhart and Thomas Fetherston and Donghee Choi and Soo Heon Kwak and Qingyu Chen and Jaewoo Kang},
  journal   = {npj Digital Medicine},
  volume    = {8},
  number    = {1},
  pages     = {240},
  year      = {2025},
  doi       = {10.1038/s41746-025-01653-8},
}

@inproceedings{ultramedical,
  title     = {{UltraMedical}: Building Specialized Generalists in Biomedicine},
  author    = {Kaiyan Zhang and Sihang Zeng and Ermo Hua and Ning Ding and Zhang-Ren Chen and Zhiyuan Ma and Haoxin Li and Ganqu Cui and Biqing Qi and Xuekai Zhu and Xingtai Lv and Jin-Fang Hu and Zhiyuan Liu and Bowen Zhou},
  booktitle = {Advances in Neural Information Processing Systems 37},
  pages     = {26045--26081},
  year      = {2024},
  doi       = {10.52202/079017-0819},
  url       = {https://doi.org/10.52202/079017-0819},
}

@article{huatuogpt_o1,
  title   = {{HuatuoGPT-o1}, Towards Medical Complex Reasoning with {LLMs}},
  author  = {Junying Chen and Zhenyang Cai and Ke Ji and Xidong Wang and Wanlong Liu and Rongsheng Wang and Jianye Hou and Benyou Wang},
  journal = {arXiv preprint arXiv:2412.18925},
  year    = {2024},
}

@misc{llama3meditron,
  title  = {{Llama-3.1-Meditron3-8B}},
  author = {{LiGHT}},
  year   = {2024},
  url    = {https://huggingface.co/EPFLiGHT/Meditron3-8B},
  note   = {Model card},
}

@article{jin2021disease,
  title     = {What Disease does this Patient Have? {A} Large-scale Open Domain Question Answering Dataset from Medical Exams},
  author    = {Di Jin and Eileen Pan and Nassim Oufattole and Wei-Hung Weng and Hanyi Fang and Peter Szolovits},
  journal   = {Applied Sciences},
  volume    = {11},
  number    = {14},
  pages     = {6421},
  year      = {2021},
  publisher = {MDPI},
  doi       = {10.3390/app11146421},
  url       = {https://doi.org/10.3390/app11146421},
}

@inproceedings{pal2022medmcqa,
  title     = {{MedMCQA}: A Large-scale Multi-Subject Multi-Choice Dataset for Medical Domain Question Answering},
  author    = {Ankit Pal and Logesh Kumar Umapathi and Malaikannan Sankarasubbu},
  booktitle = {Proceedings of the Conference on Health, Inference, and Learning (CHIL)},
  series    = {Proceedings of Machine Learning Research},
  volume    = {174},
  pages     = {248--260},
  year      = {2022},
  publisher = {PMLR},
  url       = {https://proceedings.mlr.press/v174/pal22a.html},
}

@inproceedings{hendrycks2021measuring,
  title     = {Measuring Massive Multitask Language Understanding},
  author    = {Dan Hendrycks and Collin Burns and Steven Basart and Andy Zou and Mantas Mazeika and Dawn Song and Jacob Steinhardt},
  booktitle = {International Conference on Learning Representations (ICLR)},
  year      = {2021},
  url       = {https://openreview.net/forum?id=d7KBjmI3GmQ},
}

@inproceedings{jin2019pubmedqa,
  title     = {{PubMedQA}: A Dataset for Biomedical Research Question Answering},
  author    = {Qiao Jin and Bhuwan Dhingra and Zhengping Liu and William Cohen and Xinghua Lu},
  booktitle = {Proceedings of the 2019 Conference on Empirical Methods in Natural Language Processing and the 9th International Joint Conference on Natural Language Processing (EMNLP-IJCNLP)},
  pages     = {2567--2577},
  year      = {2019},
  publisher = {Association for Computational Linguistics},
  doi       = {10.18653/v1/D19-1259},
  url       = {https://aclanthology.org/D19-1259/},
}

@article{medexpqa2024,
  title   = {{MedExpQA}: Multilingual Benchmarking of Large Language Models for Medical Question Answering},
  author  = {I{\~n}igo Alonso and Maite Oronoz and Rodrigo Agerri},
  journal = {Artificial Intelligence in Medicine},
  volume  = {155},
  pages   = {102938},
  year    = {2024},
  doi     = {10.1016/j.artmed.2024.102938},
}

@inproceedings{zuo2025medxpertqa,
  title     = {{MedXpertQA}: Benchmarking Expert-Level Medical Reasoning and Understanding},
  author    = {Yuxin Zuo and Shang Qu and Yifei Li and Zhang-Ren Chen and Xuekai Zhu and Ermo Hua and Kaiyan Zhang and Ning Ding and Bowen Zhou},
  booktitle = {Proceedings of the 42nd International Conference on Machine Learning},
  series    = {Proceedings of Machine Learning Research},
  volume    = {267},
  pages     = {80961--80990},
  year      = {2025},
  publisher = {PMLR},
  url       = {https://proceedings.mlr.press/v267/zuo25a.html},
}

@article{wu2025medcasereasoning,
  title   = {{MedCaseReasoning}: Evaluating and Learning Diagnostic Reasoning from Clinical Case Reports},
  author  = {Kevin Wu and Eric Wu and Rahul Thapa and Kevin Wei and Angela Zhang and Arvind Suresh and Jacqueline J. Tao and Min Woo Sun and Alejandro Lozano and James Zou},
  journal = {arXiv preprint arXiv:2505.11733},
  year    = {2025},
}

@article{aho1972transitive,
  title     = {The Transitive Reduction of a Directed Graph},
  author    = {Alfred V. Aho and Michael R. Garey and Jeffrey D. Ullman},
  journal   = {SIAM Journal on Computing},
  volume    = {1},
  number    = {2},
  pages     = {131--137},
  year      = {1972},
  doi       = {10.1137/0201008},
}

\section*{Ethics Statement}

This research uses publicly available medical QA datasets that do not contain identifiable patient information. All data sources used in MedPRMBench construction are established academic benchmarks used under their respective licenses. The expert review process was conducted by qualified medical professionals who provided informed consent for their participation. We acknowledge that medical AI evaluation benchmarks, including MedPRMBench, should not be used as a sole basis for clinical deployment decisions; comprehensive clinical validation remains essential before any AI system is used in patient care settings. Our benchmark is designed to advance research in medical AI safety evaluation and should be used in conjunction with other safety assessment frameworks.

\section*{AI Assistance Disclosure}

Large language models were used as research tools to generate candidate reasoning traces, structured reasoning representations, controlled candidate errors, consistency signals, and selected model-evaluation outputs, as described in the paper. AI tools also assisted with literature discovery and manuscript drafting and editing. Research outputs were subjected to the deterministic checks, physician review where specified, and author verification described in our protocols. The authors reviewed the complete manuscript and take full responsibility for its content, claims, and citations.

\section*{Reproducibility Statement}

For the double-blind review version, all author-identifying repository names, acknowledgments, and institutional metadata are removed. The public artifact release will include the train/evaluation split files, source-dataset identifiers and reconstruction scripts where redistribution is license-restricted, the physician annotation guideline, error-injection and evaluation prompts, evaluation scripts, quality-control audit fields, and the benchmark-specific PRM checkpoint. If a source dataset's license does not permit redistributing derived text directly, we will release stable source IDs plus reconstruction code and provide an evaluation server for the corresponding held-out split. The release will preserve the fixed evaluation split used in this paper and will not require using the test set for prompt, threshold, or checkpoint selection. A machine-readable manifest records the nine overlap removals, integrity checks, and SHA-256 hashes for both split files. Construction details and thresholds are specified in Appendices~\ref{app:blueprint_details}--\ref{app:dataset_splitting}, PRM training hyperparameters in Appendix~\ref{app:prm_details}, evaluation protocols and prompts in Appendix~\ref{app:model_details} and Appendix~\ref{app:prompt_eval}, and the transfer analysis in Section~\ref{sec:generalization}.

\appendix


\section{Limitations}
\label{app:limitations}

MedPRMBench evaluates English, text-only reasoning. Some traces and candidate
errors are model-generated before deterministic checks and physician review;
their distribution may differ from naturally occurring clinical mistakes.
The step-only and question-replacement probes characterize local signal and
input sensitivity under their stated protocols
(Appendices~\ref{app:shortcut_probe} and~\ref{app:context_counterfactual}).

The 14-type taxonomy defines the benchmark's coverage. Its rule-derived
severity grades describe constructed errors rather than independently rated
clinical harm. The independent physician agreement estimate comes from
Class~A expert-authored traces; Class~B generated traces undergo the same
physician validation workflow but are outside that agreement sample.

The source questions come from general medical QA benchmarks, so specialized
settings such as radiology reporting or surgical planning, multilingual clinical
reasoning, and multimodal inputs such as images and laboratory tables are not
covered. Severity is a four-level rule-based simplification, and most
constructed errors fall in the Major or Critical grades, which limits
severity-stratified conclusions. A physician-annotated subset of naturally
occurring model errors would test whether constructed-error performance
predicts detection of real failures.

The exact-output protocol provides complete coverage for six open critics but changes
their response protocol relative to unconstrained free-form critique. Our
fixed-threshold primary metrics allow comparison with label-only API models;
for models with continuous scores, ranking metrics and validation-only
calibration also allow threshold-free comparison. The R-2
diagnostic in Appendix~\ref{app:attributed_diagnostics} illustrates why
fixed-threshold sensitivity and ranking can differ.


\section{Detailed Related Work}
\label{app:detailed_related_work}

\subsection{Process-Level Reward Models}
\label{sec:rw_prm}

Reward models are central to advancing large language models (LLMs) beyond pre-training and supervised fine-tuning, playing key roles in both reinforcement learning-based alignment and test-time compute scaling~\citep{ouyang2022training}. Based on supervision granularity, reward models fall into two categories: \emph{Outcome Reward Models} (ORMs) assign a single correctness score to an entire reasoning trace, while \emph{Process Reward Models} (PRMs) independently score each intermediate reasoning step~\citep{lightman2023lets,uesato2022solving}. \citet{lightman2023lets} constructed the PRM800K dataset through human annotation, demonstrating for the first time that PRMs significantly outperform ORMs on mathematical reasoning tasks. However, human annotation is costly and difficult to scale~\citep{setlur2024rewarding}.

To reduce annotation costs, subsequent work has explored various automatic labeling strategies. Inspired by Monte Carlo Tree Search (MCTS), Math-Shepherd~\citep{wang2023mathshepherd} estimates the quality of each reasoning step by sampling continuations and checking how often they reach the correct final answer. ReasonEval~\citep{xia2024reasoneval} leverages model-generated reasoning trajectories for automatic evaluation. In the medical domain, MedS$^3$~\citep{jiang2025meds3} adapts the MCTS approach to clinical QA, constructing a domain-specific PRM. However, as \citet{yun2025medprm} point out, MCTS-based labeling strategies suffer from a fundamental limitation: they tend to undervalue early reasoning steps that are logically sound but fail to lead to the correct final answer, thereby distorting the learning signal.

Med-PRM~\citep{yun2025medprm} proposes a retrieval-augmented process reward modeling framework that employs a RAG-AS-A-JUDGE approach for stepwise evaluation, reporting over 80\% MedQA accuracy with 8B-parameter models. MedCEG~\citep{mu2025medceg} takes a different approach, proposing a graph-based alignment framework that converts clinical narratives into traceable Critical Evidence Graphs (CEGs) and designs a Clinical Reasoning Procedure Reward evaluating reasoning quality along three dimensions: Node Coverage, Structural Correctness, and Chain Completeness.

All of the above works focus on PRM \emph{training methodologies}---how to construct better reward signals to improve model reasoning capabilities. Notably, reinforcement learning for medical reasoning has also attracted growing interest. Methods such as GRPO~\citep{shao2024deepseekmath} reduce computational overhead by eliminating auxiliary value models, but when paired with outcome-oriented reward functions, models may learn to exploit misleading patterns in the data to take shortcuts~\citep{mu2025medceg,wu2025cot}---arriving at conclusions that appear correct but are based on clinically invalid reasoning processes. This phenomenon further underscores the necessity of systematic PRM evaluation. In contrast to the training-focused works above, our work addresses how to \emph{systematically evaluate} PRMs' ability to detect various types of errors in medical reasoning: beyond assessing whether models can arrive at correct answers, we must evaluate whether they can identify errors in the reasoning process itself.

\begin{table*}[t]
\centering
\caption{Positioning of MedPRMBench against representative PRM training methods and reasoning-process benchmarks. Prior medical PRM works primarily study how to train reward or reasoning models, while prior PRM benchmarks primarily target mathematics or general critique. MedPRMBench focuses on medical step-level error-detection evaluation with clinical error taxonomy and severity labels.}
\label{tab:related_positioning}
\begingroup
\scriptsize
\setlength{\tabcolsep}{2pt}
\begin{tabularx}{\textwidth}{>{\raggedright\arraybackslash}p{1.65cm} >{\raggedright\arraybackslash}p{1.0cm} >{\raggedright\arraybackslash}X >{\raggedright\arraybackslash}p{1.4cm} >{\raggedright\arraybackslash}p{1.7cm} >{\raggedright\arraybackslash}p{0.9cm} >{\raggedright\arraybackslash}p{1.7cm} >{\raggedright\arraybackslash}p{1.6cm}}
\toprule
\textbf{Work} & \textbf{Domain} & \textbf{Primary focus} & \textbf{Step labels} & \textbf{Taxonomy} & \textbf{Severity} & \textbf{Physician review} & \textbf{Benchmark} \\
\midrule
Med-PRM~\citep{yun2025medprm} & Medical & Stepwise guideline-verified process rewards for training medical reasoners/reward models & Yes & No fine-grained taxonomy & No & Guideline/RAG-based judging & Training resource \\
MedCEG~\citep{mu2025medceg} & Medical & Critical Evidence Graph alignment and reward design for verifiable medical reasoning & Process dimensions & No fine-grained taxonomy & No & Not central & Training method \\
MedS$^3$~\citep{jiang2025meds3} & Medical & Search/self-evolved process supervision for medical slow thinking & Process supervision & No fine-grained taxonomy & No & Not central & Training method \\
Process\-Bench~\citep{zheng2024processbench} & Math & Step-level process-error localization benchmark & Yes & Single error notion & No & Human annotation & Evaluation benchmark \\
PRM\-Bench~\citep{song2025prmbench} & Math & Fine-grained PRM evaluation over designed mathematical reasoning errors & Yes & 9 types & No & Human annotation & Evaluation benchmark \\
\textbf{MedPRMBench} & Medical & Step-level PRM error-detection evaluation with controlled errors & 113{,}910 labels & 14 medical/general types & 4 levels & Full-corpus physician review & Evaluation benchmark + baseline \\
\bottomrule
\end{tabularx}
\endgroup
\end{table*}

\subsection{Benchmarks for Reasoning Process Evaluation}
\label{sec:rw_benchmark}

Benchmarks for evaluating reasoning process quality have attracted growing attention in recent years. Early efforts such as MR-GSM8K~\citep{zeng2023mrgsm8k} constructed mathematical datasets with erroneous reasoning steps through human annotation, but only support detection of a single error type. RMBench~\citep{liu2024rmbench} and CriticBench~\citep{lin2024criticbench} evaluate models' capabilities as critics but do not provide step-level labels. MR-Ben~\citep{zeng2024mrben} extends the evaluation scale but remains limited to a single error category.

ProcessBench~\citep{zheng2024processbench} is a widely used process-level benchmark designed for PRMs, comprising 3{,}400 human-annotated mathematical reasoning samples with step-level error localization. However, its error taxonomy contains only a single category, unable to distinguish different types of reasoning defects. PRMBench~\citep{song2025prmbench} achieves a significant advance by proposing a fine-grained error taxonomy spanning three major categories---Simplicity, Soundness, and Sensitivity---with nine sub-categories, containing 6{,}216 carefully designed problems and 83{,}456 step-level labels. Its systematic evaluation of 25 models reveals significant weaknesses in current PRMs: the best-performing model, Gemini-2-Thinking, achieves only 68.8 points, far below the human performance of 83.8.

\begin{figure}[htbp]
    \noindent\makebox[\columnwidth]{\includegraphics[width=\columnwidth]{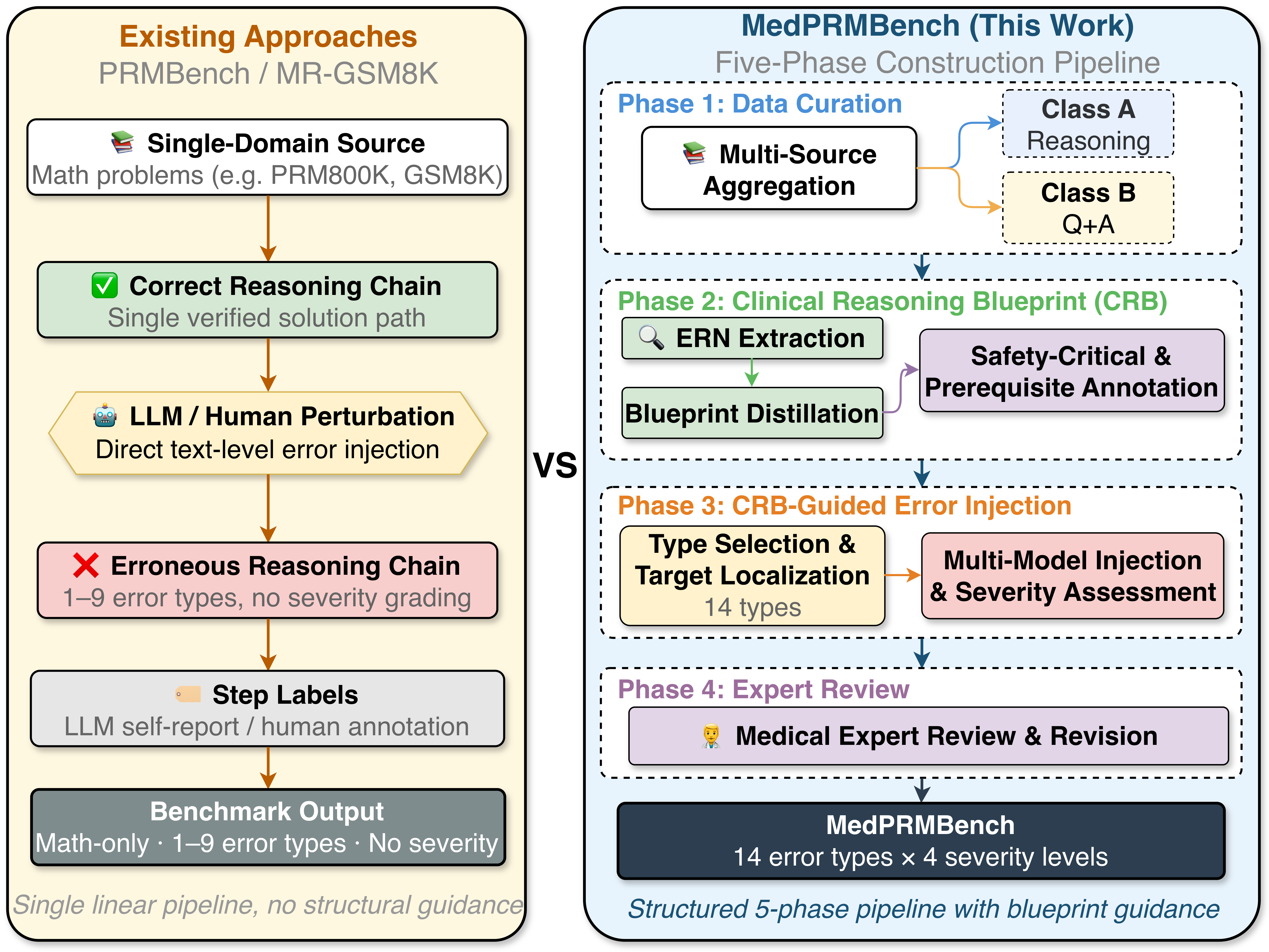}}
    \caption{Comparison of existing benchmark construction approaches (left) and the MedPRMBench three-phase pipeline (right). Existing methods employ a linear perturbation process with limited error categories and no severity grading; MedPRMBench combines structured construction records with deterministic verification and multi-round quality revision, supporting 14 fine-grained error types and 4-level severity annotation.}
    \label{fig:construction_comparison}
\end{figure}

Despite the important progress achieved by the above works in general domains (particularly mathematical reasoning), they do not directly cover medical PRM evaluation. Medical reasoning differs from mathematical reasoning in several important aspects: (1)~medical reasoning is highly dependent on domain expertise (pharmacology, pathology, clinical guidelines, etc.) rather than purely logical rules; (2)~medical reasoning errors can carry \emph{safety-sensitive} implications, such as missing a drug contraindication or allergy check; and (3)~medical reasoning involves integrating multi-source heterogeneous information (patient history, physical examination, laboratory tests, imaging, etc.), resulting in more diverse error types. We therefore retain PRMBench's useful three-category scaffold, clinically instantiate seven inherited labels, re-scope two, and add five types with no PRMBench counterpart---including safety awareness and differential-diagnosis coverage (Table~\ref{tab:error_taxonomy})---alongside a four-level severity grading system.

As illustrated in Figure~\ref{fig:construction_comparison}, existing benchmarks commonly construct erroneous reasoning chains through direct perturbation by LLMs or human annotators. MedPRMBench instead uses a three-phase implementation in which Clinical Reasoning Blueprints (CRBs) provide structured metadata for type-specific candidate construction, followed by deterministic verification and physician review (Section~\ref{sec:method_overview}). This design records target locations, error types, and severity metadata explicitly; without a controlled random-target or direct-perturbation comparison, we do not attribute empirical quality gains specifically to the CRB component.


\section{Additional Evaluation Analyses}
\label{app:additional_main_analyses}

\subsection{Inference Bias Analysis}
\label{sec:inference_bias}

Beyond evaluating overall model performance, we further analyze whether models exhibit systematic bias when judging reasoning steps. We separately compute each model's accuracy on positive-labeled (correct) and negative-labeled (erroneous) steps within erroneous reasoning chains, to reveal bias patterns in medical reasoning verification.

Specifically, for each evaluated model, we partition its step-level predictions on all erroneous reasoning chains in the MedPRMBench test set into two groups by ground-truth label. \textbf{Acc Pos.} denotes the accuracy on all correct steps (i.e., the proportion of correct steps correctly predicted as correct), and \textbf{Acc Neg.} denotes the accuracy on all erroneous steps (i.e., the proportion of erroneous steps correctly predicted as erroneous). \textbf{Bias Gap} is defined as $\text{Acc}_{\text{pos}} - \text{Acc}_{\text{neg}}$; positive values indicate positive bias (tendency to label steps as correct), while negative values indicate negative bias (tendency to label steps as erroneous).

\begin{table}[!t]
\centering
\caption{Comparison of model accuracy on correct and erroneous steps within erroneous reasoning chains on MedPRMBench. \textbf{Acc Pos.} and \textbf{Acc Neg.} denote the step-level accuracy on correct and erroneous steps, respectively. \textbf{PRMScore} is the overall PRMScore from the main evaluation. \textbf{Bias Gap} is Acc Pos. $-$ Acc Neg.; positive values indicate positive bias, negative values indicate negative bias.}
\label{tab:inference_bias}
\footnotesize
\begin{tabularx}{\columnwidth}{@{}X|cc|c|c@{}}
\toprule
\multirow{2}{*}{\textbf{Model}} & \multicolumn{2}{c|}{\textbf{Accuracy}} & \textbf{PRM} & \textbf{Bias} \\
 & \textbf{Pos.} & \textbf{Neg.} & \textbf{Score} & \textbf{Gap} \\
\midrule
\multicolumn{5}{c}{\textit{Frontier API Models}} \\
Claude-Opus-4.5 & 69.1 & 82.5 & 73.7 & $-$13.4 \\
GPT-5.4 & 74.3 & 79.0 & 75.4 & $-$4.7 \\
DeepSeek-R1 & 72.3 & 64.0 & 67.8 & +8.3 \\
\midrule
\multicolumn{5}{c}{\textit{Open-source Reasoning Models}} \\
Sky-T1-32B & 63.0 & 71.0 & 65.4 & $-$8.0 \\
QwQ-32B & 15.9 & 85.7 & 39.0 & $-$69.8 \\
DS-R1-Qwen-32B & 31.4 & 75.0 & 47.4 & $-$43.5 \\
\midrule
\multicolumn{5}{c}{\textit{Open-source Medical Models}} \\
TxGemma-9B & 69.8 & 45.4 & 57.7 & +24.5 \\
HuatuoGPT-o1 & 75.8 & 39.9 & 57.9 & +35.9 \\
Meditron3-8B & 96.4 & 9.7 & 46.8 & +86.7 \\
\midrule
\multicolumn{5}{c}{\textit{Medical Process Reward Models}} \\
Med-PRM-Reward v1.0 & 39.9 & 83.8 & 56.1 & $-$43.9 \\
\rowcolor{gray!15}
Our PRM & 95.2 & 75.7 & 86.6 & +19.5 \\
\bottomrule
\end{tabularx}
\end{table}

\smallskip
\noindent\textbf{Critic models exhibit polarized bias patterns, and the trained PRM retains a moderate correct-step bias.}
As shown in Table~\ref{tab:inference_bias}, critic models exhibit polarized operating points. Claude-Opus-4.5, GPT-5.4, QwQ-32B, and DS-R1-Qwen-32B over-flag errors (negative gaps), while HuatuoGPT-o1 and Meditron3-8B strongly favor correct labels (positive gaps). The released Med-PRM-Reward v1.0 likewise over-flags, with a $-$43.9-point gap. Our one-epoch PRM accepts 95.2\% of correct steps and detects 75.7\% of erroneous steps, yielding a +19.5-point gap. Its 86.6 PRMScore therefore reflects strong balanced-F1 performance in distribution, but not a perfectly balanced operating point; this acceptance tendency becomes more pronounced out of distribution (Section~\ref{sec:transfer}).




\subsection{PRM Baseline as a Plug-and-Play Verifier}
\label{sec:prm_verifier}

Beyond evaluating models' ability to detect reasoning errors, we further investigate whether MedPRMBench-RM (see Appendix~\ref{app:prm_details} for training details) can serve as a \emph{verifier} for downstream medical question answering. We deploy the frozen one-epoch MedPRMBench-RM as a process-level verifier on top of three policy models---Llama-3.1-8B-Instruct, Meerkat-8B, and UltraMedical-8B---and evaluate on MedQA (USMLE, 1{,}273 questions, 4 options) and MedMCQA (4{,}183 questions, 4 options).

\smallskip
\noindent\textbf{Experimental Setup.}
For each policy model, we sample $N{=}64$ candidate solutions using temperature $T{=}0.7$ and top-$p{=}0.9$ with vLLM and a step-by-step chain-of-thought prompt. A trajectory's pre-specified reward is the minimum step-level $P(+)$ across its chain. We compare (1)~\textbf{Self-Consistency (SC)}, majority voting over the 64 final answers; (2)~\textbf{Best-of-N (BoN)}, selecting the single highest-reward trajectory; and (3)~\textbf{weighted-SC}, summing trajectory rewards within each answer group and selecting the group with the largest total. All 1{,}047{,}552 trajectories are rescored with the same frozen checkpoint, with no empty scores, step mismatches, or truncation.

\begin{table}[!htb]
\centering
\caption{Accuracy (\%) of the frozen MedPRMBench-RM as a verifier at $N{=}64$. Parentheses report changes from self-consistency. The trajectory reward is minimum step $P(+)$.}
\label{tab:prm_verifier}
\small
\begin{tabular}{l|cc|c}
\toprule
\textbf{Model / Strategy} & \textbf{MedQA} & \textbf{MedMCQA} & \textbf{Avg.} \\
\midrule
\textbf{Llama-3.1-8B-Instruct} & & & \\
\quad SC                         & 75.10 & 62.59 & 68.84 \\
\quad BoN                        & 73.06 {\scriptsize($-$2.04)} & 61.22 {\scriptsize($-$1.36)} & 67.14 {\scriptsize($-$1.70)} \\
\rowcolor{gray!15}
\quad weighted-SC                & \textbf{76.12} {\scriptsize(+1.02)} & \textbf{62.85} {\scriptsize(+0.26)} & \textbf{69.48} {\scriptsize(+0.64)} \\
\midrule
\textbf{Meerkat-8B} & & & \\
\quad SC                         & 75.81 & 62.71 & 69.26 \\
\quad BoN                        & 70.07 {\scriptsize($-$5.73)} & 60.46 {\scriptsize($-$2.25)} & 65.27 {\scriptsize($-$3.99)} \\
\rowcolor{gray!15}
\quad weighted-SC                & \textbf{76.04} {\scriptsize(+0.24)} & \textbf{63.23} {\scriptsize(+0.53)} & \textbf{69.64} {\scriptsize(+0.38)} \\
\midrule
\textbf{UltraMedical-8B} & & & \\
\quad SC                         & 78.00 & 66.24 & 72.12 \\
\quad BoN                        & 78.79 {\scriptsize(+0.79)} & 65.67 {\scriptsize($-$0.57)} & 72.23 {\scriptsize(+0.11)} \\
\rowcolor{gray!15}
\quad weighted-SC                & \textbf{79.10} {\scriptsize(+1.10)} & \textbf{66.77} {\scriptsize(+0.53)} & \textbf{72.94} {\scriptsize(+0.81)} \\
\bottomrule
\end{tabular}
\end{table}

The results are presented in Table~\ref{tab:prm_verifier}. We summarize the key findings as follows.

\smallskip
\noindent\textbf{Reward weighting provides a small, policy-dependent signal on top of consensus.}
Across all six policy--dataset settings, weighted-SC improves over SC by 0.24--1.10 points (descriptive micro-average +0.52). After Holm correction across the six comparisons, only UltraMedical on MedQA remains significant ($+1.10$ points, adjusted $p{=}0.0397$); the other adjusted $p$-values range from 0.073 to 0.648. These results support a modest consensus-conditioned benefit, rather than a large or uniformly significant downstream gain.

\smallskip
\noindent\textbf{The PRM is not a reliable standalone best-of-$N$ ranker.}
BoN underperforms SC in five of six settings, including significant corrected declines for Meerkat on both datasets, and is 1.61 points worse in the descriptive micro-average. In contrast, the $N$-scaling analysis shows weighted-SC remaining slightly above SC from $N\geq2$. The useful signal is therefore conditional on answer-level consensus; step rewards alone do not reliably rank complete trajectories.


\subsection{Ablation Study}
\label{sec:ablation}

To assess which training-data components contribute most, we run two controlled ablations. All models use Qwen3-8B, the same 6{,}870 question IDs and 13{,}740-chain budget, and the same pre-specified one-epoch recipe; only the indicated data component changes:

\begin{itemize}[leftmargin=1.5em, itemsep=2pt]
    \item \textbf{w/o Error Construction}: Removes the candidate error construction phase; training data contains only correct reasoning chains (all steps labeled as correct), providing no negative samples.
    \item \textbf{w/o Expert Review}: Removes the expert review phase, using automatically generated data without human verification for training.
\end{itemize}

\begin{table}[!t]
\centering
\caption{Controlled one-epoch data ablations. Full and no-review results are mean $\pm$ sample SD over seeds 42, 1234, and 2026; parentheses report the mean paired change from Full. The all-positive no-error-construction sanity control ($\dagger$) was run with seed 42, so its parenthesized changes are relative to the seed-42 Full model (86.6/75.7/95.2/89.0/2.0). PRMScore, error recall, correct-step accuracy, and strict first-error localization are measured on the 6{,}500 corrupted chains; clean false alarm is the percentage of 6{,}500 clean chains with at least one predicted error.}
\label{tab:ablation}
\resizebox{\columnwidth}{!}{
\begin{tabular}{l|ccccc}
\toprule
\textbf{Configuration} & \textbf{PRMScore} & \textbf{Error recall} & \textbf{Correct acc.} & \textbf{Strict first} & \textbf{Clean FA} \\
\midrule
Full Model (Ours) & $86.5{\pm}0.4$ & $76.7{\pm}2.6$ & $94.3{\pm}1.3$ & $88.3{\pm}1.0$ & $2.1{\pm}0.6$ \\
w/o Expert Review & $84.8{\pm}0.1$ ($-$1.7) & $70.1{\pm}0.5$ ($-$6.6) & $96.3{\pm}0.2$ (+2.0) & $88.5{\pm}0.2$ (+0.2) & $5.6{\pm}0.5$ (+3.5) \\
w/o Error Construction$^{\dagger}$ & 38.5 ($-$48.1) & 0.0 ($-$75.7) & 100.0 (+4.8) & 0.0 ($-$89.0) & 0.0 ($-$2.0) \\
\bottomrule
\end{tabular}
}
\end{table}

As shown in Table~\ref{tab:ablation}, constructed negative chains provide the largest necessary learning signal: without them, the model sees no negative labels, converges to an all-positive classifier, and loses 48.1 PRMScore points in the seed-42 control (95\% CI: $-$48.5 to $-$47.8) while error recall and strict localization fall to zero. This arm is therefore a class-exposure sanity control, not evidence that every part of the construction pipeline contributes 48 points. Physician-reviewed labels provide the more informative quality comparison. Across three matched seeds, removing review lowers PRMScore by $1.72{\pm}0.33$ points (range: $-$2.02 to $-$1.36), lowers error recall by $6.55{\pm}2.45$ points, and raises the clean-chain false-alarm rate by $3.54{\pm}0.91$ points. The ranking difference is also consistent without choosing a threshold: removing review changes corrupted-chain error AUROC by $-0.82{\pm}0.18$ points and error-AUPRC by $-0.81{\pm}0.31$ points. The no-review arm is more conservative on correct steps within corrupted chains (+$2.00{\pm}1.35$ points) but misses substantially more errors, while strict first-error localization has no stable paired change ($+0.24{\pm}1.15$ points).


\section{Additional Robustness and Validity Analyses}
\label{app:robustness}

This appendix collects secondary analyses supporting Section~\ref{sec:generalization} and the validity of the main evaluation protocol: cross-backbone replication, code-attributed diagnostics, parser coverage, Class~A/B stratification, repeated-sampling evaluation, construction-weight sensitivity, step-local and question-replacement probes, and annotation reliability.

\subsection{Cross-Backbone Replication Details}
\label{app:backbone_replication}

\begin{table*}[!htb]
\centering
\caption{Frozen-protocol scale and cross-family replication (\%). MedPRMBench metrics use 6{,}500 corrupted chains except Clean FA, the fraction of 6{,}500 clean chains with any false alarm. Strict first requires the model's earliest predicted error to equal the gold first error. PB is ProcessBench macro F1; PRMB is official PRMBench PRMScore. Each row is the sole epoch-one checkpoint from seed 42.}
\label{tab:backbone_replication}
\small
\resizebox{0.96\textwidth}{!}{
\begin{tabular}{lccccccc}
\toprule
\textbf{Backbone} & \textbf{MedPRMScore} & \textbf{Error recall} & \textbf{Correct acc.} & \textbf{Strict first} & \textbf{Clean FA}$\downarrow$ & \textbf{PB} & \textbf{PRMB} \\
\midrule
Qwen3-8B & 86.6 & 75.7 & 95.2 & 89.0 & 2.0 & 55.3 & 64.3 \\
Qwen3-4B & 86.4 & 83.0 & 89.8 & 87.4 & 6.0 & 50.7 & 64.4 \\
Llama-3.1-8B-Instruct & 87.2 & 81.1 & 92.3 & 88.7 & 2.5 & 25.2 & 56.1 \\
\bottomrule
\end{tabular}}
\end{table*}

All runs use the same 6{,}870-question training split, learning rate, effective batch size, prompt, boundary-token objective, and one pre-specified epoch. No checkpoint, threshold, or architecture was selected using test performance, and all cover 13{,}000 MedPRMBench chains and 113{,}910 steps without truncation. Qwen3-4B trades higher error recall for lower correct-step accuracy and more clean-chain false alarms. The Llama result confirms cross-family in-domain replication while showing that external transfer is not architecture invariant.

\subsection{Code-Attributed Error Recall}
\label{app:attributed_diagnostics}

The taxonomy columns in Table~\ref{tab:error_detection_by_type_en} are overlapping chain slices: if one chain contains R-1 and E-1 errors, every step in that chain contributes to both columns. To isolate detection of the named phenomenon, Table~\ref{tab:attributed_recall} instead evaluates only physician-finalized step--code pairs. A step assigned multiple codes contributes once to each applicable code.

\begin{table*}[!htb]
\centering
\caption{Code-attributed error-step recall (\%) of the canonical one-epoch PRM. $n$ is the number of finalized error-step assignments for each code.}
\label{tab:attributed_recall}
\small
\resizebox{\textwidth}{!}{%
\begin{tabular}{l|cc|ccccccc|ccccc}
\toprule
 & \textbf{S-1} & \textbf{S-2} & \textbf{R-1} & \textbf{R-2} & \textbf{R-3} & \textbf{R-4} & \textbf{R-5} & \textbf{R-6} & \textbf{R-7} & \textbf{E-1} & \textbf{E-2} & \textbf{E-3} & \textbf{E-4} & \textbf{E-5} \\
\midrule
$n$ & 2{,}880 & 1{,}044 & 7{,}248 & 1{,}764 & 1{,}264 & 1{,}411 & 1{,}900 & 3{,}888 & 1{,}882 & 1{,}124 & 1{,}034 & 691 & 904 & 2{,}135 \\
Recall & 94.6 & 91.5 & 67.2 & 71.3 & 81.0 & 86.3 & 68.8 & 86.4 & 73.6 & 53.4 & 95.5 & 96.8 & 97.5 & 84.9 \\
\bottomrule
\end{tabular}}
\end{table*}

Prerequisite Sensitivity (E-1) is the clearest weakness (53.4\%), followed by Evidence-Based Soundness (R-1, 67.2\%) and Safety Awareness (R-5, 68.8\%). Quantitative Correctness (E-4), Multi-Solution Consistency (E-3), and Deception Resistance (E-2) exceed 95\% recall. These descriptive differences motivated the matched leave-one-type-out experiment in
Table~\ref{tab:type_loto}.  For each target, the LOTO arm removes every training row
containing that code, while the control removes the same number of rows matched within
source, severity, chain length, and error-count strata.  Because rows can carry multiple
codes, the estimand is the contribution of rows \emph{containing} the target rather than
an isolated single-code intervention.

\begin{table*}[!t]
\centering
\caption{Matched leave-one-type-out training ablations (percentage points). $\Delta$ is matched control minus LOTO; positive values mean that retaining target-bearing rows improves the metric. Rows/arm is the training-set size after removing all target-bearing rows from the 6{,}870-question split (e.g., $6{,}870-1{,}487=5{,}383$ for R-2); Target rows kept is the number of target-bearing rows that the size-matched control retains. Brackets are paired question-bootstrap 95\% confidence intervals. Target recall uses physician-finalized step--code assignments; Clean FA is lower-is-better, so a positive $\Delta$ indicates more false alarms in the control.}
\label{tab:type_loto}
\scriptsize
\resizebox{\textwidth}{!}{%
\begin{tabular}{lrrrrrrr}
\toprule
\textbf{Target} & \textbf{Rows/arm} & \textbf{Target rows kept} & \textbf{Control PRM} & \textbf{LOTO PRM} & \textbf{$\Delta$ PRM [95\% CI]} & \textbf{$\Delta$ target recall [95\% CI]} & \textbf{$\Delta$ Clean FA [95\% CI]} \\
\midrule
E-1 & 5,763 & 1,098 & 86.5 & 86.5 & -0.01 [-0.15, +0.14] & +7.38 [+5.73, +9.18] & +1.29 [+0.98, +1.62] \\
R-1 & 2,608 & 2,363 & 85.2 & 85.0 & +0.16 [-0.07, +0.39] & +10.09 [+9.38, +10.85] & +1.32 [+0.91, +1.72] \\
R-5 & 5,450 & 1,416 & 86.1 & 85.5 & +0.57 [+0.31, +0.84] & +20.21 [+18.51, +21.90] & +1.57 [+1.25, +1.92] \\
R-2 & 5,383 & 1,466 & 86.3 & 86.6 & -0.29 [-0.45, -0.14] & -2.61 [-3.77, -1.49] & +0.48 [+0.18, +0.77] \\
R-7 & 5,556 & 1,307 & 85.8 & 86.2 & -0.37 [-0.56, -0.19] & +6.11 [+4.94, +7.34] & +0.35 [+0.05, +0.68] \\
\bottomrule
\end{tabular}}
\end{table*}

The paired design controls training-set size and the principal observed strata, but it
does not make multi-label rows single-type interventions.  We therefore interpret the
confidence intervals as matched contributions of target-bearing data and retain the
full operating-point trade-offs rather than ranking error types by PRMScore alone.

\paragraph{Operating-point diagnostic.}
The frozen $P(+){\geq}0.5$ operating point remains the primary protocol.  As a
uniform post-hoc diagnostic, Table~\ref{tab:type_operating_point} additionally reports
error-detection AUROC over all labeled steps in corrupted chains carrying each target
code.  This threshold-free quantity measures ranking discrimination on an overlapping
target-chain slice; it is not code-specific AUROC because those chains can contain
other error codes. 

\begin{table*}[!t]
\centering
\caption{Post-hoc operating-point diagnostic for the five matched type ablations (percentage points). $\Delta$ is matched control minus LOTO. Fixed-threshold target recall uses physician-finalized step--code assignments; target-chain AUROC uses all labeled steps in corrupted chains carrying the target. Brackets are paired question-bootstrap 95\% confidence intervals (2,000 draws).}
\label{tab:type_operating_point}
\scriptsize
\resizebox{\textwidth}{!}{%
\begin{tabular}{lrrrrr}
\toprule
\textbf{Target} & \textbf{Test target chains} & \textbf{$\Delta$ recall at 0.5 [95\% CI]} & \textbf{Control AUROC} & \textbf{LOTO AUROC} & \textbf{$\Delta$ AUROC [95\% CI]} \\
\midrule
E-1 & 998 & +7.38 [+5.73, +9.18] & 94.05 & 93.49 & +0.55 [+0.43, +0.67] \\
R-1 & 4,020 & +10.09 [+9.38, +10.85] & 93.42 & 93.39 & +0.03 [-0.04, +0.11] \\
R-5 & 1,483 & +20.21 [+18.51, +21.90] & 94.68 & 94.49 & +0.20 [+0.10, +0.30] \\
R-2 & 1,481 & -2.61 [-3.77, -1.49] & 94.36 & 94.35 & +0.01 [-0.08, +0.10] \\
R-7 & 1,306 & +6.11 [+4.94, +7.34] & 94.43 & 94.32 & +0.11 [+0.03, +0.20] \\
\bottomrule
\end{tabular}}
\end{table*}

R-2 is the only target with a negative fixed-threshold recall contrast
($-2.61$ points), while its target-chain AUROC contrast is effectively zero
($+0.01$ points; 95\% CI $[-0.08,+0.10]$).  Moreover, only 9 of the 1,487
R-2-bearing rows in the original training set are R-2-only; the rest are
multi-label.  Thus the experiment cannot identify R-2 supervision as harmful or
separate calibration from interactions with co-occurring labels.

\subsection{Same-Backbone Qwen3-8B Diagnostic}
\label{app:matched_base_control}

We evaluate the untrained Qwen3-8B initialization (the post-trained Qwen3-8B checkpoint from which our PRM is fine-tuned) on the complete PRMBench official task
with the trained model's chat template, step delimiter, \texttt{+}/\texttt{-}
verbalizers, and 0.5 decision rule. Eight GPU shards cover all 6{,}969 records
with no duplicate, missing, or step-mismatched outputs. As in the official task
score, the 758 synthesized correct-reference records are excluded from the
6{,}211-record, 83{,}371-step comparison. A second base run uses a concise
system instruction while holding the remaining interface fixed.

\begin{table}[!t]
\centering
\caption{Same-backbone PRMBench diagnostic on the official non-reference pool.
The base model has no native PRM head; its fixed-threshold scores are
interface-dependent. AUROC uses the continuous step correctness score.}
\label{tab:matched_base_control}
\small
\begin{tabular}{lcccc}
\toprule
\textbf{Qwen3-8B} & \textbf{PRMScore} & \textbf{Error rec.} & \textbf{Correct acc.} & \textbf{AUROC} \\
\midrule
Untrained, matched prompt & 23.4 & 97.5 & 10.0 & 65.9 \\
Untrained, concise prompt & 29.6 & 92.1 & 17.9 & 64.7 \\
MedPRMBench trained & 64.3 & 24.7 & 97.8 & 78.6 \\
\bottomrule
\end{tabular}
\end{table}

The trained model's AUROC exceeds the matched-prompt base by
12.71 points (95\% paired chain-cluster bootstrap CI: 12.07--13.47; 300 draws)
and the concise-prompt base by 13.90 points (13.27--14.61).
This measures improved ranking under the shared verbalizer interface. The
large shift in the base model's thresholded score across the two prompts makes
its 23.4/29.6 PRMScore unsuitable as a native zero-shot PRM baseline. These
comparisons establish the effect of training the backbone for this interface;
they do not isolate the contribution of medical data from supervised PRM
training in general.

At a comparable correct-step operating point, the recall difference is easier
to interpret. We partition the 6{,}211 official non-reference records into
five folds by record ID. For each held-out fold, a base-model threshold is
chosen on the other four folds to match the trained model's correct-step
acceptance at its fixed 0.5 threshold. On held-out records, correct-step
acceptance is 97.9\% for either base prompt and 97.8\% for the trained model;
error recall is 7.6\% (matched prompt), 8.5\% (concise prompt), and 24.7\%
(trained). The trained-minus-base recall differences are $+17.0$ points
(95\% record-cluster bootstrap CI: $+16.2$ to $+18.0$) and $+16.2$ points
($+15.4$ to $+16.9$), respectively. This post-hoc, cross-fitted comparison
uses the common prompted verbalizer interface and is separate from the
official fixed-threshold PRMScore. It shows improved error sensitivity at a
similar false-alarm operating point without attributing the entire gain
specifically to medical data.

On ProcessBench, under the same interface and 0.5 rule, the untrained
initialization reaches 2.0 (matched prompt) and 10.8 (concise prompt) macro F1,
versus 55.3 after training. The pre-trained-only Qwen3-8B-Base behaves
similarly: 1.9/4.8 ProcessBench macro F1 and 47.4/49.6 PRMBench PRMScore, as it
accepts nearly all steps. Table~\ref{tab:transfer} reports the concise-prompt
row of the actual initialization.

\subsection{Native-Interface Reverse Transfer}
\label{app:reverse_transfer}

The trained model's PRMBench category breakdown is weakest for redundancy
(52.8), circularity (55.6), and missing-condition errors (56.9). These are
diagnostic associations, not error-type-specific training effects.

Table~\ref{tab:reverse_transfer} applies the native scoring interface for each released
PRM and conditions medical interpretation on a ProcessBench reproduction control.
Three of six models pass that gate; failed controls remain visible but their medical
scores are intentionally suppressed. This guards against attributing interface or
tokenization failures to domain transfer.

For Table~\ref{tab:transfer}, we fill cells unreported by the benchmark papers with
the same harnesses. Before scoring the missing benchmark, each model must reproduce its
published value on the other: PURE-PRM-7B reaches 65.3 PRMBench PRMScore (published
65.3) and then 56.7 ProcessBench macro F1; ReasonEval-7B reaches 60.0 (published 60.1)
and then 35.3. Qwen2.5-Math-7B-PRM800K reaches 56.4 ProcessBench macro F1 (published
56.5) and 58.3 PRMBench PRMScore. For the bootstrap below, we also rerun Skywork-PRM-7B
on PRMBench, which reaches 65.1 (published 65.1). All thresholds follow each model's native
decision rule without test-set tuning.

To test the close comparisons in Table~\ref{tab:transfer}, we run a paired bootstrap
(10{,}000 draws) over the per-record outputs of MedPRM and the five neighbors with
available records, resampling within ProcessBench subsets and within PRMBench
categories. On ProcessBench, MedPRM minus PURE-PRM-7B is $-1.5$ (95\% CI $-4.1$ to
$+1.2$) and minus Qwen2.5-Math-7B-PRM800K is $-1.1$ ($-3.8$ to $+1.5$), so both
differences are within resampling noise. On PRMBench, MedPRM trails Skywork-PRM-7B by 0.8
(0.3 to 1.4), PURE-PRM-7B by 1.1 (0.5 to 1.7), and Qwen2.5-Math-PRM-7B by 1.2 (0.7 to 1.8),
and exceeds Qwen2.5-Math-7B-PRM800K by 6.0 (5.4 to 6.5) and ReasonEval-7B by 4.2 (3.6 to
4.8). Under Holm correction over all nine comparisons (five on PRMBench, four on
ProcessBench), the five PRMBench gaps remain significant (adjusted $p<0.01$) and both
ProcessBench ties remain non-significant. These intervals reflect test-set sampling for fixed checkpoints, not training-seed or
scoring-interface variation.

\begin{table*}[!t]
\centering
\caption{Native-interface reverse transfer from mathematical/general PRMs. PB is ProcessBench macro F1. A model passes the pre-specified positive control only if every ProcessBench subset reaches at least 90\% of its published F1; MedPRMBench values are suppressed when that gate fails. All shown medical runs use the native threshold 0.5 and exact 13,000-chain/113,910-step coverage.}
\label{tab:reverse_transfer}
\small
\resizebox{0.92\textwidth}{!}{%
\begin{tabular}{lccccc}
\toprule
\textbf{Native PRM} & \textbf{PB macro F1} & \textbf{Control gate} & \textbf{MedPRMScore} & \textbf{Error recall} & \textbf{Correct acc.} \\
\midrule
Qwen2.5-Math-PRM-7B & 73.1 & pass & 62.7 & 31.7 & 94.5 \\
Qwen2.5-Math-7B-PRM800K & 56.4 & pass & 43.3 & 4.7 & 99.7 \\
Math-Shepherd-PRM-7B & 28.6 & fail & -- & -- & -- \\
RLHFlow-Llama3.1-8B-PRM & 28.6 & pass & 42.5 & 5.5 & 94.7 \\
Skywork-o1-Open-PRM-7B & 39.5 & fail & -- & -- & -- \\
Universal-PRM-7B & 0.0 & fail & -- & -- & -- \\
\bottomrule
\end{tabular}}
\end{table*}

\subsection{Parsed-Output Coverage}
\label{app:parse_coverage}

We evaluate six open generative critics with an exact single-label-per-step output
protocol.  All six runs cover all 6,500 corrupted chains and 59,174 corrupted-chain
steps with zero parse failures; main and detailed tables use these full-coverage results.

\begin{table}[!htb]
\centering
\caption{Exact-output chain coverage for the six rerun open generative critics.}
\label{tab:parse_coverage}
\small
\begin{tabular}{lrr}
\toprule
\textbf{Model} & \textbf{Parsed / 6,500} & \textbf{Coverage} \\
\midrule
HuatuoGPT-o1-8B & 6{,}500 / 6{,}500 & 100.00\% \\
llama-3-meerkat-8b & 6{,}500 / 6{,}500 & 100.00\% \\
Qwen3-8B & 6{,}500 / 6{,}500 & 100.00\% \\
DeepSeek-R1-Distill-Qwen-32B & 6{,}500 / 6{,}500 & 100.00\% \\
QwQ-32B & 6{,}500 / 6{,}500 & 100.00\% \\
DeepSeek-R1-Distill-Llama-70B & 6{,}500 / 6{,}500 & 100.00\% \\
\bottomrule
\end{tabular}
\end{table}

\subsection{Class A/B Composition and Stratified Performance}
\label{app:classab}

A natural concern is that Class~B instances---whose correct chains are themselves LLM-generated before error injection---might reduce the task to ``distinguish uncorrupted from corrupted LLM text,'' an easier and different problem than detecting errors in expert-authored reasoning. We report the exact composition and the stratified effect. In the effective model-training set (6{,}870 instances after overlap removal), Class~A (expert-authored, MedCaseReasoning) accounts for 47.2\% (3{,}244) and Class~B for 52.8\% (3{,}626); on the evaluation split (6{,}500 instances), Class~A accounts for 57.1\% (3{,}711) and Class~B for 42.9\% (2{,}789). The two splits differ in source-dataset mix (MMLU-medical and PubMedQA appear only on the evaluation side), so we report both rather than a pooled ratio.

\begin{table}[!htb]
\centering
\caption{Error-step recall stratified by reasoning-chain provenance using the full-coverage outputs. The benchmark-specific PRM and four of six exact-output critics find Class~B harder; two aggressive error-flagging critics show a modest reversal.}
\label{tab:classab}
\small
\begin{tabular}{lccc}
\toprule
\textbf{Model} & \textbf{Class A recall} & \textbf{Class B recall} & \textbf{Gap (A$-$B)} \\
\midrule
Our PRM                        & 81.29 & 69.07 & $+$12.22 \\
DeepSeek-R1-Distill-Llama-70B  & 47.29 & 30.00 & $+$17.29 \\
DeepSeek-R1-Distill-Qwen-32B   & 73.07 & 77.23 & $-$4.16 \\
HuatuoGPT-o1-8B                & 43.61 & 35.42 & $+$8.20 \\
llama-3-meerkat-8b             & 39.99 & 35.87 & $+$4.12 \\
QwQ-32B                        & 83.85 & 87.90 & $-$4.04 \\
Qwen3-8B                       & 65.53 & 54.29 & $+$11.24 \\
\bottomrule
\end{tabular}
\end{table}

The benchmark-specific PRM detects substantially fewer Class~B errors (69.07\%) than Class~A errors (81.29\%), opposing the hypothesis that its high score arises because corrupted LLM-generated chains are uniformly easier to recognize. Four of six full-coverage generative critics show the same direction. QwQ-32B and DS-R1-Qwen-32B reverse it by about four points while operating at unusually high overall error-flagging rates, so the provenance effect is not universal across critics. These results rule out a simple ``Class~B is easier'' account for the trained PRM.

\subsection{API Evaluation Protocol Robustness}
\label{app:api_protocol}

PRM-style models are scored from calibrated $P(+)/P(-)$ token probabilities at a fixed 0.5 threshold, while frontier API models are scored from generated labels---these APIs do not consistently expose per-step token probabilities, so the two groups are measured with different instruments. To check whether the one-shot generative protocol underestimates frontier models, we reran an 800-question stratified subset ($\sim$57 questions per error type) with repeated API sampling: each model outputs a JSON validity score per step, we sample $k{=}5$ independent completions (temperature 0.7), and compare three aggregations---first sample only, mean validity score, and binary majority vote (Table~\ref{tab:api_k5}).

\begin{table}[!htb]
\centering
\caption{Repeated-sampling aggregation for API critics on an 800-question stratified subset. Aggregation moves PRMScore by at most $+0.87$ points. Absolute scores differ from Table~\ref{tab:error_detection_by_type_en} because of the protocol and subset difference.}
\label{tab:api_k5}
\small
\begin{tabular}{lccc}
\toprule
\textbf{Model} & \textbf{Single-sample} & \textbf{$k{=}5$ mean} & \textbf{$k{=}5$ majority} \\
\midrule
GPT-5.4         & 80.59 & 80.53 & 80.59 \\
GPT-5.2         & 79.07 & 79.94 & 79.48 \\
Claude-Opus-4.5 & 79.44 & 79.34 & 79.48 \\
GLM-4.7         & 79.91 & 80.11 & 80.11 \\
\bottomrule
\end{tabular}
\end{table}

Repeated sampling moves frontier-model PRMScore by at most $+0.87$ points (GPT-5.2, single $\to$ mean); error-step detection shows the same pattern, with $k{=}5$ aggregation shifting the recall/precision tradeoff only modestly. Coverage is near-complete for GPT-5.4, GPT-5.2, and Claude-Opus-4.5; for GLM-4.7, 1{,}587 of 1{,}600 question/chain groups have valid data (1{,}369 with full $k{=}5$). Accordingly, the frontier-model rows in the main tables should be read as \emph{prompted-critic baselines under the same step-verification task}, not as calibrated PRMs.

\subsection{Auxiliary Frontier-Model Repeated-Sampling Check}
\label{app:variance}

Using the same 800-question stratified subset and protocol as Appendix~\ref{app:api_protocol} (so the two analyses are directly comparable), we compute PRMScore separately for each of the $k{=}5$ replicates (Table~\ref{tab:variance}).

\begin{table}[!htb]
\centering
\caption{Auxiliary repeated-sampling check for four API critics on an 800-question subset under the JSON validity-score protocol (temperature 0.7): five independent replicates per model. This is not a variance estimate for the full-set main-table protocol.}
\label{tab:variance}
\small
\begin{tabular}{lcccc}
\toprule
\textbf{Model} & \textbf{Replicate scores} & \textbf{Mean} & \textbf{Std} & \textbf{Range} \\
\midrule
GPT-5.4         & 80.59, 80.32, 80.61, 80.59, 80.53 & 80.53 & 0.12 & 0.29 \\
GPT-5.2         & 79.07, 79.05, 79.04, 79.43, 79.18 & 79.15 & 0.16 & 0.39 \\
Claude-Opus-4.5 & 79.44, 79.55, 79.46, 79.46, 79.31 & 79.44 & 0.09 & 0.24 \\
GLM-4.7         & 80.51, 80.39, 80.21, 80.58, 80.58 & 80.45 & 0.16 & 0.37 \\
\bottomrule
\end{tabular}
\end{table}

Across this auxiliary protocol, replicate ranges are 0.24--0.39pp. Because both the subset and decoding interface differ from the full-set main evaluation, we use these numbers only to characterize local sampling stability under this protocol; they do not statistically resolve the close ordering of frontier models in Table~\ref{tab:error_detection_by_type_en}. Coverage is $\ge$7{,}990 of 8{,}000 parsed requests per model for the first three; GLM-4.7 replicates each cover 1{,}487--1{,}515 of the 1{,}600 question/chain groups (93--95\%) due to endpoint timeouts on long clinical prompts.

\subsection{Logged Construction-Weight Sensitivity}
\label{app:threshold_sens}

The construction pipeline logs per-component scores separately before combining them, which allows perturbing the manually chosen weights without regenerating data. \textbf{Node-criticality weights} ($\alpha{=}0.4$, $\beta{=}0.35$, $\gamma{=}0.25$; Appendix~\ref{app:bnc}): perturbing each weight by $\pm$0.1 (renormalized) across 2{,}997 blueprint instances and measuring how often the top-ranked implementation target changes gives: $\alpha\pm0.1$: top-1 changed in 1.5--2.1\% of instances, top-3 Jaccard overlap 0.96; $\beta\pm0.1$: 2.6--14.0\%, 0.81--0.93; $\gamma\pm0.1$: 2.8--12.4\%, 0.83--0.92. Thus this heuristic is stable under $\alpha$ and moderately sensitive to $\beta$/$\gamma$; these are implementation-sensitivity statistics, not evidence that the targeting heuristic is superior to alternative construction strategies. \textbf{Severity-score weights} ($\alpha_1{=}\alpha_2{=}0.35$, $\alpha_3{=}0.30$; Appendix~\ref{app:severity_assessment}), on 3{,}000 corrupted variants: $w_1$ (step disruption) $\pm$0.1 shifts the continuous severity score by 0.046 on average (42--44\% of instances shift by $>$0.05); $w_2$ (target safety level): 0.024 (7.7--8.1\%); $w_3$ (error-type weight): 0.013 (0\%). The reported \emph{severity level} (Critical/Major/Moderate/Minor---the label behind the paper's headline severity statistics) is assigned by a separate discretization rule on step disruption and target safety level, not on the weighted severity score, so the paper's severity distribution is invariant to the 0.35/0.35/0.30 weighting by construction; only the continuous score used for secondary ranking (e.g., Hard Subset selection and best-variant choice) is affected. These analyses cover only the logged node-criticality and severity weights. The semantic-matching thresholds $\delta_e$ and $\delta_r$ were fixed implementation settings during construction.

\subsection{Step-Local Signal Probe}
\label{app:shortcut_probe}

To quantify how much label information is available in individual steps, we train a TF--IDF classifier that sees only an isolated step---not the question, answer choices, neighboring steps, error type, or severity. Correct and erroneous steps are matched within construction source, chain length, step position, exact word count, and a 20-character-length bin. On the resulting 7{,}778-step cohort, the probe reaches 67.6 PRMScore (95\% question-cluster bootstrap CI: 66.6--68.7) and 75.2 AUROC. This measures step-local predictability on a matched, label-selected cohort; it does not identify the features used by the trained PRM. The question-replacement analysis below separately measures its input sensitivity.

\subsection{Question-Replacement Sensitivity}
\label{app:context_counterfactual}

To probe how strongly the frozen PRM depends on question context, we replace the question and options for every benchmark item with a donor from the same source dataset and option-count stratum, matched by adjacent word length, while leaving both clean and corrupted reasoning chains unchanged. The pre-specified primary gate accepts 6{,}488/6{,}500 questions (absolute word-count difference at most five words or 10\%); all 13{,}000 chains are nevertheless scored and pass integrity checks. Across the 113{,}610 primary-cohort steps, original and swapped scores have Pearson $r{=}0.939$, mean absolute movement 0.0385, and 96.68\% threshold-label agreement. The question-cluster bootstrap mean absolute shift is 0.0392 (95\% CI: 0.0384--0.0401). On corrupted chains alone, the corresponding values are $r{=}0.949$, 0.0502, and 95.44\%. Separation under the original labels decreases from 0.978 to 0.964 AUROC overall (0.952 to 0.934 on corrupted chains), but this is only a separation diagnostic because the donor question makes those labels semantically invalid.

The paired movement quantifies response to unrelated question replacements. Because a donor question can make the original reasoning chain clinically incoherent, we report score and label movement without treating the donor question as a new correctness label. Code-attributed error recall on the original, physician-reviewed cases provides the complementary label-valid analysis.

In a post-hoc code-attributed slice of the same 6{,}488-question cohort, we
compare original and swapped decisions only at physician-finalized erroneous
steps. Agreement is 96.59\% on 1{,}261 R-3 steps from 1{,}094 questions
(question-cluster bootstrap 95\% CI: 95.64--97.48\%) and 92.69\% on 1{,}888
R-5 steps from 1{,}476 questions (91.50--93.87\%). Mean absolute score
movement is 0.0420 and 0.0522, respectively. Codes can overlap on a step;
these are descriptive sensitivity slices, not accuracy estimates under the
donor questions.

\subsection{Physician Annotation and Inter-Annotator Agreement}
\label{app:annotation_reliability}

\textbf{Review design and IAA stage.} Physicians judge both the original reasoning and the system-generated error annotation, correcting the reasoning or error step--type mapping when needed. A shared subset of 1{,}000 MedCaseReasoning (Class~A) candidates was independently and blindly annotated twice, with a different primary physician assigned to every paired candidate; the question, original chain, corrupted chain, and system-proposed labels are identical across the two assignments. IAA compares these two initial-annotation responses only. The subsequent second-stage check and model-consistency filter determine the released labels and do not enter the IAA calculation.

\textbf{Reviewers.} All 86 annotators and checkers are practicing physicians from Grade-A tertiary hospitals in China, the highest tier of the national hospital accreditation system. Each candidate is reviewed by two physicians in sequence: an annotator gives the initial judgment and corrections, and a checker verifies and, when needed, corrects them. No third physician adjudicates; released labels are the checked labels after the model-consistency filter.

\textbf{Initial-annotation observed agreement.} Each physician's proposed binary error-step labels are derived from the independent initial response: whenever a physician supplies a corrected step--type map, that map is used; otherwise, the system map is retained when the physician accepts it as accurate. Across the 5{,}763 steps in the 1{,}000-candidate shared subset, the two independent annotations agree on 84.71\% of binary error-step labels (candidate-cluster bootstrap 95\% CI: 83.65--85.77\%).

\textbf{Chance-corrected agreement.} Computed on the same two independent initial step-label vectors, Cohen's $\kappa=0.691$ (candidate-cluster bootstrap 95\% CI: 0.669--0.713), indicating substantial chance-corrected agreement. Second-stage-checked labels, rather than either individual initial response, are used in the released benchmark.


\section{Illustration of 14 Error Types}
\label{app:error_types}

\begin{figure}[!htbp]
    \centering
    \includegraphics[width=0.98\columnwidth]{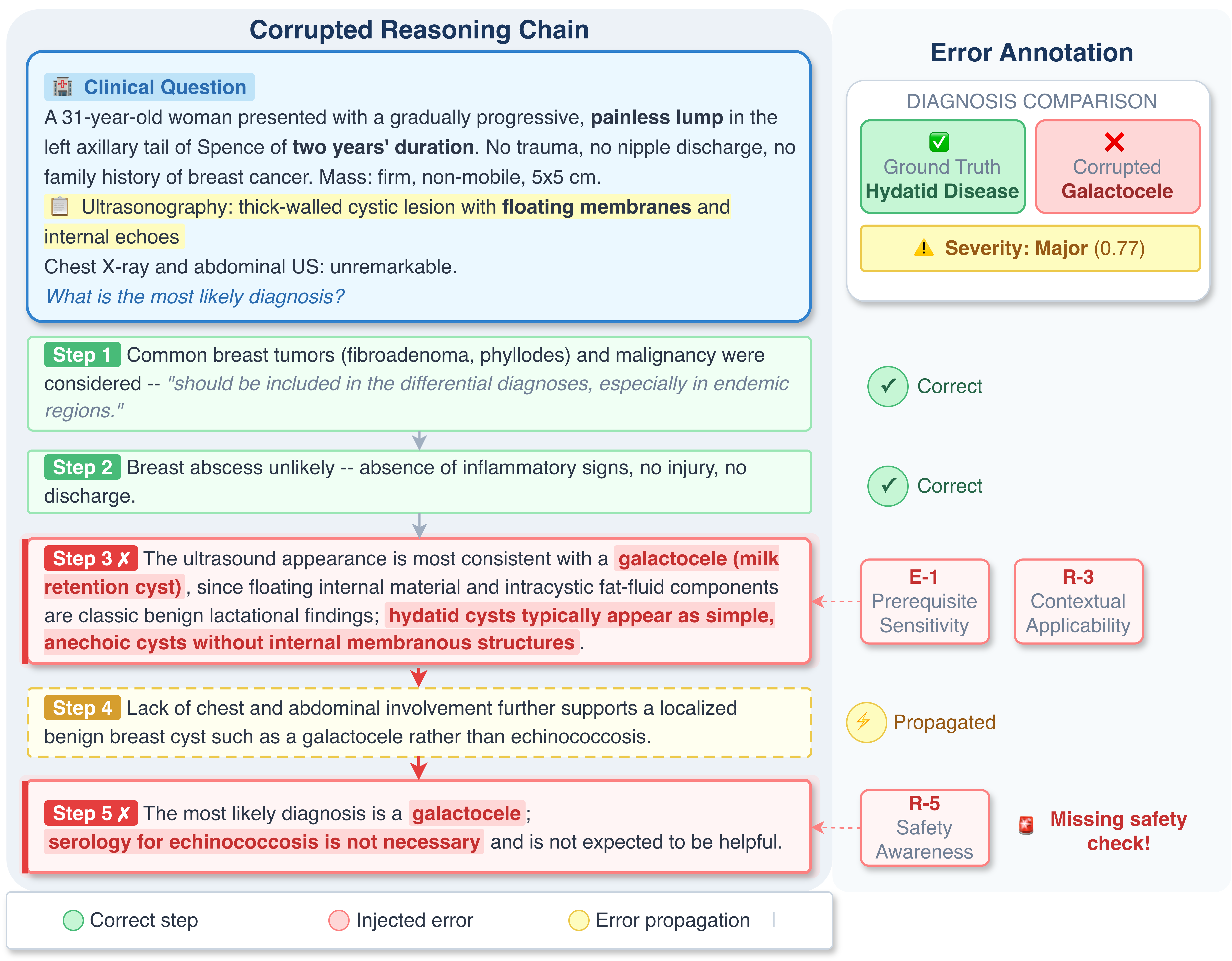}
    \caption{Full MedPRMBench example: the corrupted reasoning chain includes errors at Steps~3 and~5, annotated with Prerequisite Sensitivity (E-1), Contextual Applicability (R-3), and Safety Awareness (R-5) and with severity metadata.}
    \label{fig:medprmbench_example}
\end{figure}

\begin{figure*}[!t]
    \centering
    \includegraphics[width=\textwidth, height=0.9\textheight, keepaspectratio]{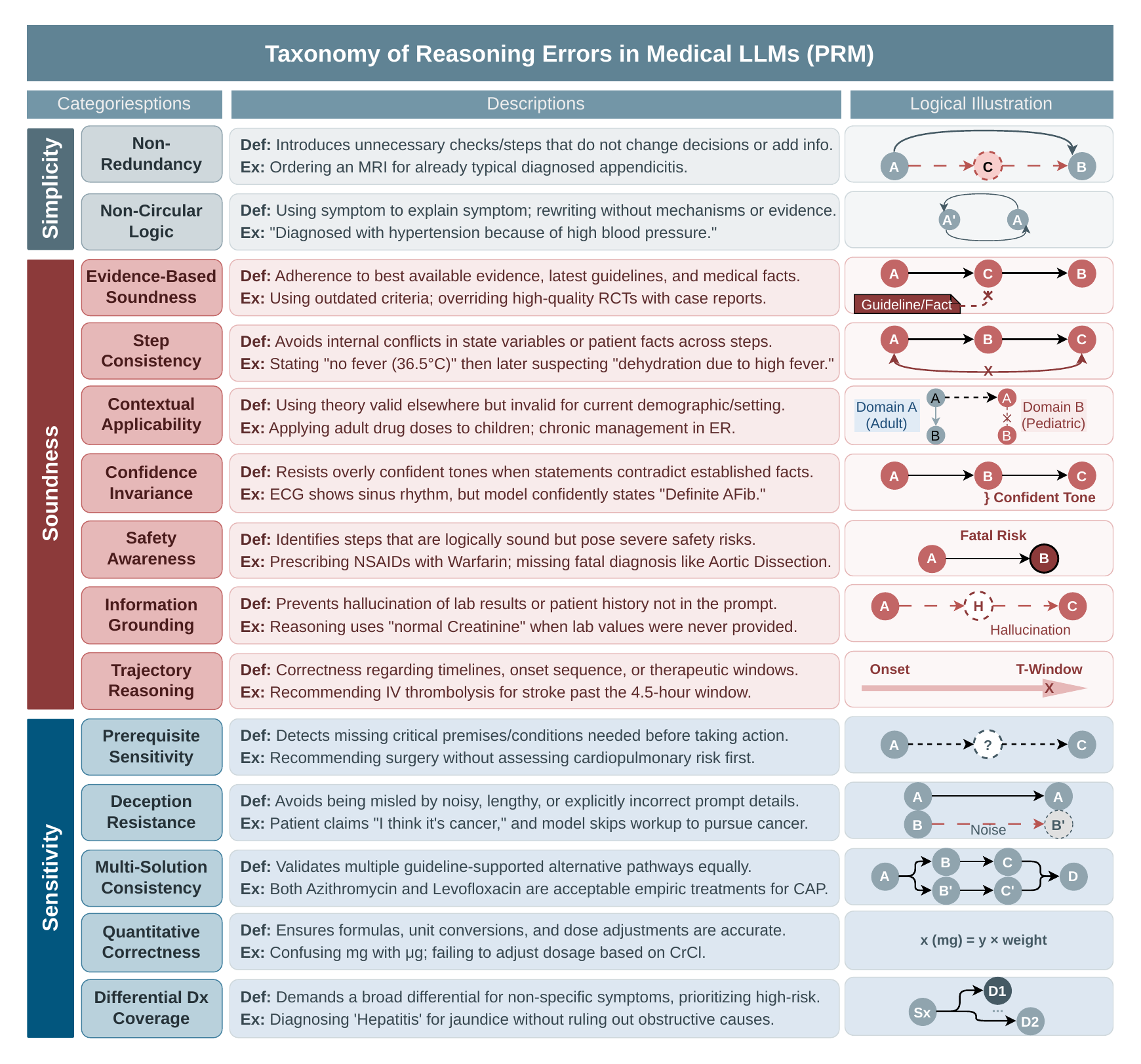}
    \caption{Illustration and description of the 14 error types in MedPRMBench. Error types are organized into three categories: Simplicity (S-1, S-2), Soundness (R-1 through R-7), and Sensitivity (E-1 through E-5). Each type includes its definition, blueprint operation, and example.}
    \label{fig:error_types}
\end{figure*}

Figure~\ref{fig:error_types} presents an illustration and detailed description of the 14 error types in MedPRMBench. These error types are organized into three categories---Simplicity, Soundness, and Sensitivity---covering reasoning deficiencies that may arise in clinical reasoning. Each error type is accompanied by its definition, blueprint operation, and illustrative example to facilitate intuitive understanding of the characteristics and injection methods for each error type.

Below we provide detailed definitions and detection objectives for each of the 14 error types.

\noindent\textbf{Category I: Simplicity}---Efficiency and non-redundancy of clinical reasoning pathways.

\subsection{S-1: Non-Redundancy}
\label{app:error_s1_en}

The reasoning chain introduces unnecessary examinations or processing steps that do not change the clinical decision, and for which more direct, safer, or more cost-effective alternatives exist. When existing information already strongly points to a conclusion, the reasoning still adds formulaic checks without clear purpose, or repeats homogeneous checks for the same objective without additional information gain. \textbf{Example}: Recommending an abdominal MRI for a patient with a clear-cut diagnosis of typical appendicitis. PRMs should identify redundant steps and assign low scores to steps that do not change clinical decisions or provide incremental information.

\subsection{S-2: Non-Circular Logic}
\label{app:error_s2_en}

Using symptoms to explain symptoms, or treating the conclusion as a premise, constituting invalid argumentation logic and tautological pseudo-causation. The reasoning is merely a synonymous restatement without introducing mechanisms, evidence, or differential logic. \textbf{Example}: ``The patient has hypertension because the blood pressure is high''---merely renaming the observed metric without adding any reasoning substance. PRMs should identify circular arguments and tautologies, requiring each step to introduce new information, mechanisms, or evidence.

\smallskip
\noindent\textbf{Category II: Soundness}---Correctness and consistency of medical knowledge and logic.

\subsection{R-1: Evidence-Based Soundness}
\label{app:error_r1_en}

This type is merged from PRMBench's Empirical Soundness and Evidence Compliance. It evaluates whether reasoning conforms to the best currently available evidence (including latest guidelines, pathophysiological knowledge, and pharmacological facts), whether reasonable trade-off justifications are provided when evidence conflicts, and whether cited evidence is applicable to the current population/scenario/problem. Violations include contradicting current guidelines, using obsolete standards, and overriding high-quality evidence with low-quality evidence without justification. \textbf{Example}: Recommending amoxicillin while ignoring a documented penicillin allergy; using obsolete diagnostic criteria as the core basis. PRMs should detect counterfactual errors, outdated evidence citations, evidence hierarchy inversions, and evidence applicability errors.

\subsection{R-2: Step Consistency}
\label{app:error_r2_en}

Whether there exist conflicts between earlier and later statements, state variables, or plans and their supporting evidence within the reasoning path. Facts, parameters, or judgments established in earlier steps are implicitly changed or contradicted in subsequent steps. \textbf{Example}: The reasoning states ``the patient is afebrile'' but a later step suspects ``dehydration due to persistent high fever.'' PRMs should track state variables throughout the reasoning chain and detect any inconsistencies.

\subsection{R-3: Contextual Applicability}
\label{app:error_r3_en}

At a certain step in the reasoning chain, knowledge or theory that is valid in other scenarios or cases but inapplicable to the current reasoning context is used. In medicine, this means incorrectly applying knowledge specific to certain populations (e.g., children, pregnant women, elderly) or settings (e.g., emergency vs.\ outpatient) to the current case. \textbf{Example}: Directly applying adult treatment standards or drug dosages to a pediatric case. PRMs should evaluate whether each reasoning step matches the population and setting characteristics of the current case.

\subsection{R-4: Confidence Invariance}
\label{app:error_r4_en}

PRMs should maintain their judgment when facing statements in the reasoning chain that contradict established facts from the question stem, especially when such statements are presented with overconfident language. When a reasoning step contains an assertion inconsistent with facts explicitly provided in the clinical vignette but stated with absolute certainty, the PRM should identify the error rather than being ``deceived'' by the confident tone. \textbf{Example}: Lab results show serum potassium of 5.8~mmol/L (explicitly given), but the reasoning states ``the patient's electrolytes are completely normal, no intervention needed.'' PRMs should not be misled by overconfident language such as ``definitively,'' ``clearly confirms,'' or ``without question.''

\subsection{R-5: Safety Awareness}
\label{app:error_r5_en}

The safety baseline in medical reasoning, evaluating whether PRMs can identify steps that are logically coherent but carry severe clinical consequences. This includes recognizing high-risk or fatal missed-diagnosis risks, drug contraindications, drug interactions, allergies, pregnancy risks, and bleeding risks. \textbf{Example}: Failing to rule out life-threatening conditions (e.g., aortic dissection, pulmonary embolism) in a patient presenting with acute chest pain; recommending high-dose NSAIDs for a patient on warfarin. PRMs should treat safety compliance as a veto criterion, assigning the lowest scores to steps involving fatal missed diagnoses or safety baseline violations.

\subsection{R-6: Information Grounding Compliance}
\label{app:error_r6_en}

Detecting whether reasoning is strictly grounded in information provided by the question stem or clinical record, without fabricating (hallucinating) test results, past medical history, or guideline content. The model must not ``fill in'' critical information not provided in the stem and use it as a reasoning basis. \textbf{Example}: The stem provides no creatinine information, but the reasoning states ``creatinine is normal, safe to use contrast agent''; fabricating ``history of diabetes'' as a reasoning basis when the stem contains no such information. PRMs should verify the information source of each reasoning step and identify any fabricated content beyond the scope of the provided information.

\subsection{R-7: Trajectory Reasoning}
\label{app:error_r7_en}

Detecting whether reasoning related to the temporal dimension is correct and whether timeline confusion exists. This includes acute vs.\ chronic determination, onset sequence inference, treatment time windows, pharmacokinetic parameters, and disease stage classification. \textbf{Example}: Misjudging an obviously acute onset as a chronic process; recommending intravenous thrombolysis outside the stroke thrombolysis time window (4.5h); evaluating treatment efficacy as inadequate and requesting a medication change before steady-state drug concentration has been reached. PRMs should evaluate the reasonableness of time-related reasoning.

\smallskip
\noindent\textbf{Category III: Sensitivity}---Capturing critical conditions or necessary implicit requirements.

\subsection{E-1: Prerequisite Sensitivity}
\label{app:error_e1_en}

PRMs should detect deficiencies in the reasoning chain where critical premises, assumptions, or necessary conditions are missing, the absence of which renders subsequent reasoning steps invalid or conclusions unreliable. The reasoning chain skips conditions that must be satisfied or confirmed before reaching a conclusion or making a decision. \textbf{Example}: Recommending contrast-enhanced CT without mentioning the need to assess renal function; recommending a medication without checking the patient's allergy history. PRMs should identify which critical prerequisites have been skipped and assess the impact of missing prerequisites on the validity of subsequent reasoning.

\subsection{E-2: Deception Resistance}
\label{app:error_e2_en}

Whether the model can identify erroneous information, distractors, or lengthy but irrelevant narratives in the clinical vignette without being misled. The model should distinguish key clinical clues from noise, maintain focus on the core clinical problem under information overload, and remain vigilant against obviously incorrect information rather than incorporating it as key evidence. \textbf{Example}: Being drawn by a lengthy but irrelevant occupational exposure history toward an incorrect occupational disease diagnosis, while overlooking a briefly mentioned but highly diagnostic key clue. PRMs should evaluate robustness against distracting information.

\subsection{E-3: Multi-Solution Consistency}
\label{app:error_e3_en}

For the same disease or clinical problem, multiple guideline-supported or contextually reasonable treatment pathways may exist (e.g., surgery vs.\ conservative treatment, different first-line drug regimens). The model should recognize and acknowledge the reasonableness of alternative approaches rather than claiming a single correct solution. \textbf{Example}: For early-stage cancer, both surgical resection and stereotactic body radiation therapy (SBRT) are guideline-supported first-line treatments; PRMs should assign high scores to correct steps along both pathways. PRMs should exhibit multi-pathway fairness, basing scoring criteria on the rigor of reasoning logic rather than the uniqueness of the chosen approach.

\subsection{E-4: Quantitative Correctness}
\label{app:error_e4_en}

When the reasoning process requires quantitative calculations or dosage adjustments, evaluating whether necessary calculations have been performed and whether the calculation process is normatively correct. This covers unit conversions, numerical operations, dosing frequency reasonableness, and individualized dosage adjustments based on body weight, body surface area, renal function, and other patient-specific parameters. \textbf{Example}: Confusing mg with $\mu$g, resulting in a 1000-fold dosage error; entering incorrect values when calculating creatinine clearance using the Cockcroft-Gault formula. PRMs should evaluate every aspect of quantitative reasoning, including formula selection, parameter substitution, computational results, and unit consistency.

\subsection{E-5: Differential Diagnosis Coverage}
\label{app:error_e5_en}

When symptoms are highly nonspecific, evaluating whether the reasoning process proposes a necessary differential diagnosis framework rather than providing only a single diagnosis. The framework should include a reasonable differential diagnosis list with next-step exclusion pathways, ordered by danger level and likelihood, and must not omit high-risk but treatable diseases. \textbf{Example}: A patient presents with acute chest pain, but the reasoning considers only acute coronary syndrome without including aortic dissection, pulmonary embolism, tension pneumothorax, and other life-threatening differential diagnoses. PRMs should evaluate the adequacy of differential diagnosis coverage, ensuring that high-risk diseases are not overlooked.


\section{Data Curation: Technical Details}
\label{app:data_curation_details}

This appendix details the three stages of Phase~1 summarized in Section~\ref{sec:data_curation}.

\paragraph{Multi-source aggregation.}
All seven source benchmarks are converted into a unified schema, yielding a raw pool $\mathcal{D}_{\text{raw}}$ of approximately 226{,}439 instances. Based on the availability of reasoning text, sources are categorized into \textbf{Class~A} (MedCaseReasoning), which comes with expert-authored step-by-step differential-diagnosis reasoning, and \textbf{Class~B} (the remaining six datasets), which contain only questions and answers and therefore require reasoning-trace generation.

\paragraph{Difficulty-aware filtering.}
Trivially easy questions contribute little discriminative signal for process-reward evaluation, while excessively ambiguous ones introduce noise. We estimate question difficulty by \emph{single-model repeated sampling}: for each question $x \in \mathcal{D}_{\text{raw}}$, we query GPT-4o-mini $K{=}8$ times at temperature $\tau{=}0.7$ and compute the empirical pass rate
\begin{equation}
    p(x) = \frac{1}{K} \sum_{k=1}^{K} \mathbb{1}\bigl[\hat{y}_k = y^*\bigr],
\end{equation}
where $\hat{y}_k$ is the predicted answer on the $k$-th sample and $y^*$ the gold answer. We retain only questions with $p(x) \leq \theta$, $\theta{=}0.5$, i.e., questions the probe model answers incorrectly at least half the time; questions with $p(x){=}0$ are explicitly retained, as they often represent genuinely challenging clinical scenarios. The same pass-rate cache underlies the Hard Subset's zero-pass-rate criterion in Section~\ref{sec:hard_subset}.

\paragraph{Rejection-sampled reasoning generation.}
Class~A reasoning text requires no LLM generation and is segmented by rule-based regex splitting. For each Class~B question in the filtered pool, we generate a step-by-step reasoning trace via \emph{rejection sampling} over a pool of three frontier models (Gemini-3-Pro-preview, DeepSeek-R1, GPT-5.2), used in round-robin order: the model is prompted to produce a comprehensive stepwise clinical analysis with the final answer in $\boxed{\cdot}$ notation, and if the answer is incorrect or the response fails JSON parsing, the system cycles to the next model, up to $N_{\text{ans}}{=}5$ attempts; the first accepted trace is retained. The output of the three stages is the curated dataset $\mathcal{D}_{\text{curated}} = \{(q_i, r_i, a_i)\}_{i=1}^{N}$ used by Phase~2.

\section{Blueprint Distillation: Technical Details}
\label{app:blueprint_details}

This appendix provides the technical details of the blueprint distillation pipeline that are summarized at a high level in Section~\ref{sec:blueprint_construction}. We describe each sub-step in full, present the formal algorithm, and conclude with a complete worked example drawn from the MedPRMBench test set.

\subsection{Multi-Model Semantic Voting Details}
\label{app:semantic_voting}

Section~\ref{sec:blueprint_construction} provides a high-level overview of the multi-model semantic voting strategy. Here we present the detailed matching rule and its fixed implementation thresholds.

Let $\mathcal{M} = \{m_1, m_2, m_3\}$ denote the model pool, where each model $m_j$ independently extracts a triplet set $\mathcal{T}^{(j)}$. For a candidate triplet $t = (v_s, p, v_o)$, we determine whether it receives sufficient support across models via semantic embeddings (bge-large-en-v1.5). Specifically, two triplets $t_a = (v_s^a, p^a, v_o^a)$ and $t_b = (v_s^b, p^b, v_o^b)$ are considered semantically equivalent if and only if:
\begin{equation}
\begin{aligned}
    \text{sim}(v_s^a, v_s^b) &\geq \delta_e \;\wedge\;
    \text{sim}(p^a, p^b) \geq \delta_r \\
    &\wedge\; \text{sim}(v_o^a, v_o^b) \geq \delta_e,
\end{aligned}
\end{equation}
where $\text{sim}(\cdot, \cdot)$ denotes cosine similarity of normalized embedding vectors, and $\delta_e = 0.7$, $\delta_r = 0.6$ are the entity and relation matching thresholds, respectively. Only triplets supported by at least $\mu = 2$ distinct models are retained:
\begin{equation}
    \mathcal{T}_{\text{ERN}} = \bigl\{ t \mid |\{m_j : \exists\, t' \in \mathcal{T}^{(j)},\; t' \approx_{\delta} t \}| \geq \mu \bigr\}.
\end{equation}

We choose embedding-based semantic matching over LLM pairwise judgment because triplet alignment requires comparing $O(n^2)$ candidate pairs---embedding computation completes batch comparisons in milliseconds, far more efficient than per-pair LLM calls; meanwhile, the multi-model voting consensus mechanism ($\mu \geq 2$) reduces the impact of single-model extraction noise.

\subsection{Blueprint Distillation Supplementary Details}
\label{app:distillation_details}

Section~\ref{sec:blueprint_construction} describes the four steps of blueprint distillation. This section provides additional technical details not covered in the main text.

\emph{Semantic bridging threshold.} In semantic bridging, a bridging edge is added when two nodes $v_a, v_b$ satisfy $\text{sim}(\phi(v_a), \phi(v_b)) \geq \delta_{\text{bridge}}$. In the medical domain, the same entity often has multiple surface forms (e.g., ``myocardial infarction'' vs.\ ``MI'' vs.\ ``heart attack''), and the choice of $\delta_{\text{bridge}}$ must balance recall against the risk of spurious connections.

\emph{Transitive reduction formalization.} The transitive reduction mentioned in the main text can be formalized as follows: for a DAG $G=(V,E)$, remove every edge $(u,v)$ for which a path of length $\geq 2$ from $u$ to $v$ exists:
\begin{equation}
\resizebox{.98\columnwidth}{!}{$
    E^*_{\text{red}} =
    \bigl\{(u,v) \in E^* \mid \nexists\; \text{path}(u, v)
    \;\text{in}\; (V^*, E^* \setminus \{(u,v)\})\;
    \text{with length} \geq 2\bigr\}.
$}
\label{eq:transitive_reduction}
\end{equation}
This encourages the blueprint to retain direct causal edges that are not already represented through other paths.

\subsection{CRB Construction Algorithm}
\label{app:crb_algorithm}

Algorithm~\ref{alg:crb_construction_en} summarizes the end-to-end CRB construction pipeline for Class~B datasets, integrating multi-model semantic voting, bidirectional BFS distillation, and sufficiency verification into a unified procedure.
\begin{algorithm*}[!t]
\footnotesize
\caption{Clinical Reasoning Blueprint Construction (Class~B Datasets)}
\label{alg:crb_construction_en}
\begin{algorithmic}[1]
\Require Curated dataset $\mathcal{D}_{\text{curated}} = \{(q_i, r_i, a_i)\}$, model pool $\mathcal{M} = \{m_1, m_2, m_3\}$, embedding encoder $\phi$, thresholds $\delta_e, \delta_r, \mu$
\Ensure Clinical Reasoning Blueprint dataset $\mathcal{D}_{\text{CRB}}$
\For{each $(q, r, a) \in \mathcal{D}_{\text{curated}}$}
    \Statex \hspace{\algorithmicindent}\textcolor{gray}{\textit{// Phase I: Multi-Model Semantic Voting ERN Extraction}}
    \For{each model $m_j \in \mathcal{M}$}
        \State $\mathcal{T}^{(j)} \gets \Call{ExtractTriplets}{m_j, q, r}$
    \EndFor
    \State $\mathcal{T}_{\text{all}} \gets \bigcup_{j} \mathcal{T}^{(j)}$
    \State $\mathcal{T}_{\text{ERN}} \gets \emptyset$
    \For{each candidate $t = (v_s, p, v_o) \in \mathcal{T}_{\text{all}}$}
        \State $\text{cnt} \gets \bigl|\{m_j : \exists\, t' \!\in\! \mathcal{T}^{(j)},\; \text{sim}(\phi(v_s), \phi(v_s')) \!\geq\! \delta_e \,\wedge\, \text{sim}(\phi(p), \phi(p')) \!\geq\! \delta_r \,\wedge\, \text{sim}(\phi(v_o), \phi(v_o')) \!\geq\! \delta_e\}\bigr|$
        \If{$\text{cnt} \geq \mu$}
            \State $\mathcal{T}_{\text{ERN}} \gets \mathcal{T}_{\text{ERN}} \cup \{t\}$
        \EndIf
    \EndFor
    \State $G_{\text{ERN}} \gets \Call{BuildGraph}{\mathcal{T}_{\text{ERN}}}$
    \Statex \hspace{\algorithmicindent}\textcolor{gray}{\textit{// Phase II: Blueprint Distillation}}
    \State $v_c \gets \arg\max_{v \in V} \text{sim}\bigl(\phi(v), \phi(a)\bigr)$ \Comment{Conclusion node identification}
    \State $G_{\text{bridged}} \gets \Call{SemanticBridging}{G_{\text{ERN}}}$ \Comment{Reconnect fragments}
    \State $V^*, E^* \gets \Call{BidirectionalBFS}{G_{\text{bridged}}, v_c}$ \Comment{Subgraph extraction}
    \State $G^* \gets \Call{TransitiveReduction}{V^*, E^*}$ \Comment{Remove shortcut edges}
    \Statex \hspace{\algorithmicindent}\textcolor{gray}{\textit{// Phase III: Sufficiency Verification \& Enhancement}}
    \If{$\neg\,\Call{SufficiencyCheck}{G^*, q, a}$ \textbf{or} $|E^*| < \eta_{\min}$}
        \State $\mathcal{T}_{\text{sup}} \gets \Call{SelectSupplementary}{G_{\text{ERN}}, G^*, q, a}$
        \State $G^* \gets G^* \cup \{t \in \mathcal{T}_{\text{sup}} \mid \Call{IsConnected}{t, G^*}\}$
    \EndIf
    \State $\mathcal{D}_{\text{CRB}} \gets \mathcal{D}_{\text{CRB}} \cup \{(q, r, a, G_{\text{ERN}}, G^*)\}$
\EndFor
\State \Return $\mathcal{D}_{\text{CRB}}$
\end{algorithmic}
\end{algorithm*}


\subsection{Blueprint Node Criticality}
\label{app:bnc}

For Class~B datasets, we compute a criticality score for each node based on graph-theoretic metrics, capturing its structural importance in the reasoning path:
  \begin{equation}
  \begin{split}
      \text{BNC}(v) = {} & \alpha \cdot \text{BC}(v) + \beta \cdot \frac{1}{\text{dist}(v, v_c)} \\
      & + \gamma \cdot \frac{\text{deg}(v)}{\max_{u \in V^*} \text{deg}(u)},
  \end{split}
      \label{eq:bnc}
  \end{equation}
where $\text{BC}(v)$ is betweenness centrality, $\text{dist}(v, v_c)$ is the shortest-path distance from node $v$ to the conclusion node $v_c$, $\text{deg}(v)$ is the node degree (in-degree + out-degree), and the weight coefficients are $\alpha = 0.4$, $\beta = 0.35$, $\gamma = 0.25$. Nodes with higher BNC scores are more critical in the reasoning chain; subsequent error injection preferentially selects high-BNC nodes as operation targets. Class~A datasets do not compute BNC.


\section{Complete Blueprint Construction Example}
\label{app:blueprint_example}

Figure~\ref{fig:crb_example} presents an end-to-end CRB construction example using a real data instance from the MedPRMBench test set (\texttt{medprm\_medxpertqa\_001902}, Class~B). The figure illustrates each stage of the pipeline: from the original medical question and gold reasoning text (Phase~1), through multi-model semantic voting ERN extraction (Phase~2; Qwen3-Max / GPT-5.2 / Claude-Opus-4.5; total input = 36 triplets, 10 accepted, acceptance rate = 27.78\%), blueprint distillation with conclusion node identification ($v_c$ = ``impact of anterior aspect of humeral head against posterior glenoid rim'', sim = 0.7884), semantic bridging (1 bridge edge added), bidirectional BFS (depth = 5, all 11 nodes reachable), and transitive reduction (compression\_rate = 0.0, no edges removed) (Phase~3), to safety-critical annotation and linearization (Phase~4). The final CRB contains 11 nodes and 11 edges (10 ERN edges + 1 semantic bridge edge), with a cycle-back edge (``reverse Hill-Sachs lesion $\to$ is mechanically intrinsic to $\to$ posterior glenohumeral dislocation'') retained as a non-causal attribute edge.

\begin{sidewaysfigure}[p]
    \centering
    \includegraphics[width=0.95\textheight, height=0.92\textwidth, keepaspectratio]{crb.png}
    \caption{End-to-end CRB construction example for \texttt{medprm\_medxpertqa\_001902} (Class~B, MedXpertQA). The figure shows all four phases: (1)~input question and gold reasoning text; (2)~ERN extraction via multi-model semantic voting; (3)~blueprint distillation including conclusion node identification, semantic bridging, bidirectional BFS, and transitive reduction; and (4)~safety-critical annotation with linearized reasoning steps. Node colors in the graph encode roles: blue = conclusion node $v_c$, yellow = bridge-connected nodes, grey = standard entity nodes. Edge styles distinguish causal edges (solid), cycle-back edges (orange dashed), and semantic bridge edges (yellow dashed).}
    \label{fig:crb_example}
\end{sidewaysfigure}


\section{Composite Error Synthesis and Deterministic Verification: Technical Details}
\label{app:composite_synthesis}

This appendix provides the full technical details of the composite error synthesis and deterministic verification pipeline summarized at a high level in Section~\ref{sec:composite_and_verification}.

\subsection{Composite Error Synthesis}
\label{app:composite_synthesis_detail}

Reasoning failures may involve interacting errors across multiple steps. To construct a challenging \emph{hard subset}, we synthesize \emph{composite error variants} containing 2--3 error types simultaneously in a single coherent reasoning chain.

For each instance with $|\mathcal{V}_{\text{single}}(i)| \geq 2$ single-error variants, we select error type combinations according to three prioritization criteria: (1)~\textbf{cross-category diversity}---preferring combinations spanning different error categories (e.g., one Simplicity + one Soundness + one Sensitivity error), maximizing the breadth of reasoning defects; (2)~\textbf{step disjointness}---avoiding combinations that modify the same step, preventing conflicting modifications that would produce incoherent text; (3)~\textbf{severity diversity}---preferring combinations mixing different severity levels (e.g., one Major + one Minor), testing whether PRMs can simultaneously detect severe and subtle errors. At most $K_{\text{comp}} = 2$ composite variants are generated per instance, each containing $n \in [2, 3]$ error types.

Unlike naive step-level merging, composite synthesis is performed by an LLM that receives the original reasoning chain $\mathbf{s}$ together with all selected single-error variants $\{(\tilde{\mathbf{s}}^{(c_1)}, c_1), \ldots, (\tilde{\mathbf{s}}^{(c_n)}, c_n)\}$, integrating all errors into a single coherent chain $\tilde{\mathbf{s}}_{\text{comp}}$ while resolving potential conflicts (e.g., one error's modification altering the context that a subsequent error depends on) and preserving each individual error's precise nature and subtlety:
\begin{equation}
\resizebox{.98\columnwidth}{!}{$
    \mathcal{V}_{\text{comp}}(i) =
    \bigl\{(\tilde{\mathbf{s}}_{\text{comp}}, \{c_1, \ldots, c_n\},
    \{\mathbf{e}^{(c_1)}, \ldots, \mathbf{e}^{(c_n)}\})\bigr\}.
$}
\end{equation}

\subsection{Deterministic Text Diff Verification}
\label{app:diff_verification}

LLM-reported error step indices may be inaccurate: models may report a step as modified when it was not actually changed, or modify a step without reporting it. To ensure label accuracy, we apply a fully deterministic verification step based on text differencing.

For each variant (both single-error and composite), we align the original step sequence $\mathbf{s} = (s_1, \ldots, s_L)$ with the corrupted sequence $\tilde{\mathbf{s}} = (\tilde{s}_1, \ldots, \tilde{s}_{L'})$ using the \texttt{SequenceMatcher} algorithm (a variant of the Ratcliff/Obershelp algorithm), handling cases where $L' \neq L$ (e.g., S-1 inserts new steps) and avoiding false positives caused by positional shifts. The alignment produces an opcode sequence that classifies each corrupted step as \textbf{equal} (text identical to the original step, confirmed correct), \textbf{replace} (text differs, confirmed modified), or \textbf{insert} (no corresponding original step, confirmed as newly inserted).

We then reconcile the LLM-reported error indices $\mathbf{e}_{\text{reported}}$ with the deterministic diff result $\mathbf{e}_{\text{diff}}$: $k \in \mathbf{e}_{\text{reported}} \cap \mathbf{e}_{\text{diff}}$ is \emph{confirmed} (retained); $k \in \mathbf{e}_{\text{reported}} \setminus \mathbf{e}_{\text{diff}}$ is a \emph{false positive} (removed); $k \in \mathbf{e}_{\text{diff}} \setminus \mathbf{e}_{\text{reported}}$ is \emph{unreported} (added). Variants with $\mathbf{e}_{\text{diff}} = \emptyset$ (no steps were actually modified) are discarded entirely. The final verified error step set is $\mathbf{e}_{\text{verified}} = \mathbf{e}_{\text{diff}}$.


\section{Quality Control: Technical Details}
\label{app:quality_control}

This appendix details the two-stage quality-assurance pipeline summarized in Section~\ref{sec:quality_control}: automatic filtering (\S\ref{app:auto_filtering}) and expert review with multi-model consistency checks (\S\ref{app:expert_review}).

\subsection{Automatic Filtering}
\label{app:auto_filtering}

Each verified variant undergoes three automatic quality assessments. \emph{Text fidelity}: We compute a text fidelity score $\text{TF}(\mathbf{s}, \tilde{\mathbf{s}})$ using the \texttt{SequenceMatcher} ratio; variants with $\text{TF} < \tau_{\text{TF}} = 0.10$ are discarded, as the error was not effectively injected. \emph{Error obviousness}: We assess detection difficulty based on error step position and the fraction of corrupted steps:
\begin{equation}
    \text{Obv}(\tilde{\mathbf{s}}) = f\!\left(\frac{|\mathbf{e}_{\text{verified}}|}{L'},\; \text{pos}(\mathbf{e}_{\text{verified}})\right),
\end{equation}
where $\text{pos}(\cdot)$ captures the relative position of error steps; variants with $\text{Obv} > 0.8$ receive a reduced sample weight ($w_{\text{sample}} = 0.3$). \emph{Answer impact analysis}: An LLM derives the final answer from $\tilde{\mathbf{s}}$ without seeing the correct answer; the extracted answer $\tilde{a}$ is compared with the gold answer $a_i$:
\begin{equation}
\resizebox{.75\columnwidth}{!}{$
    \text{AnswerChanged}(i) = \begin{cases}
        \text{true} & \text{if } \tilde{a} \neq a_i, \\
        \text{false} & \text{if } \tilde{a} = a_i, \\
        \text{unknown} & \text{if extraction fails}.
    \end{cases}
$}
\end{equation}
This signal is retained as metadata rather than used for filtering, as both answer-changing and answer-preserving errors are valuable for PRM evaluation.

\subsection{Expert Review and Multi-Model Consistency Checks}
\label{app:expert_review}

Medical professionals review each instance via a structured annotation interface along two dimensions: (1)~\emph{original reasoning correctness}---whether the original steps $\mathbf{s}$ are factually correct and logically coherent, with specific erroneous steps annotated and corrected reasoning provided when errors are found; and (2)~\emph{error annotation accuracy}---whether the system-generated error step--code mapping is accurate, with a corrected mapping provided otherwise. Reviewers also record rationales; instances with incomplete annotations are flagged and excluded from subsequent processing. As described in Section~\ref{sec:quality_control}, a shared 1{,}000-candidate MedCaseReasoning (Class~A) subset receives independent blinded re-annotation to estimate reliability. Every exported record also passes a second-stage check (86 physicians: 82 annotator and 17 checker accounts, with overlap; the checker differs from the annotator in 15{,}763 of 15{,}982 records, which count the 1{,}000 doubly annotated candidates twice). Checkers flag and correct 2{,}735 of the 15{,}982 records (17.1\%), and an additional acceptance-stage spot check of 504 records finds no further errors. Review is substantive rather than confirmatory: annotators judge the original reference reasoning erroneous in 12.2\% of records (17.7\% for expert-authored Class~A, 5.3\% for generated Class~B), supplying corrected reasoning in those cases.

To guard against individual-reviewer subjectivity, expert corrections are additionally cross-checked by a heterogeneous three-model voting mechanism that acts as a consistency filter. For corrections to the original reasoning, three heterogeneous models (Gemini-3.1-Pro, Claude-Sonnet-4.6, GPT-5.4) independently judge whether the expert's correction is medically sound and well-justified, with a $\geq 2/3$ majority required; voting follows a conservative principle---corrections are supported only when the expert identifies genuine factual errors, logical flaws, or inaccuracies, and are flagged when the original reasoning is medically correct and the critique is overly strict or preference-based. For corrections to the error-annotation mapping, an analogous three-model vote (Gemini-3.1-Pro, Claude-Opus-4.5, GPT-5.4) assesses whether the corrected error-type classification is justified. Model votes never rewrite physician labels; they only decide whether a physician correction is adopted. Of the 14{,}981 completed annotations, 13{,}381 are adopted and 1{,}600 are excluded because the vote rejects the physician's reasoning or error-map correction (1{,}536 low- and 64 high-conflict cases). The remaining attrition is one physician-discarded out-of-scope case and two adopted cases whose revised maps contain no error code, giving 13{,}379 questions (6{,}879 train, 6{,}500 test) before the nine train--test overlaps are removed (Section~\ref{sec:quality_control}).

For adopted revisions, three updates are performed sequentially: (a)~\emph{original reasoning step rewriting}---expert-identified error steps and corrections are used to rewrite the corresponding steps in $\mathbf{s}$ via an LLM, keeping the corrupted chain $\tilde{\mathbf{s}}$ unchanged since expert annotations are based on it; (b)~\emph{error annotation replacement}---the expert-corrected mapping replaces the system mapping, with synchronized updates to error codes, categories, and step indices; and (c)~\emph{rule-based \texttt{modified\_steps} update}---step-by-step comparison of the revised $\mathbf{s}$ and $\tilde{\mathbf{s}}$ recomputes the difference set.


\section{Applicability Tagging and Distribution-Aware Sampling: Technical Details}
\label{app:applicability_sampling}

For each instance $(q_i, \mathbf{s}_i, a_i)$, an LLM evaluates the applicability of all 14 types based on the question content, reasoning chain, and blueprint annotations ($\text{SC}_i$, $\text{Pre}_i$, and $\text{BNC}$ scores for Class~B), outputting a binary applicability vector $\mathbf{a}_i \in \{0, 1\}^{14}$, where $a_i^{(c)} = 1$ indicates that error type $c$ can be meaningfully injected. Three types---R-1 (Evidence-Based Soundness), R-6 (Information Grounding Compliance), and E-2 (Deception Resistance)---are designated as universally applicable ($a_i^{(c)} = 1$ for all $i$).

From the applicable types, we select 1--3 target error types per instance using distribution-aware sampling. Let $\pi(c)$ denote the target proportion for error type $c$ (e.g., $\pi(\text{R-1}) = 0.12$, $\pi(\text{E-5}) = 0.10$), and let $\hat{\pi}(c)$ denote the current empirical proportion. We compute a \emph{demand score} for each applicable type:
\begin{equation}
    d(c) = \pi(c) - \hat{\pi}(c),
\end{equation}
and sample types with probability proportional to $\max(d(c), \epsilon)$, where $\epsilon > 0$ is a small floor ensuring nonzero selection probability. This mechanism drives the final error type distribution toward the design target $\pi$.

\paragraph{Target step selection.}
Given a selected error type $c$ and step sequence $\mathbf{s} = (s_1, \ldots, s_L)$, the implementation uses the construction annotations from Section~\ref{sec:blueprint_construction} in four type-dependent target-selection rules: \emph{safety-critical types} (R-5, E-1) select steps with high safety criticality $\text{SC}(s_k) \in \{\text{Critical}, \text{Major}\}$ or steps flagged as prerequisites; \emph{consistency types} (R-2, R-7) select pairs of steps referencing related clinical findings; \emph{knowledge types} (R-1, R-3, R-6) select steps containing specific medical claims; and \emph{structural types} (S-1, S-2) select positions for redundant-step insertion or circular-dependency construction. For Class~B datasets, the Blueprint Node Criticality score (Appendix~\ref{app:bnc}) is an additional input and high-BNC nodes are preferentially selected. These rules specify the released construction procedure; they are not validated as optimal against alternative target-selection procedures.


\section{Severity Assessment: Technical Details}
\label{app:severity_assessment}

Each corrupted variant is assigned a continuous construction severity score combining three factors:
\begin{equation}
\begin{aligned}
    \text{Sev}(\tilde{\mathbf{s}}, c) =
    &\alpha_1 \underbrace{\frac{|\mathbf{e}|}{L}}_{\text{disruption}}
    + \alpha_2 \underbrace{w_{\text{SC}}(s_k)}_{\text{SC weight}} \\
    &+ \alpha_3 \underbrace{w_{\text{type}}(c)}_{\text{severity}},
\end{aligned}
\end{equation}
where $\alpha_1 = \alpha_2 = 0.35$, $\alpha_3 = 0.30$; $|\mathbf{e}|/L$ is the fraction of steps affected; $w_{\text{SC}}(s_k)$ maps the safety criticality level to a numerical weight ($\text{Critical} \!\mapsto\! 1.0$, $\text{Major} \!\mapsto\! 0.7$, $\text{Moderate} \!\mapsto\! 0.4$, $\text{Minor} \!\mapsto\! 0.1$); and $w_{\text{type}}(c)$ is the inherent severity weight from Table~\ref{tab:error_taxonomy}. For Class~B datasets, the BNC value augments $w_{\text{SC}}$ when available, reflecting the structural importance of the corrupted node.

Separately, a deterministic rule assigns one of four severity levels:
\begin{equation}
\resizebox{.88\columnwidth}{!}{$
    \text{SevLevel} = \begin{cases}
        \text{Critical} & |\mathbf{e}|/L \!\geq\! 0.7 \,\wedge\, \text{SC}(s_k) \!\in\! \{\text{Crit., Maj.}\} \\
        \text{Major} & |\mathbf{e}|/L \!\geq\! 0.4 \,\vee\, \text{SC}(s_k) \!\in\! \{\text{Crit., Maj.}\} \\
        \text{Moderate} & |\mathbf{e}|/L \!\geq\! 0.2 \,\vee\, \text{SC}(s_k) \!=\! \text{Moderate} \\
        \text{Minor} & \text{otherwise}
    \end{cases}
$}
\end{equation}

The four rule-generated levels are imbalanced in the release. The training split contains 53 Minor (0.77\%), 234 Moderate (3.41\%), 3{,}551 Major (51.69\%), and 3{,}032 Critical (44.13\%) instances; the evaluation split contains 17 Minor (0.26\%), 157 Moderate (2.42\%), 3{,}008 Major (46.28\%), and 3{,}318 Critical (51.05\%) instances. These are descriptive construction indicators, not independently physician-rated clinical harm levels; severity-stratified analyses must account for both the labeling rule and class prevalence.


\section{Dataset Splitting: Technical Details}
\label{app:dataset_splitting}

\begin{table}[!htb]
\centering
\caption{Statistics of MedPRMBench by error-type occurrence. Averages cover all 13{,}370 corrupted chains and questions across the training and evaluation splits; the test-count row covers only evaluation instances. Per-type columns overlap. Abbreviations are defined in Table~\ref{tab:error_taxonomy}.}
\label{tab:medprmbench_stats}
\small
\resizebox{\textwidth}{!}{%
\begin{tabular}{l|c|cc|ccccccc|ccccc}
\toprule
 & \textbf{Overall} & \textbf{NR} & \textbf{NCL} & \textbf{EBS} & \textbf{SC} & \textbf{CA} & \textbf{CI} & \textbf{SA} & \textbf{IGC} & \textbf{TR} & \textbf{PS} & \textbf{DR} & \textbf{MSC} & \textbf{QC} & \textbf{DDC} \\
\midrule
Avg.\ Steps & 9.6 & 10.9 & 8.2 & 9.2 & 9.5 & 9.8 & 9.3 & 9.7 & 8.0 & 9.5 & 10.0 & 9.7 & 12.1 & 9.4 & 10.0 \\
Avg.\ Error Steps & 3.3 & 4.3 & 3.1 & 3.5 & 3.5 & 3.4 & 3.4 & 3.5 & 3.4 & 3.6 & 3.4 & 3.4 & 3.2 & 3.3 & 3.5 \\
Avg.\ 1st Error & 3.2 & 3.3 & 2.8 & 3.1 & 3.2 & 3.2 & 2.9 & 3.4 & 2.7 & 3.2 & 3.3 & 3.2 & 4.1 & 3.3 & 3.0 \\
Avg.\ Q Length & 917 & 871 & 892 & 935 & 887 & 996 & 1077 & 1040 & 1018 & 939 & 924 & 823 & 939 & 1006 & 997 \\
\# Test Instances & 6{,}500 & 1{,}555 & 1{,}028 & 4{,}020 & 1{,}481 & 1{,}098 & 1{,}137 & 1{,}483 & 2{,}640 & 1{,}306 & 998 & 977 & 667 & 892 & 1{,}860 \\
\# Instances & 13{,}370 & 3{,}149 & 1{,}964 & 8{,}282 & 2{,}968 & 2{,}240 & 2{,}204 & 2{,}903 & 5{,}287 & 2{,}620 & 2{,}105 & 2{,}057 & 1{,}475 & 1{,}759 & 3{,}760 \\
\bottomrule
\end{tabular}%
}
\end{table}

Beyond the per-type statistics in Table~\ref{tab:medprmbench_stats}, two aggregate comparisons are worth noting. MedPRMBench features substantially longer question texts than PRMBench (917 vs.\ 153 characters on average), reflecting the rich clinical context of medical questions, and per-type differences track inherent error semantics: Multi-Solution Consistency has the latest average first-error position (4.1), since such errors require sufficient reasoning context to establish, while Information Grounding Compliance has the earliest (2.7), since fabricated-information errors can be injected early in a chain.

The final dataset is split into train/test partitions based on the original source datasets' splits. Instances are first triaged by adoption status: only those with complete physician annotations/checks and completed model-assisted consistency checks enter the train/test split; instances with unresolved consistency failures or missing annotations are filtered out. For instances entering the split, each is mapped back to its source dataset's \texttt{\_split} field via \texttt{instance\_id}, preserving the original partition (train $\to$ train, test/val/dev $\to$ test). Importantly, MedQA-USMLE and MedMCQA are designated as \emph{protected datasets}: their training data is retained in the training set and never reassigned to the test set, keeping the train/test partition for these two datasets consistent with the original split and reducing leakage risk in downstream experiments (Section~\ref{sec:prm_verifier}). When the test set proportion is insufficient, only training records from non-protected datasets are reassigned to the test set, maintaining stratified balance across \texttt{dataset\_type} $\times$ primary error type. A final normalized-question audit identified nine overlapping training records, which are removed from the 6{,}870-instance training release.

\paragraph{Release-integrity audit.}
Release checks never alter reasoning text or finalized \texttt{error\_steps}; they remove the nine overlapping training records and normalize code-map bookkeeping fields (counts in the artifact manifest). The resulting 6{,}870/6{,}500 split files have unique IDs; for every record, the union of mapped steps equals \texttt{error\_steps}, top-level and mapped type sets agree, and all indices are in range. The artifact manifest records these operations and split hashes.



\section{PRM Baseline Training Details}
\label{app:prm_details}
Our benchmark-specific PRM baseline is obtained by fine-tuning
Qwen3-8B~\citep{yang2025qwen3} on the MedPRMBench training split. We follow the
standard process-supervision training paradigm introduced by
\citet{lightman2023lets}.

\smallskip
\noindent\textbf{Input Format.}
Each training sample consists of a medical question (with answer options, if
present) and a step-by-step reasoning chain. We prepend a system prompt
instructing the model to evaluate the logicality and validity of each reasoning
step, outputting ``+'' for correct steps and ``-'' for erroneous ones. Each
reasoning step is terminated by a special delimiter token ``\texttt{ki}''
(added to the tokenizer as an additional special token). The input is formatted
using the model's chat template with a system message, a user message containing
the question and explanation, and a generation prompt.

\smallskip
\noindent\textbf{Training Objective.}
The loss is computed only at step boundary positions (i.e., the delimiter token
``\texttt{ki}''). At each such position, we extract the logits for the ``+''
and ``-'' tokens, form a two-class distribution via softmax, and compute
cross-entropy against the ground-truth label. For hard labels, the target is
$(1, 0)$ for correct steps and $(0, 1)$ for erroneous steps. All other token
positions are masked with label $-100$ and excluded from the loss computation.

\smallskip
\noindent\textbf{Hyperparameters.}
We fine-tune the full model (no LoRA or adapter) using DeepSpeed ZeRO-3 on
8$\times$NVIDIA H20 GPUs. The training configuration is summarized in
Table~\ref{tab:prm_hyperparams}.

\begin{table}[t]
\centering
\caption{Hyperparameters for MedPRMBench-RM training.}
\label{tab:prm_hyperparams}
\small
\begin{tabular}{ll}
\toprule
\textbf{Hyperparameter} & \textbf{Value} \\
\midrule
Initialization & Qwen3-8B \\
Training data & 6{,}870 instances / 13{,}740 chains \\
Precision & bfloat16 \\
Max sequence length & 2{,}048 tokens \\
Per-device batch size & 2 \\
Gradient accumulation steps & 4 \\
Effective batch size & $2 \times 8 \times 4 = 64$ \\
Learning rate & $1 \times 10^{-5}$ \\
LR scheduler & Cosine with warmup \\
Warmup ratio & 0.05 \\
Number of epochs & 1 \\
Optimizer & AdamW \\
Max gradient norm & 1.0 \\
Distributed strategy & DeepSpeed ZeRO Stage-3 \\
Random seed & 42 \\
\bottomrule
\end{tabular}
\end{table}

\smallskip
\noindent\textbf{Training Data.}
The MedPRMBench training split contains 6{,}870 medical instances after removing
nine normalized-question train--test overlaps.
For each retained instance, we construct paired process
supervision data consisting of one correct reasoning chain and one
error-annotated reasoning chain, yielding 13{,}740 reasoning chains in total.
Each chain is segmented into reasoning steps and annotated with binary hard
labels indicating step-level correctness. Samples whose number of delimiter
tokens does not match the number of step labels are filtered out.

\smallskip
\noindent\textbf{Model Selection.}
We train for one pre-specified epoch and report checkpoint-215, frozen at the
epoch boundary before any downstream test result is inspected. MedPRMBench
test-set metrics are not used to select checkpoints, tune thresholds, or choose
prompt variants.

\section{Evaluated Model Details}
\label{app:model_details}
\subsection{Process Reward Models (PRMs)}
PRMs assess the correctness of intermediate reasoning steps, typically through
binary predictions or scalar step-level scores \citep{lightman2023lets}. We evaluate two PRM baselines. For the released Med-PRM-Reward v1.0 model \citep{yun2025medprm}, we use the released reward-model checkpoint without additional fine-tuning. We convert each MedPRMBench reasoning chain into the same step-boundary verification format used for our PRM and extract the normalized probabilities of the ``+'' and ``-'' tokens at each step boundary. No threshold tuning is performed on the MedPRMBench test set; the default threshold $P(+) \geq 0.5$ is used for all PRM-style models. The second PRM baseline is our benchmark-specific model obtained by 
fine-tuning Qwen3-8B on the MedPRMBench training split under a standard PRM objective. 
In our model, each reasoning step is separated by a boundary token and labeled 
as either ``+'' or ``-'', and the probability of ``+'' is used as the step-level 
correctness score. Full training details of our PRM baseline are deferred to 
Appendix~\ref{app:prm_details}.

\subsection{Critic Models}
In addition to specialized PRMs, we also evaluate critic models, i.e., general
language models used as step-level verifiers for medical reasoning. We consider
two groups of critic models.

The first group comprises open-source medical models, including TX-Gemma \citep{txgemma}, 
Llama-3.1-Meditron3-8B \citep{llama3meditron}, Meerkat \citep{meerkat}, UltraMedical 
\citep{ultramedical}, and HuatuoGPT-o1 \citep{huatuogpt_o1}. These models are 
specialized for medical or biomedical use, but differ in emphasis, ranging from 
therapeutics-oriented modeling and clinically adapted LLMs to textbook-enhanced 
medical reasoning and verifier-guided complex medical reasoning.

The second group comprises open-source reasoning models, including 
DeepSeek-R1-Distill-\allowbreak Qwen-32B and 
DeepSeek-R1-Distill-\allowbreak Llama-70B \citep{guo2025deepseek}, Marco-o1 
\citep{marco_o1}, QwQ-32B \citep{qwq32b}, Sky-T1-32B-\allowbreak Flash 
\citep{sky_t1}, and Qwen3-8B \citep{yang2025qwen3}. These models emphasize general 
reasoning, but adopt different strategies, including distillation from strong 
reasoning teachers, chain-of-thought fine-tuning with reflection and search, 
reinforcement-learning-based reasoning enhancement, and support for switching 
between thinking and non-thinking modes.

\subsection{Frontier API Models}
To provide a more comprehensive assessment of representative frontier
models, we further evaluate nine API-served frontier models, both proprietary and open-weight, spanning six
major vendors. From \textbf{OpenAI}, we evaluate GPT-5.4
and GPT-5.2,\footnote{\url{https://platform.openai.com/docs/models}} two high-performing frontier models with
strong reasoning and instruction-following capabilities. From
\textbf{Google}, we evaluate Gemini-3.1-Pro,\footnote{\url{https://ai.google.dev/gemini-api/docs/models}}
which represents a frontier multimodal reasoning model with strong
long-context understanding. From \textbf{Anthropic}, we evaluate
Claude-Opus-4.5,\footnote{\url{https://docs.anthropic.com/en/docs/about-claude/models}} a frontier model
optimized for complex analysis and nuanced reasoning. From
\textbf{DeepSeek}, we evaluate DeepSeek-R1~\citep{guo2025deepseek}, a
reasoning-specialized model trained with reinforcement learning for complex
multi-step reasoning; DeepSeek-V3.2 and
DeepSeek-V3.1,\footnote{\url{https://api-docs.deepseek.com/news/news250324}} high-performance
open-weight models with strong reasoning benchmark results. From
\textbf{Alibaba}, we evaluate Qwen3-Max~\citep{alibabacloud2026qwen3max}, a Qwen3-series model with
strong multilingual and reasoning capabilities. From
\textbf{Zhipu AI}, we evaluate GLM-4.7,\footnote{\url{https://open.bigmodel.cn/dev/howuse/model}} a competitive
general-purpose model with strong instruction-following and reasoning
capabilities. All nine models are evaluated via API in a generative
setting: each model receives the medical question and numbered reasoning steps,
and is prompted to output a JSON array of ``+'' (correct) or ``-'' (erroneous)
symbols, one per step. This generative evaluation protocol differs from the
probability-based protocol used for PRM-style models, as API-based models do
not expose token-level probabilities.

\section{Additional Experimental Results and Protocols}
\label{app:additional_experiments}

\subsection{Direct Medical QA Evaluation}
\label{sec:medical_qa_eval}

Beyond step-level reasoning verification, we further evaluate 22 standalone models (9 served via API and 13 run locally) on direct medical question answering using MedQA \citep{jin2021disease} and MedMCQA \citep{pal2022medmcqa}, the same source datasets from which MedPRMBench is constructed. This experiment contextualizes each policy model's direct answer ability and is not used to rank PRMs. Closed-source models are evaluated via API with greedy decoding (temperature = 0); open-source models are evaluated using vLLM with max\_new\_tokens = 2048 to accommodate thinking models. All standalone models receive a zero-shot prompt instructing them to act as a medical expert and output only the answer letter (A/B/C/D). We report Accuracy (Acc) and Macro-F1 across the four answer options. Verifier-enhanced Llama policy results are reported separately in Table~\ref{tab:prm_verifier}.

\begin{table}[htbp]
\centering
\caption{Zero-shot performance of standalone LLMs on medical QA benchmarks. Acc: Accuracy (\%), F1: Macro-F1 (\%). Among locally run standalone models, \textbf{bold} indicates best and \underline{underline} indicates second best. Policy rows here report direct single-pass QA; verifier-enhanced multi-sample Llama results are reported separately in Table~\ref{tab:prm_verifier}.}
\label{tab:medical_qa_eval}
\footnotesize
\begin{tabular}{lcccc}
\toprule
\multirow{2}{*}{\textbf{Model}} & \multicolumn{2}{c}{\textbf{MedMCQA}} & \multicolumn{2}{c}{\textbf{MedQA}} \\
\cmidrule(lr){2-3} \cmidrule(lr){4-5}
 & Acc & F1 & Acc & F1 \\
\midrule
GPT-5.4 & 79.11 & 79.00 & 92.07 & 92.05 \\
DeepSeek-R1 & 78.44 & 78.52 & 89.79 & 89.97 \\
DeepSeek-V3.1 & 78.96 & 78.69 & 89.55 & 89.46 \\
Claude Opus 4.5 & 78.68 & 79.92 & 87.43 & 89.85 \\
Qwen3-Max & 74.68 & 75.04 & 85.70 & 85.48 \\
Gemini-3.1-Pro & 70.19 & 77.12 & 77.06 & 84.84 \\
DeepSeek-V3.2 & 68.75 & 70.12 & 76.90 & 78.11 \\
GLM-4.7 & 56.35 & 66.24 & 64.18 & 76.18 \\
GPT-5.2 & 56.30 & 65.08 & 60.72 & 72.45 \\
\midrule
\midrule
DS-R1-Distill-Llama-70B & \textbf{72.34} & \textbf{72.64} & \textbf{85.39} & \textbf{86.92} \\
QwQ-32B & \underline{67.73} & \underline{67.74} & \underline{81.15} & \underline{81.65} \\
DS-R1-Distill-Qwen-32B & 65.07 & 65.36 & 77.69 & 79.23 \\
Sky-T1-32B-Flash & 65.22 & 64.83 & 69.60 & 69.27 \\
Qwen3-8B & 62.90 & 62.80 & 74.23 & 75.20 \\
\textbf{Llama-3.1-8B-Instruct} & 58.69 & 58.83 & 69.52 & 69.82 \\
MedSSS-Policy & 51.18 & 51.06 & 57.50 & 57.62 \\
HuatuoGPT-o1-8B & 54.36 & 54.70 & 70.78 & 71.06 \\
Meditron3-8B & 60.22 & 60.03 & 67.16 & 67.07 \\
UltraMedical-8B & 56.49 & 55.68 & 61.19 & 61.10 \\
Meerkat-8B & 58.26 & 58.11 & 39.67 & 37.27 \\
TX-Gemma-9B & 49.27 & 48.97 & 52.87 & 52.68 \\
Marco-o1 & 51.54 & 50.76 & 54.91 & 54.36 \\
\bottomrule
\end{tabular}
\end{table}

\paragraph{Analysis.}
As shown in Table~\ref{tab:medical_qa_eval}, GPT-5.4 achieves the highest overall accuracy on MedQA (92.07\%), followed by DeepSeek-R1 (89.79\%), DeepSeek-V3.1 (89.55\%), and Claude Opus 4.5 (87.43\%). On MedMCQA, GPT-5.4 (79.11\%), DeepSeek-V3.1 (78.96\%), Claude Opus 4.5 (78.68\%), and DeepSeek-R1 (78.44\%) perform comparably at the top. All models achieve substantially higher accuracy on MedQA than on MedMCQA, likely because MedQA's USMLE-style clinical vignettes are well-represented in training corpora, whereas MedMCQA covers a broader range of medical subjects including basic sciences.

Among locally run standalone models, DS-R1-Distill-Llama-70B leads with 85.39\% on MedQA and 72.34\% on MedMCQA, followed by QwQ-32B (81.15\%/67.73\%) and DS-R1-Distill-Qwen-32B (77.69\%/65.07\%). The best open-source standalone model trails the best closed-source model by 6.68 pp on MedQA and 6.77 pp on MedMCQA. The direct single-pass policy results remain below the strongest open-source reasoning models: Llama-3.1-8B-Instruct obtains 69.52\% on MedQA and 58.69\% on MedMCQA, while MedSSS-Policy~\citep{jiang2025meds3} obtains 57.50\% and 51.18\%, respectively. Among medical-specialized open-source models, HuatuoGPT-o1-8B (70.78\% on MedQA) outperforms Meditron3-8B (67.16\%) and UltraMedical-8B (61.19\%), though all remain below general-purpose reasoning models of comparable size such as Qwen3-8B (74.23\%). We caution that this strict letter-only protocol disadvantages models tuned toward explanatory chat formats, so cross-protocol conclusions about medical fine-tuning should be drawn with care. We retain MedSSS-Policy only in this direct QA comparison because it is a released policy model rather than a standalone PRM checkpoint.

Verifier-enhanced Llama policy results are reported separately in Table~\ref{tab:prm_verifier}, where MedPRMBench-RM is used to rerank or aggregate multiple candidate reasoning traces.


\subsection{Hard Subset Analysis}
\label{sec:hard_subset}

To further stress-test model capabilities on the most challenging cases, we construct a \textbf{Hard Subset} of 700 samples from the full MedPRMBench test set (6{,}500 samples) using the following four filtering criteria:
\textbf{(1)~No step overlap}: each error step corresponds to exactly one error type, preventing models from exploiting error co-occurrence;
\textbf{(2)~Zero pass rate}: GPT-4o-mini answers incorrectly in all $K{=}8$ difficulty-probe samples of the original question;
\textbf{(3)~Answer unchanged}: the injected error does not alter the final answer, making the error more covert and harder to detect via outcome-based shortcuts;
\textbf{(4)~Low severity score}: samples are ranked by the rule-derived severity score in ascending order, and the first 700 are selected.
This operational challenge slice contains 10.8\% of the benchmark. Its low rule-derived severity scores are a selection criterion, not an independently validated measure of clinical subtlety.

\begin{figure}[!htb]
\centering
\includegraphics[width=0.7\textwidth]{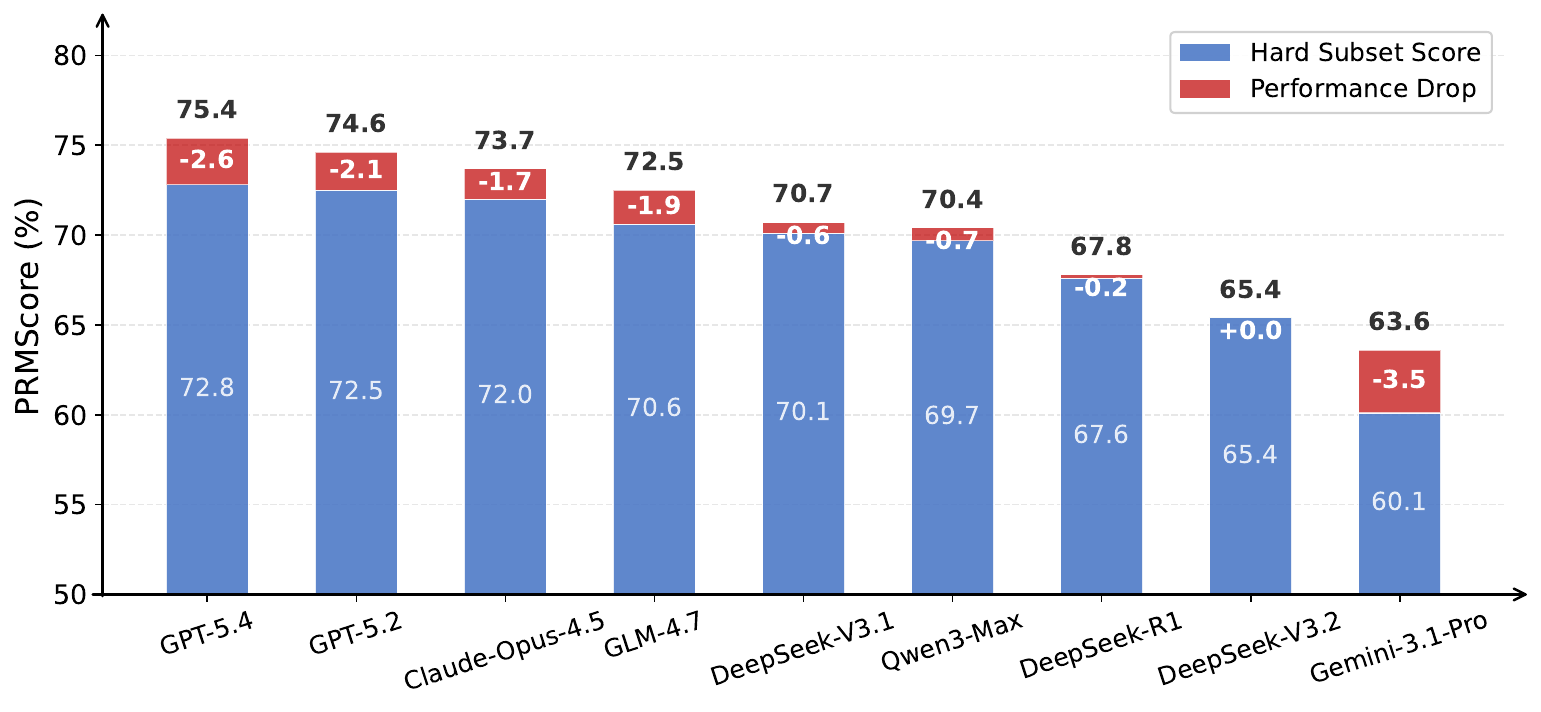}
\caption{PRMScore (\%) comparison between the full MedPRMBench test set and the Hard Subset (700 samples) for nine frontier API models. Blue bars represent the Hard Subset score; red segments represent the performance drop ($\Delta$). Eight of nine models score lower on the Hard Subset (DeepSeek-V3.2 is unchanged), with an average decline of $-$1.5 pp.}
\label{fig:hard_subset}
\end{figure}

\noindent\textbf{The Hard Subset exposes additional difficulty for frontier API models.}
As shown in Figure~\ref{fig:hard_subset}, eight of nine frontier API models score lower on the Hard Subset, with a mean decline of 1.5 points. GPT-5.4 drops from 75.4\% to 72.8\%, GPT-5.2 from 74.6\% to 72.5\%, and Claude-Opus-4.5 from 73.7\% to 72.0\%; Gemini-3.1-Pro has the largest decline (3.5 points), whereas DeepSeek-R1 ($-$0.2) and DeepSeek-V3.2 ($+$0.0) are essentially unaffected. The subset thus serves as a model-dependent stress test under low rule-derived severity, not a uniformly harder split.

\clearpage
\subsection{Detailed Results by Error Type}
\label{app:error_type_details}

Tables~\ref{tab:error_type_S_1}--\ref{tab:error_type_E_5} present detailed step-level diagnostics for the main critic models, the released Med-PRM-Reward v1.0 baseline, and our benchmark-specific medical PRM on each of the 14 error types defined in MedPRMBench. For each error type, we report PRMScore ($0.5 \times F1_{\mathrm{err}} + 0.5 \times F1_{\mathrm{cor}}$), F1 (with erroneous steps as the positive class), Negative F1 (with correct steps as the positive class), step-level Accuracy, Positive Accuracy (fraction of correct steps correctly identified), Negative Accuracy (error detection rate), and First Error Accuracy (whether the gold first-error step is detected).

\begin{table}[!htbp]
\centering
\caption{A performance comparison of popular models across detailed metrics in \textbf{S-1 (Non-Redundancy)} sub-category of MedPRMBench. Category: Simplicity. The best performance for each metric is highlighted in \textbf{bold}, while the second-best performance is \underline{underlined}.}
\label{tab:error_type_S_1}
\resizebox{0.8\textwidth}{!}{
\begin{tabular}{lccccccc}
\toprule
\textbf{Model Name} & \textbf{PRMScore} & \textbf{F1} & \textbf{Negative F1} & \textbf{Acc} & \textbf{Positive Acc} & \textbf{Negative Acc} & \textbf{First} \\
\midrule
\rowcolor{gray!15} \multicolumn{8}{c}{\textbf{\emph{Open-source Medical Models, Prompted as Critic Models}}} \\
txgemma-9b-chat & 55.6 & 43.8 & 67.4 & 58.7 & 71.9 & 39.5 & 39.8 \\
Llama-3.1-8B-UltraMedical & 58.1 & 47.4 & 68.8 & 60.8 & 72.8 & 43.4 & 28.7 \\
llama-3-meerkat-8b & 55.9 & 44.2 & 67.7 & 59.1 & 74.3 & 38.3 & 24.1 \\
HuatuoGPT-o1-8B & 57.7 & 45.2 & 70.1 & 61.3 & 78.7 & 37.7 & 22.3 \\
Meditron3-8B & 43.6 & 12.5 & \underline{74.7} & 60.7 & \textbf{97.7} & 6.9 & 2.6 \\
\textbf{Avg.} & 54.2 & 38.6 & 69.7 & 60.1 & 79.1 & 33.2 & 23.5 \\
\midrule
\rowcolor{gray!15} \multicolumn{8}{c}{\textbf{\emph{Open-source Reasoning Models, Prompted as Critic Models}}} \\
Sky-T1-32B-Flash & 61.7 & 55.9 & 67.5 & 62.6 & 65.5 & 58.3 & 53.7 \\
QwQ-32B & 40.1 & 54.8 & 25.3 & 43.7 & 16.6 & \underline{80.6} & \underline{82.6} \\
Marco-o1 & 49.3 & 25.3 & 73.2 & 60.6 & 90.9 & 16.4 & 16.7 \\
DeepSeek-R1-Distill-Llama-70B & 57.2 & 42.4 & 72.0 & 62.3 & 84.0 & 32.8 & 9.5 \\
Qwen3-8B & 54.4 & 47.6 & 61.3 & 55.5 & 61.1 & 47.8 & 44.5 \\
DeepSeek-R1-Distill-Qwen-32B & 48.9 & 52.6 & 45.3 & 49.2 & 36.5 & 66.5 & 63.9 \\
\textbf{Avg.} & 51.9 & 46.4 & 57.4 & 55.6 & 59.1 & 50.4 & 45.1 \\
\midrule
\rowcolor{gray!15} \multicolumn{8}{c}{\textbf{\emph{Frontier API Models, Prompted as Critic Models}}} \\
Claude-Opus-4.5 & \underline{69.0} & \underline{64.7} & 73.3 & \underline{69.6} & 70.4 & 68.5 & 70.9 \\
DeepSeek-V3.2 & 63.4 & 61.6 & 65.2 & 63.5 & 57.7 & 72.0 & 74.4 \\
GPT-5.4 & 68.0 & 61.5 & 74.4 & 69.3 & 75.4 & 60.3 & 69.3 \\
GPT-5.2 & 66.3 & 59.1 & 73.5 & 67.9 & 75.2 & 57.1 & 67.0 \\
GLM-4.7 & 66.6 & 60.0 & 73.1 & 67.9 & 73.8 & 59.3 & 63.6 \\
Qwen3-Max & 64.3 & 57.0 & 71.6 & 65.8 & 72.7 & 55.8 & 63.6 \\
DeepSeek-V3.1 & 64.0 & 55.5 & 72.5 & 66.0 & 75.5 & 52.2 & 57.1 \\
DeepSeek-R1 & 63.7 & 55.7 & 71.7 & 65.5 & 73.8 & 53.3 & 54.6 \\
Gemini-3.1-Pro & 55.4 & 37.3 & 73.5 & 62.8 & 87.1 & 27.2 & 57.5 \\
\textbf{Avg.} & 64.5 & 56.9 & 72.1 & 66.5 & 73.5 & 56.2 & 64.2 \\
\midrule
\rowcolor{gray!15} \multicolumn{8}{c}{\textbf{\emph{Medical Process Reward Models}}} \\
Med-PRM-Reward v1.0 & 57.1 & 61.7 & 52.5 & 57.6 & 40.7 & \textbf{80.7} & 65.1 \\
Ours & \textbf{87.5} & \textbf{84.9} & \textbf{90.2} & \textbf{88.1} & \underline{95.0} & 78.7 & \textbf{92.4} \\
\bottomrule
\end{tabular}}
\end{table}

\begin{table}[!htbp]
\centering
\caption{A performance comparison of popular models across detailed metrics in \textbf{S-2 (Non-Circular Logic)} sub-category of MedPRMBench. Category: Simplicity. The best performance for each metric is highlighted in \textbf{bold}, while the second-best performance is \underline{underlined}.}
\label{tab:error_type_S_2}
\resizebox{0.8\textwidth}{!}{
\begin{tabular}{lccccccc}
\toprule
\textbf{Model Name} & \textbf{PRMScore} & \textbf{F1} & \textbf{Negative F1} & \textbf{Acc} & \textbf{Positive Acc} & \textbf{Negative Acc} & \textbf{First} \\
\midrule
\rowcolor{gray!15} \multicolumn{8}{c}{\textbf{\emph{Open-source Medical Models, Prompted as Critic Models}}} \\
txgemma-9b-chat & 57.9 & 47.5 & 68.2 & 60.4 & 69.2 & 46.4 & 47.6 \\
Llama-3.1-8B-UltraMedical & 55.5 & 43.7 & 67.3 & 58.6 & 69.3 & 41.6 & 24.5 \\
llama-3-meerkat-8b & 53.9 & 40.4 & 67.4 & 57.8 & 70.2 & 37.7 & 19.3 \\
HuatuoGPT-o1-8B & 57.0 & 43.4 & 70.7 & 61.4 & 75.1 & 38.9 & 24.5 \\
Meditron3-8B & 47.0 & 17.9 & 76.1 & 62.9 & \underline{95.9} & 10.5 & 4.5 \\
\textbf{Avg.} & 54.3 & 38.6 & 69.9 & 60.2 & 75.9 & 35.0 & 24.1 \\
\midrule
\rowcolor{gray!15} \multicolumn{8}{c}{\textbf{\emph{Open-source Reasoning Models, Prompted as Critic Models}}} \\
Sky-T1-32B-Flash & 65.8 & 61.1 & 70.5 & 66.4 & 65.3 & 68.3 & 64.2 \\
QwQ-32B & 40.6 & 52.8 & 28.4 & 43.1 & 18.2 & \textbf{83.8} & \underline{89.1} \\
Marco-o1 & 55.1 & 35.7 & 74.6 & 63.6 & 87.2 & 26.1 & 26.9 \\
DeepSeek-R1-Distill-Llama-70B & 58.8 & 44.0 & 73.6 & 64.1 & 80.6 & 37.1 & 12.4 \\
Qwen3-8B & 57.8 & 52.0 & 63.6 & 58.6 & 58.3 & 59.1 & 53.9 \\
DeepSeek-R1-Distill-Qwen-32B & 48.4 & 52.5 & 44.3 & 48.7 & 32.9 & 74.7 & 73.5 \\
\textbf{Avg.} & 54.4 & 49.7 & 59.2 & 57.4 & 57.1 & 58.2 & 53.3 \\
\midrule
\rowcolor{gray!15} \multicolumn{8}{c}{\textbf{\emph{Frontier API Models, Prompted as Critic Models}}} \\
Claude-Opus-4.5 & 73.6 & 70.1 & 77.1 & 74.0 & 71.1 & 78.7 & 75.9 \\
DeepSeek-V3.2 & 65.1 & 63.5 & 66.6 & 65.1 & 56.6 & 78.7 & 79.4 \\
GPT-5.4 & \underline{75.4} & \underline{70.9} & \underline{79.9} & \underline{76.2} & 76.9 & 75.1 & 72.6 \\
GPT-5.2 & 75.0 & 70.4 & 79.7 & 75.9 & 76.9 & 74.3 & 73.6 \\
GLM-4.7 & 72.5 & 67.8 & 77.1 & 73.3 & 73.4 & 73.0 & 68.4 \\
Qwen3-Max & 69.5 & 63.5 & 75.5 & 70.7 & 73.7 & 65.9 & 66.1 \\
DeepSeek-V3.1 & 70.5 & 64.1 & 77.0 & 71.9 & 76.4 & 64.9 & 62.5 \\
DeepSeek-R1 & 66.8 & 59.2 & 74.5 & 68.6 & 74.7 & 58.9 & 55.8 \\
Gemini-3.1-Pro & 64.8 & 52.9 & 76.7 & 68.9 & 83.7 & 45.3 & 67.4 \\
\textbf{Avg.} & 70.4 & 64.7 & 76.0 & 71.6 & 73.7 & 68.3 & 69.1 \\
\midrule
\rowcolor{gray!15} \multicolumn{8}{c}{\textbf{\emph{Medical Process Reward Models}}} \\
Med-PRM-Reward v1.0 & 55.6 & 58.0 & 53.2 & 55.7 & 40.5 & \underline{80.5} & 66.5 \\
Ours & \textbf{88.1} & \textbf{84.4} & \textbf{91.8} & \textbf{89.2} & \textbf{96.8} & 76.7 & \textbf{91.0} \\
\bottomrule
\end{tabular}}
\end{table}

\begin{table}[!htbp]
\centering
\caption{A performance comparison of popular models across detailed metrics in \textbf{R-1 (Evidence-Based Soundness)} sub-category of MedPRMBench. Category: Soundness. The best performance for each metric is highlighted in \textbf{bold}, while the second-best performance is \underline{underlined}.}
\label{tab:error_type_R_1}
\resizebox{0.8\textwidth}{!}{
\begin{tabular}{lccccccc}
\toprule
\textbf{Model Name} & \textbf{PRMScore} & \textbf{F1} & \textbf{Negative F1} & \textbf{Acc} & \textbf{Positive Acc} & \textbf{Negative Acc} & \textbf{First} \\
\midrule
\rowcolor{gray!15} \multicolumn{8}{c}{\textbf{\emph{Open-source Medical Models, Prompted as Critic Models}}} \\
txgemma-9b-chat & 57.2 & 46.0 & 68.3 & 60.1 & 70.8 & 43.5 & 44.0 \\
Llama-3.1-8B-UltraMedical & 56.1 & 44.5 & 67.7 & 59.1 & 70.2 & 41.9 & 28.0 \\
llama-3-meerkat-8b & 54.8 & 42.5 & 67.1 & 58.2 & 72.3 & 37.7 & 22.0 \\
HuatuoGPT-o1-8B & 57.8 & 45.8 & 69.8 & 61.2 & 75.8 & 40.1 & 25.4 \\
Meditron3-8B & 45.9 & 16.1 & 75.7 & 62.3 & \textbf{96.4} & 9.2 & 4.4 \\
\textbf{Avg.} & 54.4 & 39.0 & 69.7 & 60.2 & 77.1 & 34.5 & 24.8 \\
\midrule
\rowcolor{gray!15} \multicolumn{8}{c}{\textbf{\emph{Open-source Reasoning Models, Prompted as Critic Models}}} \\
Sky-T1-32B-Flash & 65.5 & 61.9 & 69.1 & 65.9 & 62.6 & 71.0 & 68.5 \\
QwQ-32B & 40.1 & 55.8 & 24.4 & 44.2 & 15.3 & \textbf{86.0} & \underline{90.4} \\
Marco-o1 & 53.8 & 33.4 & 74.3 & 62.9 & 88.1 & 23.7 & 24.2 \\
DeepSeek-R1-Distill-Llama-70B & 59.8 & 47.6 & 71.9 & 63.5 & 79.3 & 40.6 & 14.5 \\
Qwen3-8B & 58.1 & 54.2 & 62.1 & 58.5 & 57.4 & 60.1 & 57.3 \\
DeepSeek-R1-Distill-Qwen-32B & 48.9 & 54.9 & 42.9 & 49.6 & 32.1 & 74.9 & 72.3 \\
\textbf{Avg.} & 54.4 & 51.3 & 57.4 & 57.4 & 55.8 & 59.4 & 54.5 \\
\midrule
\rowcolor{gray!15} \multicolumn{8}{c}{\textbf{\emph{Frontier API Models, Prompted as Critic Models}}} \\
Claude-Opus-4.5 & 73.5 & 71.4 & 75.5 & 73.6 & 66.8 & 84.2 & 84.9 \\
DeepSeek-V3.2 & 65.1 & 64.7 & 65.6 & 65.1 & 54.6 & 81.5 & 83.3 \\
GPT-5.4 & \underline{75.9} & \underline{72.9} & \underline{79.0} & \underline{76.3} & 73.1 & 81.4 & 85.0 \\
GPT-5.2 & 75.3 & 71.9 & 78.7 & 75.7 & 73.5 & 79.2 & 82.8 \\
GLM-4.7 & 72.6 & 69.2 & 76.1 & 73.1 & 70.3 & 77.3 & 78.3 \\
Qwen3-Max & 70.3 & 65.7 & 74.9 & 71.0 & 71.0 & 71.0 & 75.3 \\
DeepSeek-V3.1 & 71.0 & 65.9 & 76.0 & 71.8 & 73.3 & 69.6 & 69.9 \\
DeepSeek-R1 & 67.5 & 61.2 & 73.8 & 68.7 & 72.4 & 63.0 & 64.3 \\
Gemini-3.1-Pro & 64.0 & 52.1 & 76.0 & 68.0 & 83.2 & 44.4 & 71.6 \\
\textbf{Avg.} & 70.6 & 66.1 & 75.1 & 71.5 & 70.9 & 72.4 & 77.3 \\
\midrule
\rowcolor{gray!15} \multicolumn{8}{c}{\textbf{\emph{Medical Process Reward Models}}} \\
Med-PRM-Reward v1.0 & 58.5 & 62.7 & 54.3 & 58.9 & 41.4 & \underline{84.3} & 70.1 \\
Ours & \textbf{85.2} & \textbf{81.3} & \textbf{89.1} & \textbf{86.2} & \underline{95.3} & 73.2 & \textbf{92.6} \\
\bottomrule
\end{tabular}}
\end{table}

\begin{table}[!htbp]
\centering
\caption{A performance comparison of popular models across detailed metrics in \textbf{R-2 (Step Consistency)} sub-category of MedPRMBench. Category: Soundness. The best performance for each metric is highlighted in \textbf{bold}, while the second-best performance is \underline{underlined}.}
\label{tab:error_type_R_2}
\resizebox{0.8\textwidth}{!}{
\begin{tabular}{lccccccc}
\toprule
\textbf{Model Name} & \textbf{PRMScore} & \textbf{F1} & \textbf{Negative F1} & \textbf{Acc} & \textbf{Positive Acc} & \textbf{Negative Acc} & \textbf{First} \\
\midrule
\rowcolor{gray!15} \multicolumn{8}{c}{\textbf{\emph{Open-source Medical Models, Prompted as Critic Models}}} \\
txgemma-9b-chat & 58.0 & 46.7 & 69.2 & 61.0 & 71.0 & 44.8 & 46.4 \\
Llama-3.1-8B-UltraMedical & 56.2 & 44.6 & 67.8 & 59.3 & 69.5 & 42.9 & 27.2 \\
llama-3-meerkat-8b & 55.5 & 42.8 & 68.2 & 59.2 & 72.4 & 38.7 & 21.7 \\
HuatuoGPT-o1-8B & 58.0 & 45.8 & 70.2 & 61.6 & 74.8 & 41.2 & 26.6 \\
Meditron3-8B & 46.2 & 16.0 & 76.3 & 63.0 & \textbf{96.4} & 9.2 & 4.6 \\
\textbf{Avg.} & 54.8 & 39.2 & 70.3 & 60.8 & 76.8 & 35.4 & 25.3 \\
\midrule
\rowcolor{gray!15} \multicolumn{8}{c}{\textbf{\emph{Open-source Reasoning Models, Prompted as Critic Models}}} \\
Sky-T1-32B-Flash & 65.4 & 61.8 & 69.0 & 65.7 & 61.7 & 72.3 & 69.4 \\
QwQ-32B & 39.4 & 54.4 & 24.4 & 43.1 & 15.2 & \textbf{86.1} & \underline{90.1} \\
Marco-o1 & 55.7 & 36.3 & 75.1 & 64.2 & 87.4 & 26.7 & 28.2 \\
DeepSeek-R1-Distill-Llama-70B & 60.7 & 47.8 & 73.5 & 64.9 & 80.5 & 40.8 & 13.8 \\
Qwen3-8B & 59.0 & 54.9 & 63.0 & 59.4 & 57.1 & 62.8 & 60.4 \\
DeepSeek-R1-Distill-Qwen-32B & 46.9 & 53.7 & 40.1 & 47.8 & 28.9 & 76.9 & 74.3 \\
\textbf{Avg.} & 54.5 & 51.5 & 57.5 & 57.5 & 55.1 & 60.9 & 56.0 \\
\midrule
\rowcolor{gray!15} \multicolumn{8}{c}{\textbf{\emph{Frontier API Models, Prompted as Critic Models}}} \\
Claude-Opus-4.5 & 73.8 & 71.1 & 76.5 & 74.1 & 68.3 & 83.3 & 83.4 \\
DeepSeek-V3.2 & 65.0 & 64.2 & 65.8 & 65.0 & 54.5 & 82.1 & 83.6 \\
GPT-5.4 & \underline{76.1} & \underline{72.3} & \underline{79.9} & \underline{76.7} & 75.1 & 79.5 & 81.3 \\
GPT-5.2 & 75.5 & 71.3 & 79.6 & 76.2 & 75.4 & 77.5 & 80.7 \\
GLM-4.7 & 72.7 & 68.8 & 76.6 & 73.3 & 70.8 & 77.2 & 77.6 \\
Qwen3-Max & 71.1 & 66.4 & 75.8 & 71.8 & 71.3 & 72.7 & 75.5 \\
DeepSeek-V3.1 & 71.6 & 66.4 & 76.9 & 72.6 & 73.7 & 70.8 & 71.5 \\
DeepSeek-R1 & 67.9 & 61.7 & 74.0 & 69.0 & 71.5 & 65.1 & 64.8 \\
Gemini-3.1-Pro & 64.9 & 52.6 & 77.3 & 69.3 & 84.5 & 44.6 & 70.3 \\
\textbf{Avg.} & 71.0 & 66.1 & 75.8 & 72.0 & 71.7 & 72.5 & 76.5 \\
\midrule
\rowcolor{gray!15} \multicolumn{8}{c}{\textbf{\emph{Medical Process Reward Models}}} \\
Med-PRM-Reward v1.0 & 58.5 & 61.9 & 55.0 & 58.8 & 41.6 & \underline{85.1} & 69.7 \\
Ours & \textbf{84.8} & \textbf{80.2} & \textbf{89.3} & \textbf{86.1} & \underline{95.6} & 71.5 & \textbf{91.7} \\
\bottomrule
\end{tabular}}
\end{table}

\begin{table}[!htbp]
\centering
\caption{A performance comparison of popular models across detailed metrics in \textbf{R-3 (Contextual Applicability)} sub-category of MedPRMBench. Category: Soundness. The best performance for each metric is highlighted in \textbf{bold}, while the second-best performance is \underline{underlined}.}
\label{tab:error_type_R_3}
\resizebox{0.8\textwidth}{!}{
\begin{tabular}{lccccccc}
\toprule
\textbf{Model Name} & \textbf{PRMScore} & \textbf{F1} & \textbf{Negative F1} & \textbf{Acc} & \textbf{Positive Acc} & \textbf{Negative Acc} & \textbf{First} \\
\midrule
\rowcolor{gray!15} \multicolumn{8}{c}{\textbf{\emph{Open-source Medical Models, Prompted as Critic Models}}} \\
txgemma-9b-chat & 57.8 & 45.7 & 70.0 & 61.3 & 69.7 & 46.0 & 48.6 \\
Llama-3.1-8B-UltraMedical & 55.7 & 42.7 & 68.8 & 59.6 & 68.8 & 42.7 & 26.3 \\
llama-3-meerkat-8b & 55.2 & 41.3 & 69.0 & 59.4 & 71.8 & 38.4 & 21.4 \\
HuatuoGPT-o1-8B & 58.3 & 44.4 & 72.2 & 63.0 & 76.5 & 39.9 & 25.8 \\
Meditron3-8B & 48.4 & 18.0 & 78.7 & 66.2 & \textbf{96.7} & 10.5 & 4.3 \\
\textbf{Avg.} & 55.1 & 38.4 & 71.7 & 61.9 & 76.7 & 35.5 & 25.3 \\
\midrule
\rowcolor{gray!15} \multicolumn{8}{c}{\textbf{\emph{Open-source Reasoning Models, Prompted as Critic Models}}} \\
Sky-T1-32B-Flash & 66.7 & 61.6 & 71.9 & 67.5 & 64.2 & 73.6 & 70.9 \\
QwQ-32B & 38.7 & 52.5 & 24.9 & 41.8 & 15.4 & \textbf{86.7} & \underline{90.5} \\
Marco-o1 & 55.2 & 33.8 & 76.6 & 65.5 & 87.6 & 25.0 & 24.1 \\
DeepSeek-R1-Distill-Llama-70B & 59.4 & 45.5 & 73.4 & 64.2 & 78.3 & 40.2 & 14.0 \\
Qwen3-8B & 57.5 & 52.6 & 62.4 & 58.1 & 55.2 & 62.9 & 60.3 \\
DeepSeek-R1-Distill-Qwen-32B & 45.6 & 51.8 & 39.5 & 46.3 & 27.8 & 77.8 & 75.0 \\
\textbf{Avg.} & 53.9 & 49.6 & 58.1 & 57.2 & 54.8 & 61.0 & 55.8 \\
\midrule
\rowcolor{gray!15} \multicolumn{8}{c}{\textbf{\emph{Frontier API Models, Prompted as Critic Models}}} \\
Claude-Opus-4.5 & 75.2 & 71.0 & 79.4 & 75.9 & 71.7 & 83.6 & 84.6 \\
DeepSeek-V3.2 & 66.1 & 63.5 & 68.8 & 66.3 & 57.3 & 82.9 & 84.5 \\
GPT-5.4 & \underline{76.8} & \underline{72.2} & \underline{81.5} & \underline{77.8} & 75.7 & 81.6 & 84.8 \\
GPT-5.2 & 76.3 & 71.3 & 81.3 & 77.4 & 76.2 & 79.5 & 83.7 \\
GLM-4.7 & 73.9 & 68.7 & 79.1 & 75.0 & 73.5 & 77.7 & 79.1 \\
Qwen3-Max & 71.9 & 65.9 & 77.8 & 73.2 & 73.0 & 73.5 & 76.5 \\
DeepSeek-V3.1 & 71.9 & 65.4 & 78.4 & 73.4 & 74.7 & 71.1 & 71.8 \\
DeepSeek-R1 & 69.6 & 62.2 & 77.0 & 71.4 & 73.9 & 66.7 & 67.3 \\
Gemini-3.1-Pro & 65.6 & 51.5 & 79.7 & 71.4 & 86.9 & 43.0 & 67.6 \\
\textbf{Avg.} & 71.9 & 65.7 & 78.1 & 73.5 & 73.7 & 73.3 & 77.8 \\
\midrule
\rowcolor{gray!15} \multicolumn{8}{c}{\textbf{\emph{Medical Process Reward Models}}} \\
Med-PRM-Reward v1.0 & 55.7 & 58.9 & 52.6 & 56.0 & 38.9 & \underline{85.0} & 67.9 \\
Ours & \textbf{86.8} & \textbf{82.5} & \textbf{91.2} & \textbf{88.3} & \underline{96.4} & 74.5 & \textbf{92.1} \\
\bottomrule
\end{tabular}}
\end{table}

\begin{table}[!htbp]
\centering
\caption{A performance comparison of popular models across detailed metrics in \textbf{R-4 (Confidence Invariance)} sub-category of MedPRMBench. Category: Soundness. The best performance for each metric is highlighted in \textbf{bold}, while the second-best performance is \underline{underlined}.}
\label{tab:error_type_R_4}
\resizebox{0.8\textwidth}{!}{
\begin{tabular}{lccccccc}
\toprule
\textbf{Model Name} & \textbf{PRMScore} & \textbf{F1} & \textbf{Negative F1} & \textbf{Acc} & \textbf{Positive Acc} & \textbf{Negative Acc} & \textbf{First} \\
\midrule
\rowcolor{gray!15} \multicolumn{8}{c}{\textbf{\emph{Open-source Medical Models, Prompted as Critic Models}}} \\
txgemma-9b-chat & 58.6 & 51.1 & 66.1 & 60.0 & 64.1 & 53.5 & 63.3 \\
Llama-3.1-8B-UltraMedical & 56.5 & 47.3 & 65.8 & 58.5 & 65.6 & 47.6 & 34.3 \\
llama-3-meerkat-8b & 54.4 & 43.6 & 65.3 & 57.0 & 66.2 & 42.7 & 25.5 \\
HuatuoGPT-o1-8B & 57.9 & 47.2 & 68.6 & 60.6 & 70.5 & 45.2 & 32.1 \\
Meditron3-8B & 50.7 & 25.9 & 75.5 & 63.2 & \underline{93.3} & 16.4 & 5.8 \\
\textbf{Avg.} & 55.6 & 43.0 & 68.3 & 59.9 & 71.9 & 41.1 & 32.2 \\
\midrule
\rowcolor{gray!15} \multicolumn{8}{c}{\textbf{\emph{Open-source Reasoning Models, Prompted as Critic Models}}} \\
Sky-T1-32B-Flash & 66.0 & 65.9 & 66.2 & 66.0 & 54.6 & 83.9 & 83.5 \\
QwQ-32B & 40.2 & 55.3 & 25.1 & 44.0 & 15.4 & \underline{88.8} & \textbf{94.6} \\
Marco-o1 & 57.2 & 43.4 & 71.0 & 61.7 & 77.2 & 37.6 & 44.0 \\
DeepSeek-R1-Distill-Llama-70B & 59.9 & 51.3 & 68.6 & 61.8 & 68.3 & 51.6 & 17.9 \\
Qwen3-8B & 56.6 & 56.8 & 56.5 & 56.6 & 46.1 & 73.1 & 71.2 \\
DeepSeek-R1-Distill-Qwen-32B & 43.6 & 54.1 & 33.2 & 45.6 & 22.2 & 82.2 & 81.9 \\
\textbf{Avg.} & 53.9 & 54.5 & 53.4 & 56.0 & 47.3 & 69.5 & 65.5 \\
\midrule
\rowcolor{gray!15} \multicolumn{8}{c}{\textbf{\emph{Frontier API Models, Prompted as Critic Models}}} \\
Claude-Opus-4.5 & 72.7 & 71.2 & 74.2 & 72.8 & 64.3 & 86.0 & 88.3 \\
DeepSeek-V3.2 & 64.1 & 65.5 & 62.7 & 64.2 & 49.5 & 87.1 & 91.2 \\
GPT-5.4 & 75.1 & \underline{72.6} & 77.7 & 75.4 & 70.3 & 83.3 & 86.9 \\
GPT-5.2 & \underline{75.2} & 72.4 & \underline{78.1} & \underline{75.6} & 71.6 & 81.8 & 85.5 \\
GLM-4.7 & 72.7 & 70.5 & 74.9 & 72.9 & 66.4 & 82.9 & 85.0 \\
Qwen3-Max & 71.8 & 68.6 & 75.0 & 72.2 & 68.6 & 77.7 & 83.5 \\
DeepSeek-V3.1 & 71.2 & 68.2 & 74.1 & 71.4 & 67.0 & 78.4 & 81.6 \\
DeepSeek-R1 & 68.6 & 64.9 & 72.4 & 69.1 & 66.5 & 73.2 & 73.6 \\
Gemini-3.1-Pro & 65.5 & 53.7 & 77.3 & 69.5 & 85.1 & 45.2 & 72.6 \\
\textbf{Avg.} & 70.8 & 67.5 & 74.0 & 71.5 & 67.7 & 77.3 & 83.1 \\
\midrule
\rowcolor{gray!15} \multicolumn{8}{c}{\textbf{\emph{Medical Process Reward Models}}} \\
Med-PRM-Reward v1.0 & 56.1 & 61.8 & 50.4 & 56.9 & 36.0 & \textbf{89.6} & 70.5 \\
Ours & \textbf{86.2} & \textbf{82.3} & \textbf{90.2} & \textbf{87.4} & \textbf{95.1} & 75.3 & \underline{94.0} \\
\bottomrule
\end{tabular}}
\end{table}

\begin{table}[!htbp]
\centering
\caption{A performance comparison of popular models across detailed metrics in \textbf{R-5 (Safety Awareness)} sub-category of MedPRMBench. Category: Soundness. The best performance for each metric is highlighted in \textbf{bold}, while the second-best performance is \underline{underlined}.}
\label{tab:error_type_R_5}
\resizebox{0.8\textwidth}{!}{
\begin{tabular}{lccccccc}
\toprule
\textbf{Model Name} & \textbf{PRMScore} & \textbf{F1} & \textbf{Negative F1} & \textbf{Acc} & \textbf{Positive Acc} & \textbf{Negative Acc} & \textbf{First} \\
\midrule
\rowcolor{gray!15} \multicolumn{8}{c}{\textbf{\emph{Open-source Medical Models, Prompted as Critic Models}}} \\
txgemma-9b-chat & 58.7 & 46.3 & 71.2 & 62.5 & 70.7 & 46.8 & 47.3 \\
Llama-3.1-8B-UltraMedical & 57.2 & 43.7 & 70.8 & 61.5 & 71.2 & 43.2 & 31.9 \\
llama-3-meerkat-8b & 55.5 & 41.2 & 69.7 & 60.0 & 73.5 & 37.5 & 22.6 \\
HuatuoGPT-o1-8B & 58.4 & 44.5 & 72.2 & 63.0 & 76.9 & 39.7 & 25.7 \\
Meditron3-8B & 46.4 & 13.7 & 79.2 & 66.4 & \textbf{97.4} & 7.7 & 3.9 \\
\textbf{Avg.} & 55.2 & 37.9 & 72.6 & 62.7 & 77.9 & 35.0 & 26.3 \\
\midrule
\rowcolor{gray!15} \multicolumn{8}{c}{\textbf{\emph{Open-source Reasoning Models, Prompted as Critic Models}}} \\
Sky-T1-32B-Flash & 67.2 & 61.7 & 72.7 & 68.1 & 64.8 & 74.4 & 75.0 \\
QwQ-32B & 38.7 & 53.6 & 23.8 & 42.3 & 14.4 & \textbf{89.1} & \textbf{93.3} \\
Marco-o1 & 54.1 & 31.2 & 77.1 & 65.6 & 88.3 & 22.5 & 22.8 \\
DeepSeek-R1-Distill-Llama-70B & 60.2 & 47.1 & 73.3 & 64.5 & 77.9 & 42.1 & 15.3 \\
Qwen3-8B & 58.5 & 54.2 & 62.7 & 58.9 & 55.3 & 65.0 & 62.4 \\
DeepSeek-R1-Distill-Qwen-32B & 47.9 & 53.2 & 42.6 & 48.4 & 30.5 & 78.4 & 74.5 \\
\textbf{Avg.} & 54.4 & 50.2 & 58.7 & 58.0 & 55.2 & 61.9 & 57.2 \\
\midrule
\rowcolor{gray!15} \multicolumn{8}{c}{\textbf{\emph{Frontier API Models, Prompted as Critic Models}}} \\
Claude-Opus-4.5 & 76.5 & 72.6 & 80.5 & 77.2 & 71.9 & \underline{87.3} & 88.3 \\
DeepSeek-V3.2 & 67.7 & 64.4 & 71.0 & 68.1 & 59.9 & 83.6 & 84.6 \\
GPT-5.4 & \underline{78.0} & \underline{73.8} & \underline{82.3} & \underline{78.9} & 75.1 & 86.0 & 88.8 \\
GPT-5.2 & 77.3 & 72.7 & 81.9 & 78.3 & 75.4 & 83.8 & 87.1 \\
GLM-4.7 & 74.6 & 69.6 & 79.6 & 75.5 & 72.7 & 81.0 & 82.0 \\
Qwen3-Max & 73.6 & 67.7 & 79.5 & 74.9 & 74.3 & 76.2 & 80.6 \\
DeepSeek-V3.1 & 74.1 & 68.1 & 80.1 & 75.5 & 75.4 & 75.8 & 77.3 \\
DeepSeek-R1 & 70.4 & 63.6 & 77.1 & 71.9 & 72.1 & 71.4 & 72.3 \\
Gemini-3.1-Pro & 63.8 & 48.5 & 79.1 & 70.3 & 85.9 & 40.5 & 66.6 \\
\textbf{Avg.} & 72.9 & 66.8 & 79.0 & 74.5 & 73.6 & 76.2 & 80.8 \\
\midrule
\rowcolor{gray!15} \multicolumn{8}{c}{\textbf{\emph{Medical Process Reward Models}}} \\
Med-PRM-Reward v1.0 & 57.1 & 59.6 & 54.7 & 57.3 & 41.1 & 84.4 & 68.4 \\
Ours & \textbf{85.1} & \textbf{80.1} & \textbf{90.1} & \textbf{86.8} & \underline{95.9} & 71.4 & \underline{91.8} \\
\bottomrule
\end{tabular}}
\end{table}

\begin{table}[!htbp]
\centering
\caption{A performance comparison of popular models across detailed metrics in \textbf{R-6 (Information Grounding Compliance)} sub-category of MedPRMBench. Category: Soundness. The best performance for each metric is highlighted in \textbf{bold}, while the second-best performance is \underline{underlined}.}
\label{tab:error_type_R_6}
\resizebox{0.8\textwidth}{!}{
\begin{tabular}{lccccccc}
\toprule
\textbf{Model Name} & \textbf{PRMScore} & \textbf{F1} & \textbf{Negative F1} & \textbf{Acc} & \textbf{Positive Acc} & \textbf{Negative Acc} & \textbf{First} \\
\midrule
\rowcolor{gray!15} \multicolumn{8}{c}{\textbf{\emph{Open-source Medical Models, Prompted as Critic Models}}} \\
txgemma-9b-chat & 57.1 & 45.8 & 68.4 & 60.1 & 70.8 & 43.3 & 42.9 \\
Llama-3.1-8B-UltraMedical & 56.1 & 43.9 & 68.2 & 59.4 & 71.3 & 40.8 & 25.5 \\
llama-3-meerkat-8b & 55.0 & 44.6 & 65.3 & 57.4 & 72.5 & 38.5 & 21.6 \\
HuatuoGPT-o1-8B & 57.1 & 46.8 & 67.4 & 59.6 & 75.5 & 39.8 & 24.5 \\
Meditron3-8B & 45.9 & 16.0 & 75.9 & 62.5 & \textbf{96.7} & 9.1 & 3.7 \\
\textbf{Avg.} & 54.2 & 39.4 & 69.0 & 59.8 & 77.4 & 34.3 & 23.6 \\
\midrule
\rowcolor{gray!15} \multicolumn{8}{c}{\textbf{\emph{Open-source Reasoning Models, Prompted as Critic Models}}} \\
Sky-T1-32B-Flash & 64.1 & 59.4 & 68.8 & 64.7 & 63.9 & 66.1 & 61.8 \\
QwQ-32B & 41.3 & 58.1 & 24.5 & 46.1 & 15.8 & \underline{83.7} & \underline{89.1} \\
Marco-o1 & 53.5 & 33.5 & 73.5 & 62.1 & 86.2 & 24.4 & 23.7 \\
DeepSeek-R1-Distill-Llama-70B & 58.5 & 47.8 & 69.1 & 61.2 & 78.5 & 39.8 & 13.2 \\
Qwen3-8B & 56.3 & 54.7 & 57.9 & 56.4 & 54.2 & 59.1 & 57.2 \\
DeepSeek-R1-Distill-Qwen-32B & 47.4 & 55.9 & 39.0 & 48.8 & 29.5 & 72.8 & 72.2 \\
\textbf{Avg.} & 53.5 & 51.6 & 55.5 & 56.6 & 54.7 & 57.7 & 52.9 \\
\midrule
\rowcolor{gray!15} \multicolumn{8}{c}{\textbf{\emph{Frontier API Models, Prompted as Critic Models}}} \\
Claude-Opus-4.5 & 73.9 & 71.2 & 76.6 & 74.1 & 69.3 & 81.8 & 81.9 \\
DeepSeek-V3.2 & 64.0 & 63.4 & 64.5 & 64.0 & 53.7 & 80.0 & 83.9 \\
GPT-5.4 & \underline{75.2} & \underline{71.3} & \underline{79.1} & \underline{75.8} & 75.0 & 77.1 & 81.6 \\
GPT-5.2 & 74.3 & 70.0 & 78.5 & 75.0 & 75.1 & 74.7 & 80.0 \\
GLM-4.7 & 72.6 & 68.9 & 76.3 & 73.1 & 71.0 & 76.3 & 78.1 \\
Qwen3-Max & 69.2 & 63.8 & 74.5 & 70.1 & 71.6 & 67.7 & 71.7 \\
DeepSeek-V3.1 & 70.1 & 64.9 & 75.4 & 71.0 & 72.7 & 68.5 & 69.5 \\
DeepSeek-R1 & 66.1 & 59.4 & 72.7 & 67.3 & 71.2 & 61.3 & 62.3 \\
Gemini-3.1-Pro & 64.6 & 52.6 & 76.6 & 68.7 & 84.1 & 44.5 & 71.5 \\
\textbf{Avg.} & 70.0 & 65.1 & 74.9 & 71.0 & 71.5 & 70.2 & 75.6 \\
\midrule
\rowcolor{gray!15} \multicolumn{8}{c}{\textbf{\emph{Medical Process Reward Models}}} \\
Med-PRM-Reward v1.0 & 59.4 & 65.3 & 53.6 & 60.3 & 41.3 & \textbf{83.8} & 71.6 \\
Ours & \textbf{86.4} & \textbf{83.9} & \textbf{88.8} & \textbf{86.8} & \underline{94.3} & 77.5 & \textbf{93.2} \\
\bottomrule
\end{tabular}}
\end{table}

\begin{table}[!htbp]
\centering
\caption{A performance comparison of popular models across detailed metrics in \textbf{R-7 (Trajectory Reasoning)} sub-category of MedPRMBench. Category: Soundness. The best performance for each metric is highlighted in \textbf{bold}, while the second-best performance is \underline{underlined}.}
\label{tab:error_type_R_7}
\resizebox{0.8\textwidth}{!}{
\begin{tabular}{lccccccc}
\toprule
\textbf{Model Name} & \textbf{PRMScore} & \textbf{F1} & \textbf{Negative F1} & \textbf{Acc} & \textbf{Positive Acc} & \textbf{Negative Acc} & \textbf{First} \\
\midrule
\rowcolor{gray!15} \multicolumn{8}{c}{\textbf{\emph{Open-source Medical Models, Prompted as Critic Models}}} \\
txgemma-9b-chat & 59.0 & 50.1 & 68.0 & 61.0 & 67.4 & 50.8 & 52.5 \\
Llama-3.1-8B-UltraMedical & 57.1 & 46.5 & 67.8 & 59.8 & 68.7 & 45.4 & 27.8 \\
llama-3-meerkat-8b & 55.4 & 43.1 & 67.7 & 58.8 & 71.6 & 39.3 & 21.7 \\
HuatuoGPT-o1-8B & 58.8 & 47.4 & 70.2 & 62.0 & 74.3 & 43.3 & 28.3 \\
Meditron3-8B & 47.9 & 19.5 & 76.3 & 63.4 & \textbf{95.8} & 11.6 & 4.7 \\
\textbf{Avg.} & 55.6 & 41.3 & 70.0 & 61.0 & 75.6 & 38.1 & 27.0 \\
\midrule
\rowcolor{gray!15} \multicolumn{8}{c}{\textbf{\emph{Open-source Reasoning Models, Prompted as Critic Models}}} \\
Sky-T1-32B-Flash & 66.2 & 64.5 & 67.8 & 66.2 & 57.7 & 79.9 & 76.0 \\
QwQ-32B & 38.5 & 55.1 & 22.0 & 43.0 & 13.3 & \textbf{88.0} & \underline{91.2} \\
Marco-o1 & 55.6 & 36.4 & 74.9 & 64.0 & 87.2 & 26.8 & 27.1 \\
DeepSeek-R1-Distill-Llama-70B & 60.0 & 48.6 & 71.4 & 63.3 & 76.1 & 43.8 & 14.6 \\
Qwen3-8B & 59.1 & 55.6 & 62.7 & 59.4 & 56.5 & 63.9 & 60.6 \\
DeepSeek-R1-Distill-Qwen-32B & 47.3 & 54.9 & 39.7 & 48.4 & 28.2 & 79.2 & 74.3 \\
\textbf{Avg.} & 54.5 & 52.5 & 56.4 & 57.4 & 53.2 & 63.6 & 57.3 \\
\midrule
\rowcolor{gray!15} \multicolumn{8}{c}{\textbf{\emph{Frontier API Models, Prompted as Critic Models}}} \\
Claude-Opus-4.5 & 73.8 & 72.1 & 75.4 & 73.9 & 65.2 & \underline{87.7} & 88.2 \\
DeepSeek-V3.2 & 66.1 & 66.5 & 65.6 & 66.1 & 52.7 & 87.6 & 87.6 \\
GPT-5.4 & \underline{75.9} & \underline{73.2} & \underline{78.6} & \underline{76.2} & 71.0 & 84.4 & 87.7 \\
GPT-5.2 & 74.9 & 72.0 & 77.8 & 75.3 & 70.6 & 82.8 & 85.6 \\
GLM-4.7 & 73.5 & 70.7 & 76.2 & 73.7 & 68.2 & 82.5 & 81.7 \\
Qwen3-Max & 71.5 & 68.1 & 74.9 & 71.9 & 68.0 & 78.0 & 80.2 \\
DeepSeek-V3.1 & 71.8 & 68.0 & 75.5 & 72.3 & 69.5 & 76.7 & 74.7 \\
DeepSeek-R1 & 68.9 & 63.7 & 74.2 & 69.8 & 70.5 & 68.8 & 66.8 \\
Gemini-3.1-Pro & 64.3 & 51.8 & 76.8 & 68.7 & 84.4 & 43.6 & 70.0 \\
\textbf{Avg.} & 71.2 & 67.3 & 75.0 & 72.0 & 68.9 & 76.9 & 80.3 \\
\midrule
\rowcolor{gray!15} \multicolumn{8}{c}{\textbf{\emph{Medical Process Reward Models}}} \\
Med-PRM-Reward v1.0 & 58.7 & 62.5 & 55.0 & 59.1 & 41.4 & 85.9 & 69.9 \\
Ours & \textbf{85.3} & \textbf{81.2} & \textbf{89.5} & \textbf{86.5} & \underline{95.1} & 73.4 & \textbf{93.7} \\
\bottomrule
\end{tabular}}
\end{table}

\begin{table}[!htbp]
\centering
\caption{A performance comparison of popular models across detailed metrics in \textbf{E-1 (Prerequisite Sensitivity)} sub-category of MedPRMBench. Category: Sensitivity. The best performance for each metric is highlighted in \textbf{bold}, while the second-best performance is \underline{underlined}.}
\label{tab:error_type_E_1}
\resizebox{0.8\textwidth}{!}{
\begin{tabular}{lccccccc}
\toprule
\textbf{Model Name} & \textbf{PRMScore} & \textbf{F1} & \textbf{Negative F1} & \textbf{Acc} & \textbf{Positive Acc} & \textbf{Negative Acc} & \textbf{First} \\
\midrule
\rowcolor{gray!15} \multicolumn{8}{c}{\textbf{\emph{Open-source Medical Models, Prompted as Critic Models}}} \\
txgemma-9b-chat & 58.3 & 45.4 & 71.2 & 62.3 & 70.6 & 46.1 & 47.6 \\
Llama-3.1-8B-UltraMedical & 56.5 & 42.9 & 70.0 & 60.7 & 69.5 & 43.5 & 30.7 \\
llama-3-meerkat-8b & 55.4 & 41.3 & 69.4 & 59.8 & 72.4 & 38.3 & 22.2 \\
HuatuoGPT-o1-8B & 58.6 & 44.7 & 72.4 & 63.2 & 76.6 & 40.2 & 28.3 \\
Meditron3-8B & 48.6 & 17.3 & 79.9 & 67.7 & \textbf{97.4} & 10.0 & 5.8 \\
\textbf{Avg.} & 55.5 & 38.3 & 72.6 & 62.7 & 77.3 & 35.6 & 26.9 \\
\midrule
\rowcolor{gray!15} \multicolumn{8}{c}{\textbf{\emph{Open-source Reasoning Models, Prompted as Critic Models}}} \\
Sky-T1-32B-Flash & 66.1 & 58.9 & 73.2 & 67.6 & 67.1 & 68.5 & 66.1 \\
QwQ-32B & 39.6 & 51.8 & 27.4 & 42.1 & 17.4 & \textbf{84.4} & \textbf{88.6} \\
Marco-o1 & 55.9 & 34.7 & 77.2 & 66.1 & 86.6 & 26.5 & 26.8 \\
DeepSeek-R1-Distill-Llama-70B & 59.9 & 45.9 & 73.9 & 64.8 & 78.9 & 40.5 & 14.4 \\
Qwen3-8B & 59.3 & 53.4 & 65.3 & 60.2 & 59.4 & 61.7 & 58.9 \\
DeepSeek-R1-Distill-Qwen-32B & 47.3 & 51.0 & 43.6 & 47.6 & 32.2 & 73.9 & 70.9 \\
\textbf{Avg.} & 54.7 & 49.3 & 60.1 & 58.1 & 56.9 & 59.2 & 54.3 \\
\midrule
\rowcolor{gray!15} \multicolumn{8}{c}{\textbf{\emph{Frontier API Models, Prompted as Critic Models}}} \\
Claude-Opus-4.5 & 74.9 & 69.0 & 80.9 & 76.3 & 75.8 & 77.4 & 78.4 \\
DeepSeek-V3.2 & 67.0 & 62.0 & 72.0 & 67.7 & 62.8 & 77.3 & 78.8 \\
GPT-5.4 & \underline{75.8} & \underline{69.4} & \underline{82.2} & \underline{77.5} & 78.7 & 75.1 & 79.3 \\
GPT-5.2 & 75.1 & 68.5 & 81.7 & 76.8 & 78.2 & 74.1 & 77.4 \\
GLM-4.7 & 73.4 & 66.6 & 80.2 & 75.2 & 76.4 & 72.7 & 74.0 \\
Qwen3-Max & 71.9 & 64.4 & 79.3 & 73.8 & 76.1 & 69.5 & 72.7 \\
DeepSeek-V3.1 & 72.0 & 64.0 & 80.0 & 74.3 & 78.0 & 67.1 & 67.3 \\
DeepSeek-R1 & 68.9 & 60.0 & 77.8 & 71.5 & 75.9 & 62.9 & 62.7 \\
Gemini-3.1-Pro & 63.7 & 47.6 & 79.9 & 70.9 & 87.4 & 38.8 & 60.2 \\
\textbf{Avg.} & 71.4 & 63.5 & 79.3 & 73.8 & 76.6 & 68.3 & 72.3 \\
\midrule
\rowcolor{gray!15} \multicolumn{8}{c}{\textbf{\emph{Medical Process Reward Models}}} \\
Med-PRM-Reward v1.0 & 55.7 & 57.5 & 53.8 & 55.8 & 40.6 & \underline{82.1} & 64.0 \\
Ours & \textbf{84.1} & \textbf{78.4} & \textbf{89.9} & \textbf{86.2} & \underline{96.6} & 68.2 & \underline{83.1} \\
\bottomrule
\end{tabular}}
\end{table}

\begin{table}[!htbp]
\centering
\caption{A performance comparison of popular models across detailed metrics in \textbf{E-2 (Deception Resistance)} sub-category of MedPRMBench. Category: Sensitivity. The best performance for each metric is highlighted in \textbf{bold}, while the second-best performance is \underline{underlined}.}
\label{tab:error_type_E_2}
\resizebox{0.8\textwidth}{!}{
\begin{tabular}{lccccccc}
\toprule
\textbf{Model Name} & \textbf{PRMScore} & \textbf{F1} & \textbf{Negative F1} & \textbf{Acc} & \textbf{Positive Acc} & \textbf{Negative Acc} & \textbf{First} \\
\midrule
\rowcolor{gray!15} \multicolumn{8}{c}{\textbf{\emph{Open-source Medical Models, Prompted as Critic Models}}} \\
txgemma-9b-chat & 55.6 & 43.5 & 67.8 & 59.0 & 73.5 & 38.3 & 37.6 \\
Llama-3.1-8B-UltraMedical & 55.3 & 45.0 & 65.5 & 57.6 & 68.6 & 42.0 & 30.8 \\
llama-3-meerkat-8b & 54.0 & 40.6 & 67.5 & 58.0 & 70.7 & 37.5 & 22.6 \\
HuatuoGPT-o1-8B & 56.5 & 42.7 & 70.3 & 60.9 & 75.1 & 38.0 & 24.3 \\
Meditron3-8B & 45.3 & 17.2 & 73.5 & 59.9 & \textbf{94.8} & 10.1 & 5.6 \\
\textbf{Avg.} & 53.3 & 37.8 & 68.9 & 59.1 & 76.5 & 33.2 & 24.2 \\
\midrule
\rowcolor{gray!15} \multicolumn{8}{c}{\textbf{\emph{Open-source Reasoning Models, Prompted as Critic Models}}} \\
Sky-T1-32B-Flash & 63.1 & 58.5 & 67.7 & 63.6 & 64.7 & 62.1 & 61.1 \\
QwQ-32B & 38.8 & 51.8 & 25.8 & 41.6 & 16.5 & \underline{81.9} & \underline{88.1} \\
Marco-o1 & 51.1 & 29.0 & 73.2 & 61.1 & 90.6 & 19.2 & 19.1 \\
DeepSeek-R1-Distill-Llama-70B & 55.1 & 36.6 & 73.7 & 62.8 & 84.4 & 28.0 & 8.6 \\
Qwen3-8B & 54.8 & 44.1 & 65.4 & 57.3 & 65.6 & 44.0 & 41.8 \\
DeepSeek-R1-Distill-Qwen-32B & 49.6 & 50.1 & 49.1 & 49.6 & 39.4 & 66.0 & 65.8 \\
\textbf{Avg.} & 52.1 & 45.0 & 59.1 & 56.0 & 60.2 & 50.2 & 47.4 \\
\midrule
\rowcolor{gray!15} \multicolumn{8}{c}{\textbf{\emph{Frontier API Models, Prompted as Critic Models}}} \\
Claude-Opus-4.5 & 71.4 & 71.4 & 71.3 & 71.4 & 60.7 & \textbf{86.6} & 87.0 \\
DeepSeek-V3.2 & 62.3 & 62.5 & 62.1 & 62.3 & 52.6 & 76.2 & 77.9 \\
GPT-5.4 & \underline{74.0} & \underline{72.0} & \underline{75.9} & \underline{74.1} & 69.4 & 80.8 & 84.5 \\
GPT-5.2 & 72.9 & 70.4 & 75.3 & 73.1 & 69.9 & 77.7 & 82.3 \\
GLM-4.7 & 69.8 & 67.5 & 72.0 & 69.9 & 66.0 & 75.6 & 76.6 \\
Qwen3-Max & 65.3 & 60.3 & 70.2 & 65.9 & 68.2 & 62.8 & 69.5 \\
DeepSeek-V3.1 & 64.7 & 58.2 & 71.3 & 66.0 & 71.9 & 57.4 & 58.7 \\
DeepSeek-R1 & 61.6 & 52.7 & 70.5 & 63.6 & 73.8 & 49.1 & 52.4 \\
Gemini-3.1-Pro & 61.3 & 50.1 & 72.5 & 64.6 & 79.6 & 43.1 & 72.9 \\
\textbf{Avg.} & 67.0 & 62.8 & 71.2 & 67.9 & 68.0 & 67.7 & 73.5 \\
\midrule
\rowcolor{gray!15} \multicolumn{8}{c}{\textbf{\emph{Medical Process Reward Models}}} \\
Med-PRM-Reward v1.0 & 55.0 & 57.7 & 52.3 & 55.1 & 39.8 & 79.8 & 60.6 \\
Ours & \textbf{87.5} & \textbf{84.2} & \textbf{90.9} & \textbf{88.5} & \underline{93.8} & 79.9 & \textbf{94.1} \\
\bottomrule
\end{tabular}}
\end{table}

\begin{table}[!htbp]
\centering
\caption{A performance comparison of popular models across detailed metrics in \textbf{E-3 (Multi-Solution Consistency)} sub-category of MedPRMBench. Category: Sensitivity. The best performance for each metric is highlighted in \textbf{bold}, while the second-best performance is \underline{underlined}.}
\label{tab:error_type_E_3}
\resizebox{0.8\textwidth}{!}{
\begin{tabular}{lccccccc}
\toprule
\textbf{Model Name} & \textbf{PRMScore} & \textbf{F1} & \textbf{Negative F1} & \textbf{Acc} & \textbf{Positive Acc} & \textbf{Negative Acc} & \textbf{First} \\
\midrule
\rowcolor{gray!15} \multicolumn{8}{c}{\textbf{\emph{Open-source Medical Models, Prompted as Critic Models}}} \\
txgemma-9b-chat & 59.2 & 45.6 & 72.8 & 63.8 & 69.6 & 50.2 & 50.2 \\
Llama-3.1-8B-UltraMedical & 57.6 & 43.1 & 72.1 & 62.6 & 69.4 & 46.8 & 37.0 \\
llama-3-meerkat-8b & 55.0 & 36.9 & 73.0 & 62.2 & 71.7 & 38.6 & 27.9 \\
HuatuoGPT-o1-8B & 59.5 & 43.0 & 76.0 & 66.3 & 75.0 & 44.5 & 32.7 \\
Meditron3-8B & 48.8 & 15.9 & 81.8 & 70.1 & \underline{96.6} & 9.3 & 5.5 \\
\textbf{Avg.} & 56.0 & 36.9 & 75.1 & 65.0 & 76.5 & 37.9 & 30.7 \\
\midrule
\rowcolor{gray!15} \multicolumn{8}{c}{\textbf{\emph{Open-source Reasoning Models, Prompted as Critic Models}}} \\
Sky-T1-32B-Flash & 69.4 & 62.6 & 76.3 & 71.0 & 67.0 & 80.1 & 80.4 \\
QwQ-32B & 36.6 & 45.7 & 27.4 & 37.9 & 16.4 & \textbf{91.5} & \underline{93.6} \\
Marco-o1 & 56.0 & 32.7 & 79.3 & 68.3 & 87.0 & 25.4 & 26.9 \\
DeepSeek-R1-Distill-Llama-70B & 60.2 & 42.5 & 77.9 & 68.1 & 78.8 & 41.2 & 16.9 \\
Qwen3-8B & 58.3 & 49.4 & 67.2 & 60.2 & 57.1 & 67.9 & 68.4 \\
DeepSeek-R1-Distill-Qwen-32B & 44.9 & 46.0 & 43.8 & 44.9 & 30.0 & 82.0 & 78.7 \\
\textbf{Avg.} & 54.2 & 46.5 & 62.0 & 58.4 & 56.1 & 64.7 & 60.8 \\
\midrule
\rowcolor{gray!15} \multicolumn{8}{c}{\textbf{\emph{Frontier API Models, Prompted as Critic Models}}} \\
Claude-Opus-4.5 & 78.5 & 73.0 & 83.9 & 79.8 & 75.5 & \underline{89.8} & 90.6 \\
DeepSeek-V3.2 & 69.0 & 63.6 & 74.5 & 70.0 & 62.8 & 86.5 & 86.9 \\
GPT-5.4 & \underline{79.6} & \underline{73.9} & \underline{85.3} & \underline{81.2} & 78.3 & 87.8 & 90.3 \\
GPT-5.2 & 79.1 & 73.1 & 85.1 & 80.9 & 78.6 & 86.0 & 88.7 \\
GLM-4.7 & 76.3 & 69.9 & 82.6 & 78.0 & 75.1 & 84.7 & 86.3 \\
Qwen3-Max & 76.5 & 69.8 & 83.2 & 78.5 & 76.8 & 82.3 & 85.4 \\
DeepSeek-V3.1 & 75.8 & 68.8 & 82.8 & 77.8 & 76.5 & 80.7 & 81.0 \\
DeepSeek-R1 & 72.8 & 64.9 & 80.8 & 75.2 & 74.9 & 75.8 & 75.7 \\
Gemini-3.1-Pro & 63.5 & 44.6 & 82.4 & 73.3 & 89.7 & 35.5 & 61.7 \\
\textbf{Avg.} & 74.6 & 66.8 & 82.3 & 77.2 & 76.5 & 78.8 & 83.0 \\
\midrule
\rowcolor{gray!15} \multicolumn{8}{c}{\textbf{\emph{Medical Process Reward Models}}} \\
Med-PRM-Reward v1.0 & 53.2 & 50.9 & 55.5 & 53.3 & 40.7 & 84.9 & 62.6 \\
Ours & \textbf{89.3} & \textbf{84.2} & \textbf{94.4} & \textbf{91.7} & \textbf{97.5} & 77.4 & \textbf{93.7} \\
\bottomrule
\end{tabular}}
\end{table}

\begin{table}[!htbp]
\centering
\caption{A performance comparison of popular models across detailed metrics in \textbf{E-4 (Quantitative Correctness)} sub-category of MedPRMBench. Category: Sensitivity. The best performance for each metric is highlighted in \textbf{bold}, while the second-best performance is \underline{underlined}.}
\label{tab:error_type_E_4}
\resizebox{0.8\textwidth}{!}{
\begin{tabular}{lccccccc}
\toprule
\textbf{Model Name} & \textbf{PRMScore} & \textbf{F1} & \textbf{Negative F1} & \textbf{Acc} & \textbf{Positive Acc} & \textbf{Negative Acc} & \textbf{First} \\
\midrule
\rowcolor{gray!15} \multicolumn{8}{c}{\textbf{\emph{Open-source Medical Models, Prompted as Critic Models}}} \\
txgemma-9b-chat & 58.4 & 47.6 & 69.1 & 61.1 & 70.3 & 46.2 & 50.3 \\
Llama-3.1-8B-UltraMedical & 57.0 & 45.9 & 68.2 & 59.9 & 69.4 & 44.5 & 30.9 \\
llama-3-meerkat-8b & 56.2 & 43.2 & 69.2 & 60.1 & 71.8 & 40.5 & 25.6 \\
HuatuoGPT-o1-8B & 59.1 & 46.3 & 71.8 & 63.0 & 75.4 & 42.5 & 27.8 \\
Meditron3-8B & 48.7 & 21.0 & 76.3 & 63.6 & \underline{94.9} & 12.7 & 5.4 \\
\textbf{Avg.} & 55.9 & 40.8 & 70.9 & 61.5 & 76.4 & 37.3 & 28.0 \\
\midrule
\rowcolor{gray!15} \multicolumn{8}{c}{\textbf{\emph{Open-source Reasoning Models, Prompted as Critic Models}}} \\
Sky-T1-32B-Flash & 66.6 & 63.8 & 69.4 & 66.8 & 60.8 & 76.5 & 76.2 \\
QwQ-32B & 36.8 & 53.8 & 19.8 & 41.4 & 11.6 & \textbf{91.2} & \textbf{94.4} \\
Marco-o1 & 55.8 & 37.0 & 74.7 & 63.9 & 86.1 & 27.8 & 30.8 \\
DeepSeek-R1-Distill-Llama-70B & 60.3 & 47.9 & 72.7 & 64.1 & 76.2 & 44.0 & 17.3 \\
Qwen3-8B & 57.6 & 54.4 & 60.7 & 57.8 & 52.2 & 67.2 & 64.5 \\
DeepSeek-R1-Distill-Qwen-32B & 48.5 & 53.3 & 43.6 & 48.9 & 31.6 & 77.7 & 74.3 \\
\textbf{Avg.} & 54.3 & 51.7 & 56.8 & 57.1 & 53.1 & 64.1 & 59.6 \\
\midrule
\rowcolor{gray!15} \multicolumn{8}{c}{\textbf{\emph{Frontier API Models, Prompted as Critic Models}}} \\
Claude-Opus-4.5 & 76.0 & 73.6 & 78.3 & 76.2 & 69.4 & 87.2 & 87.8 \\
DeepSeek-V3.2 & 66.4 & 66.4 & 66.3 & 66.4 & 53.4 & \underline{87.3} & \underline{87.9} \\
GPT-5.4 & \underline{77.8} & \underline{74.8} & \underline{80.8} & \underline{78.2} & 74.2 & 84.7 & \underline{87.9} \\
GPT-5.2 & 77.0 & 73.7 & 80.3 & 77.4 & 74.2 & 82.7 & 85.4 \\
GLM-4.7 & 74.5 & 71.3 & 77.7 & 74.9 & 70.6 & 81.8 & 83.3 \\
Qwen3-Max & 72.4 & 68.4 & 76.4 & 73.0 & 70.8 & 76.5 & 81.7 \\
DeepSeek-V3.1 & 74.1 & 70.2 & 78.0 & 74.7 & 72.7 & 77.9 & 80.7 \\
DeepSeek-R1 & 70.3 & 65.6 & 75.0 & 71.1 & 70.3 & 72.3 & 72.3 \\
Gemini-3.1-Pro & 65.7 & 53.5 & 77.9 & 70.0 & 85.4 & 45.1 & 72.0 \\
\textbf{Avg.} & 72.7 & 68.6 & 76.7 & 73.5 & 71.2 & 77.3 & 82.1 \\
\midrule
\rowcolor{gray!15} \multicolumn{8}{c}{\textbf{\emph{Medical Process Reward Models}}} \\
Med-PRM-Reward v1.0 & 57.9 & 60.6 & 55.1 & 58.0 & 41.3 & 86.0 & 70.5 \\
Ours & \textbf{88.3} & \textbf{84.8} & \textbf{91.8} & \textbf{89.4} & \textbf{95.6} & 79.1 & \textbf{94.4} \\
\bottomrule
\end{tabular}}
\end{table}

\begin{table}[!htbp]
\centering
\caption{A performance comparison of popular models across detailed metrics in \textbf{E-5 (Differential Diagnosis Coverage)} sub-category of MedPRMBench. Category: Sensitivity. The best performance for each metric is highlighted in \textbf{bold}, while the second-best performance is \underline{underlined}.}
\label{tab:error_type_E_5}
\resizebox{0.8\textwidth}{!}{
\begin{tabular}{lccccccc}
\toprule
\textbf{Model Name} & \textbf{PRMScore} & \textbf{F1} & \textbf{Negative F1} & \textbf{Acc} & \textbf{Positive Acc} & \textbf{Negative Acc} & \textbf{First} \\
\midrule
\rowcolor{gray!15} \multicolumn{8}{c}{\textbf{\emph{Open-source Medical Models, Prompted as Critic Models}}} \\
txgemma-9b-chat & 57.9 & 47.9 & 67.9 & 60.3 & 67.2 & 48.7 & 49.7 \\
Llama-3.1-8B-UltraMedical & 56.3 & 43.9 & 68.8 & 59.9 & 70.9 & 41.7 & 24.1 \\
llama-3-meerkat-8b & 54.1 & 39.9 & 68.4 & 58.5 & 72.0 & 36.3 & 18.8 \\
HuatuoGPT-o1-8B & 57.7 & 43.6 & 71.7 & 62.3 & 76.8 & 38.6 & 20.1 \\
Meditron3-8B & 45.7 & 14.5 & 77.0 & 63.7 & \textbf{97.1} & 8.2 & 2.4 \\
\textbf{Avg.} & 54.3 & 38.0 & 70.8 & 60.9 & 76.8 & 34.7 & 23.0 \\
\midrule
\rowcolor{gray!15} \multicolumn{8}{c}{\textbf{\emph{Open-source Reasoning Models, Prompted as Critic Models}}} \\
Sky-T1-32B-Flash & 64.9 & 62.2 & 67.6 & 65.1 & 58.2 & 76.5 & 72.9 \\
QwQ-32B & 38.3 & 53.4 & 23.1 & 42.0 & 14.0 & \textbf{87.9} & \textbf{93.5} \\
Marco-o1 & 53.9 & 32.5 & 75.3 & 63.9 & 88.3 & 23.2 & 22.9 \\
DeepSeek-R1-Distill-Llama-70B & 58.8 & 46.1 & 71.6 & 62.8 & 75.4 & 42.1 & 11.4 \\
Qwen3-8B & 57.0 & 54.2 & 59.8 & 57.2 & 51.2 & 66.9 & 63.4 \\
DeepSeek-R1-Distill-Qwen-32B & 44.1 & 52.1 & 36.1 & 45.2 & 24.9 & 78.6 & 75.0 \\
\textbf{Avg.} & 52.8 & 50.1 & 55.6 & 56.0 & 52.0 & 62.5 & 56.5 \\
\midrule
\rowcolor{gray!15} \multicolumn{8}{c}{\textbf{\emph{Frontier API Models, Prompted as Critic Models}}} \\
Claude-Opus-4.5 & 71.2 & 68.9 & 73.4 & 71.4 & 63.4 & 84.5 & 86.4 \\
DeepSeek-V3.2 & 64.2 & 64.0 & 64.4 & 64.2 & 51.9 & 84.7 & 85.5 \\
GPT-5.4 & \underline{74.0} & \underline{70.9} & 77.2 & \underline{74.4} & 69.2 & 83.1 & 88.0 \\
GPT-5.2 & 73.4 & 69.9 & 77.0 & 73.9 & 70.0 & 80.5 & 85.3 \\
GLM-4.7 & 70.7 & 67.2 & 74.2 & 71.2 & 66.6 & 78.8 & 80.8 \\
Qwen3-Max & 70.0 & 65.3 & 74.7 & 70.7 & 69.1 & 73.4 & 77.2 \\
DeepSeek-V3.1 & 70.0 & 65.5 & 74.4 & 70.6 & 68.4 & 74.3 & 76.5 \\
DeepSeek-R1 & 67.4 & 62.3 & 72.5 & 68.2 & 67.1 & 70.0 & 70.5 \\
Gemini-3.1-Pro & 64.8 & 52.2 & \underline{77.4} & 69.3 & 84.2 & 44.6 & 72.4 \\
\textbf{Avg.} & 69.5 & 65.1 & 73.9 & 70.4 & 67.8 & 74.9 & 80.3 \\
\midrule
\rowcolor{gray!15} \multicolumn{8}{c}{\textbf{\emph{Medical Process Reward Models}}} \\
Med-PRM-Reward v1.0 & 53.8 & 59.1 & 48.6 & 54.4 & 34.7 & \underline{87.0} & 63.9 \\
Ours & \textbf{86.2} & \textbf{81.9} & \textbf{90.5} & \textbf{87.5} & \underline{95.3} & 74.7 & \underline{93.2} \\
\bottomrule
\end{tabular}}
\end{table}


\FloatBarrier
\section{Prompt Templates}
\label{app:prompts}

This appendix presents the key prompt templates used throughout the MedPRMBench construction and evaluation pipeline. We organize them into three parts: (1)~the ERN extraction prompt used in Phase~2 for Clinical Reasoning Blueprint construction (\S\ref{app:prompt_ern}), (2)~representative error injection prompts from Phase~3 covering all three error categories (\S\ref{app:prompt_injection}), and (3)~the step-level verification prompts used for model evaluation (\S\ref{app:prompt_eval}).


\subsection{Phase 2: Evidence Reasoning Network Extraction}
\label{app:prompt_ern}

The following prompt is used to extract medical knowledge triplets from reasoning text, forming the Evidence Reasoning Network (ERN) that serves as the foundation for Clinical Reasoning Blueprint (CRB) construction. Each of three models (Qwen3-Max, GPT-5.2, Claude-Opus-4.5) independently receives this prompt, and the outputs are fused via semantic voting (\S\ref{sec:blueprint_construction}).

\begin{tcolorbox}[breakable,colback=cyan!4, colframe=cyan!35!gray, title={\textbf{ERN Extraction Prompt}}, fonttitle=\small, left=4pt, right=4pt, top=3pt, bottom=3pt]
\footnotesize
You are an expert medical knowledge engineer. Your task is to extract a CONNECTED reasoning graph from a medical reasoning process. The graph must form a single connected chain of causal/logical relationships---not a collection of isolated facts.

\smallskip
\textbf{CRITICAL REQUIREMENT: GRAPH CONNECTIVITY}

The triplets you produce must form a SINGLE CONNECTED DIRECTED GRAPH:
\begin{itemize}[nosep]
    \item Every triplet must share at least one entity (subject or object) with another triplet.
    \item There must be a traceable path from the initial clinical findings to the final diagnosis/answer.
    \item NO isolated triplets or disconnected subgraphs are allowed.
    \item Reuse the SAME entity string when referring to the same concept across triplets.
\end{itemize}

\smallskip
\textbf{Processing Rules:}

\textbf{1. Entity Standardization (CRITICAL for Connectivity)}
\begin{itemize}[nosep]
    \item \textbf{Canonical Names}: For each medical concept, choose ONE canonical name and use it consistently across ALL triplets.
    \item \textbf{Atomic Entities}: Entities must be core nouns/noun phrases. Move modifiers into relationships.
\end{itemize}

\textbf{2. Causal Chain Construction}
\begin{itemize}[nosep]
    \item Follow the reasoning text's logical flow. For each step, create triplets that CONNECT to the previous step's entities.
    \item Use causal/inferential predicates: ``suggests'', ``leads to'', ``indicates'', ``rules out'', ``confirms'', ``requires'', ``results in''.
    \item The chain should flow: clinical findings $\to$ pathophysiology $\to$ differential reasoning $\to$ diagnosis $\to$ treatment.
\end{itemize}

\textbf{3. Bridging Triplets}: When two consecutive steps share no entity, add a bridging triplet that links them.

\smallskip
\textbf{Output Format:} Return a JSON object with a single key \texttt{"triplets"} containing a list of \texttt{[subject, predicate, object]} triples.

\smallskip
\textbf{Example Output:}

\smallskip
{\ttfamily\small
\{\\
\quad"triplets": [\\
\quad\quad["biopsy", "reveals", "invasive ductal carcinoma"],\\
\quad\quad["invasive ductal carcinoma", "is type of", "malignancy"],\\
\quad\quad["invasive ductal carcinoma", "tested for", "HER2/neu"],\\
\quad\quad["HER2/neu", "status is", "positive"],\\
\quad\quad["HER2/neu positive", "is indication for", "Trastuzumab"],\\
\quad\quad["invasive ductal carcinoma", "is treated with", "Trastuzumab"]\\
\quad]\\
\}
}
\end{tcolorbox}
\nopagebreak[4]
\captionof{figure}{Prompt used for Evidence Reasoning Network (ERN) extraction during Clinical Reasoning Blueprint construction (Phase~2).}
\label{fig:prompt_ern}
\par\medskip

\subsection{Phase 3: Error Injection Prompts}
\label{app:prompt_injection}

We select one representative error type from each of the three categories---Simplicity, Soundness, and Sensitivity---to illustrate the prompt design. Each prompt includes: (1)~the error type definition, (2)~relationship to other error types for disambiguation, (3)~two few-shot examples with original and corrupted reasoning chains, and (4)~structured injection instructions. All 14 error type prompts follow the same template structure.


\subsection{S-1: Non-Redundancy (Simplicity)}
\label{app:prompt_s1}
\begin{tcolorbox}[breakable,colback=gray!3, colframe=blue!40!gray, title={\textbf{S-1: Non-Redundancy --- Full Injection Prompt}}, fonttitle=\small, left=4pt, right=4pt, top=4pt, bottom=4pt]
\small\linespread{1.05}\selectfont

\textbf{\textcolor{blue!70!black}{\#\# Error Type Definition}}

A correct reasoning chain should not contain unnecessary or irrelevant steps. The injected error adds verbose, tangential, or duplicated reasoning steps that do not contribute to reaching the answer and do not change the clinical decision. For example, ordering an unnecessary confirmatory test when the diagnosis is already well-established.

\smallskip
\textbf{\textcolor{blue!70!black}{\#\# Relationship to Other Error Types}}

S-1 (Non-Redundancy) is a Simplicity-category error. Unlike S-2 (Non-Circular Logic) which loops back to a prior conclusion, S-1 inserts tangential steps that never circle back---they simply add noise. S-1 differs from R-6 (Information Grounding) in that the inserted content may be factually correct but clinically unnecessary.

\smallskip
\textbf{\textcolor{blue!70!black}{\#\# Few-Shot Example}}

\textbf{Question:} A 25-year-old male presents with acute right lower quadrant pain, rebound tenderness, fever (38.5\textdegree C), and WBC 15,000/$\mu$L. What is the most likely diagnosis?

\smallskip
\textbf{Original correct reasoning:}
\begin{enumerate}[leftmargin=*, nosep]
    \item The patient presents with acute RLQ pain, rebound tenderness, fever, and leukocytosis.
    \item These findings are classic for acute appendicitis---McBurney's point tenderness with peritoneal signs.
    \item The combination of clinical presentation and lab findings strongly supports acute appendicitis.
    \item Therefore, the most likely diagnosis is acute appendicitis, and surgical consultation for appendectomy is indicated.
\end{enumerate}

\smallskip
\textbf{Corrupted reasoning (with S-1 error):}
\begin{enumerate}[leftmargin=*, nosep]
    \item The patient presents with acute RLQ pain, rebound tenderness, fever, and leukocytosis.
    \item These findings are classic for acute appendicitis---McBurney's point tenderness with peritoneal signs.
    \item[] \colorbox{red!8}{\parbox[t]{\dimexpr\linewidth-2\fboxsep}{\textbf{Step 3:} To further confirm the diagnosis, we should order an abdominal MRI with contrast to visualize the appendix in detail and rule out any atypical presentations. \hfill$\leftarrow$ \textit{error}}}
    \item[] \colorbox{red!8}{\parbox[t]{\dimexpr\linewidth-2\fboxsep}{\textbf{Step 4:} Additionally, a barium enema study could help delineate the cecal anatomy and provide supplementary diagnostic information. \hfill$\leftarrow$ \textit{error}}}
    \item The combination of clinical presentation and lab findings strongly supports acute appendicitis.
    \item Therefore, the most likely diagnosis is acute appendicitis, and surgical consultation for appendectomy is indicated.
\end{enumerate}

\smallskip
\textbf{Error root causes:} Steps 3--4 \\
\textbf{Explanation:} Steps 3--4 insert redundant tests (abdominal MRI, barium enema) when the diagnosis is already well-established clinically. These steps are factually plausible but clinically unnecessary, adding noise without changing the final decision.

\smallskip
\textbf{\textcolor{blue!70!black}{\#\# Output Format}}

{\ttfamily\small
\{"corrupted\_steps": [...], "modified\_steps": [...],\\
\quad"error\_steps": [...], "error\_step\_indices": [...],\\
\quad"error\_description": "...", "reason": "..."\}
}
\end{tcolorbox}
\nopagebreak[4]
\captionof{figure}{Prompts used for test case construction during data generation (1). S-1 Non-Redundancy error injection prompt from the Simplicity category.}
\label{fig:prompt_s1}
\par\medskip

\subsection{R-4: Confidence Invariance (Soundness)}
\label{app:prompt_r4}
\begin{tcolorbox}[breakable,colback=gray!3, colframe=orange!50!gray, title={\textbf{R-4: Confidence Invariance --- Full Injection Prompt}}, fonttitle=\small, left=4pt, right=4pt, top=4pt, bottom=4pt]
\small\linespread{1.05}\selectfont

\textbf{\textcolor{orange!70!black}{\#\# Error Type Definition}}

Error injection introduces a statement that directly contradicts established facts from the question stem/patient history, but presents this contradiction with overconfident, authoritative language. The PRM should detect the factual inconsistency regardless of how confidently it is stated. Key markers: ``definitively,'' ``clearly,'' ``without doubt,'' ``confirms,'' ``completely normal.''

\smallskip
\textbf{\textcolor{orange!70!black}{\#\# Relationship to Other Error Types}}

R-4 (Confidence Invariance) is a Soundness-category error where overconfident language disguises factual contradictions with the question stem. It differs from R-2 (Step Consistency) which involves contradictions between reasoning steps (not with the question stem). It also differs from R-1 (Evidence-Based Soundness) where the medical knowledge itself is wrong---in R-4, the knowledge may be correct but is applied to contradict given facts.

\smallskip
\textbf{\textcolor{orange!70!black}{\#\# Few-Shot Example}}

\textbf{Question:} A 50-year-old male presents with chest pain. Labs show: Troponin I 2.5~ng/mL (elevated), K$^+$ 5.8~mmol/L (elevated), Creatinine 3.2~mg/dL (elevated). ECG shows ST-segment elevation in leads II, III, aVF. What is the diagnosis and initial management?

\smallskip
\textbf{Original correct reasoning:}
\begin{enumerate}[leftmargin=*, nosep]
    \item Elevated troponin I with ST-elevation in inferior leads indicates an acute inferior STEMI.
    \item The patient also has hyperkalemia (K$^+$ 5.8) and elevated creatinine (3.2), suggesting concurrent renal impairment that must be considered in management.
    \item Immediate management includes dual antiplatelet therapy, anticoagulation, and emergent PCI. Renal function must be considered when choosing contrast agents.
    \item Hyperkalemia requires urgent treatment with calcium gluconate for cardiac protection, insulin/glucose, and monitoring.
\end{enumerate}

\smallskip
\textbf{Corrupted reasoning (with R-4 error):}
\begin{enumerate}[leftmargin=*, nosep]
    \item Elevated troponin I with ST-elevation in inferior leads indicates an acute inferior STEMI.
    \item[] \colorbox{red!8}{\parbox[t]{\dimexpr\linewidth-2\fboxsep}{\textbf{Step 2:} The patient's electrolyte panel is \textit{completely normal and reassuring}---potassium and renal function are \textit{well within acceptable limits}, \textit{confirming} that there are no metabolic complications to address. \hfill$\leftarrow$ \textit{error}}}
    \item Immediate management includes dual antiplatelet therapy, anticoagulation, and emergent PCI. No dose adjustments are needed given the normal metabolic profile.
    \item The focus should be entirely on coronary reperfusion without any need for electrolyte correction or renal-protective measures.
\end{enumerate}

\smallskip
\textbf{Error root cause:} Step 2 \\
\textbf{Explanation:} Step 2 directly contradicts the question stem---K$^+$ is 5.8~mmol/L (hyperkalemia) and creatinine is 3.2~mg/dL (renal impairment), but the corrupted step claims ``completely normal'' with overconfident language (``completely,'' ``confirming''). Steps 3--4 propagate the error by omitting necessary dose adjustments and electrolyte correction.

\smallskip
\textbf{\textcolor{orange!70!black}{\#\# Injection Instructions}}
\begin{enumerate}[leftmargin=*, nosep]
    \item Replace the target step with a statement that:
    \begin{itemize}[nosep]
        \item Directly contradicts a specific fact given in the question stem (lab value, vital sign, imaging finding, history)
        \item Uses overconfident, authoritative language: ``definitively,'' ``clearly confirms,'' ``without question,'' ``completely normal''
        \item Sounds professionally written and medically fluent
    \end{itemize}
    \item Adjust subsequent steps to be consistent with the false claim
    \item The contradiction must be with the QUESTION STEM facts, not with other reasoning steps (that would be R-2)
\end{enumerate}
\end{tcolorbox}
\nopagebreak[4]
\captionof{figure}{Prompts used for test case construction during data generation (2). R-4 Confidence Invariance error injection prompt from the Soundness category.}
\label{fig:prompt_r4}
\par\medskip

\subsection{E-1: Prerequisite Sensitivity (Sensitivity)}
\label{app:prompt_e1}
\begin{tcolorbox}[breakable,colback=gray!3, colframe=teal!50!gray, title={\textbf{E-1: Prerequisite Sensitivity --- Full Injection Prompt}}, fonttitle=\small, left=4pt, right=4pt, top=4pt, bottom=4pt]
\small\linespread{1.05}\selectfont

\textbf{\textcolor{teal!70!black}{\#\# Error Type Definition}}

Error injection removes a critical prerequisite step from the reasoning chain---a step that must be completed before subsequent steps can be valid. The omission makes the conclusion unreliable even if it happens to be correct. Examples: skipping allergy check before prescribing, omitting renal function assessment before contrast CT, not confirming fasting status before interpreting fasting glucose.

\smallskip
\textbf{\textcolor{teal!70!black}{\#\# Relationship to Other Error Types}}

E-1 (Prerequisite Sensitivity) is a Sensitivity-category error involving skipped prerequisite checks. It differs from R-5 (Safety Awareness) which specifically targets safety-critical omissions. E-1 is broader---it includes any prerequisite that must be satisfied before a clinical action, not just safety checks. It also differs from R-3 (Contextual Applicability) which misapplies knowledge rather than skipping checks.

\smallskip
\textbf{\textcolor{teal!70!black}{\#\# Few-Shot Example}}

\textbf{Question:} A 45-year-old woman with suspected pulmonary embolism needs a CT pulmonary angiography (CTPA). She has no known allergies. Her last creatinine was 1.0~mg/dL (3 months ago). What should be done before the scan?

\smallskip
\textbf{Original correct reasoning:}
\begin{enumerate}[leftmargin=*, nosep]
    \item CTPA requires IV iodinated contrast, which carries risks of contrast-induced nephropathy (CIN) and allergic reactions.
    \item Before administering contrast, we must verify current renal function. The last creatinine (1.0~mg/dL) was 3 months ago---a repeat creatinine or eGFR should be obtained.
    \item Although no known allergies are documented, the patient should be asked about prior contrast reactions.
    \item If renal function is adequate (eGFR $>$30) and no contrast allergy history, proceed with CTPA with adequate hydration.
\end{enumerate}

\smallskip
\textbf{Corrupted reasoning (with E-1 error):}
\begin{enumerate}[leftmargin=*, nosep]
    \item CTPA requires IV iodinated contrast for visualization of pulmonary vasculature.
    \item[] \colorbox{red!8}{\parbox[t]{\dimexpr\linewidth-2\fboxsep}{\textbf{Step 2:} The patient has no known allergies, so contrast administration is safe. Proceed directly with CTPA. \hfill$\leftarrow$ \textit{error}}}
    \item The CTPA will definitively confirm or rule out pulmonary embolism.
    \item If PE is confirmed, initiate anticoagulation therapy.
\end{enumerate}

\smallskip
\textbf{Error root cause:} Step 2 \\
\textbf{Explanation:} Step 2 skips two critical prerequisites---verifying current renal function (last creatinine was 3 months ago) and asking about prior contrast reactions. Steps 3--4 propagate the error by proceeding without completing necessary safety assessments.

\smallskip
\textbf{\textcolor{teal!70!black}{\#\# Injection Instructions}}
\begin{enumerate}[leftmargin=*, nosep]
    \item Remove or skip the prerequisite check at the target step:
    \begin{itemize}[nosep]
        \item Omit allergy/drug reaction verification before prescribing
        \item Skip renal/hepatic function assessment before nephrotoxic/hepatotoxic drugs
        \item Bypass surgical fitness evaluation before recommending surgery
        \item Skip confirmation of test conditions (fasting, medication washout) before interpreting results
    \end{itemize}
    \item The reasoning should jump directly to the conclusion/action that depends on the missing prerequisite
    \item The omission should be natural---don't draw attention to what's missing
\end{enumerate}
\end{tcolorbox}
\nopagebreak[4]
\captionof{figure}{Prompts used for test case construction during data generation (3). E-1 Prerequisite Sensitivity error injection prompt from the Sensitivity category.}
\label{fig:prompt_e1}
\par\medskip


\subsection{Prompt Template Structure}
\label{app:prompt_structure}

All 14 error injection prompts follow a unified template structure, as summarized below:

\begin{tcolorbox}[breakable,colback=yellow!4, colframe=yellow!40!gray, title={\textbf{Unified Prompt Template}}, fonttitle=\small, left=4pt, right=4pt, top=3pt, bottom=3pt]
\footnotesize
{\ttfamily
You are a medical reasoning error injection expert.\\
Your task is to inject a **[Error Type Name]** error\\
into the given clinical reasoning chain.\\[0.5em]
\#\# Error Type Definition\\
\{definition\}\\[0.3em]
\#\# Relationship to Other Error Types\\
\{error\_type\_relation\}\\[0.3em]
\#\# Few-Shot Example 1\\
**Question:** \{example1\_question\}\\
**Original correct reasoning:** \{example1\_original\}\\
**Corrupted reasoning:** \{example1\_corrupted\}\\
**Explanation:** \{example1\_explanation\}\\[0.3em]
\#\# Few-Shot Example 2\\
{[}Same structure as Example 1{]}\\[0.3em]
\#\# Your Task\\
**Question:** \{question\}\\
**Correct reasoning steps:** \{reasoning\_steps\}\\
**Suggested injection region:** around Step \{N\}\\
**Additional context:** \{context\}\\[0.3em]
\#\# Instructions\\
{[}Error-type-specific injection guidelines{]}\\[0.3em]
Output strictly as JSON:\\
\{"corrupted\_steps": [...],\\
\quad"modified\_steps": [...],\\
\quad"error\_steps": [...],\\
\quad"error\_step\_indices": [...],\\
\quad"error\_description": "...",\\
\quad"reason": "..."\}
}
\end{tcolorbox}
\nopagebreak[4]
\captionof{figure}{Unified prompt template structure used for all 14 error type injection prompts during data generation.}
\label{fig:prompt_template}
\par\medskip

Each prompt template includes: (1)~a precise error type definition to ensure injection specificity; (2)~explicit disambiguation from related error types to prevent category confusion; (3)~two carefully crafted few-shot examples with original and corrupted reasoning chains, error root cause annotations, and explanations; and (4)~structured output format requiring step-level error labels compatible with the benchmark's annotation schema.

\subsection{Evaluation: Step-Level Verification Prompts}
\label{app:prompt_eval}

We evaluate the main benchmark results with the \textbf{basic} binary prompt (no metadata). The \textbf{enhanced} prompt, which includes safety-level and prerequisite annotations, is provided for metadata-aware diagnostic analysis but is not used for the main tables. Each mode includes a system prompt and few-shot example. For API-based models, the prompt is used in a generative evaluation protocol; for PRM-style models, we extract token probabilities at delimiter positions.

\Needspace{8\baselineskip}
\subsubsection{Basic Mode}
\label{app:prompt_eval_basic}

\noindent
\begin{tcolorbox}[breakable,colback=gray!3, colframe=blue!40!gray, title={\textbf{Basic Mode --- System Prompt \& Few-Shot Example}}, fonttitle=\small, left=4pt, right=4pt, top=3pt, bottom=3pt]
\footnotesize

\textbf{\textcolor{blue!70!black}{\#\# System Prompt}}

You are a medical reasoning evaluator. Your task is to analyze clinical reasoning steps and assess whether each step is correct or contains an error.

\smallskip
For each reasoning step, assign a binary validity label:
\begin{itemize}[nosep]
    \item \texttt{+}: The step is medically correct and logically sound.
    \item \texttt{-}: The step contains a clear medical or logical error.
\end{itemize}

\smallskip
Requirements:
\begin{itemize}[nosep]
    \item Evaluate each step independently based on medical knowledge and clinical logic.
    \item Return results in strict JSON format: \texttt{\{"labels": ["+", "-", ...]\}}
    \item The array length must equal the number of reasoning steps.
    \item Do NOT add any explanation outside the JSON.
\end{itemize}

\smallskip
\textbf{\textcolor{blue!70!black}{\#\# Few-Shot Example}}

\textbf{Question:} A 45-year-old male presents with chest pain radiating to the left arm, diaphoresis, and shortness of breath for 2 hours. ECG shows ST elevation in leads II, III, and aVF. What is the most appropriate immediate management?

\smallskip
\textbf{Reasoning Steps:}
\begin{enumerate}[leftmargin=*, nosep]
    \item The patient presents with classic symptoms of acute myocardial infarction: chest pain radiating to the left arm, diaphoresis, and shortness of breath.
    \item The ECG shows ST elevation in leads II, III, and aVF, which indicates an inferior STEMI.
    \item For STEMI, the standard of care is immediate reperfusion therapy via primary PCI.
    \item Aspirin 325~mg should be administered immediately as antiplatelet therapy, and the cardiac catheterization lab should be activated for primary PCI.
    \item Waiting for troponin results or scheduling an elective angiogram would delay reperfusion and worsen outcomes.
\end{enumerate}

\smallskip
\textbf{Expected Output:} \texttt{\{"labels": ["+", "+", "+", "+", "+"]\}}
\end{tcolorbox}
\nopagebreak[4]
\captionof{figure}{Prompts used for model evaluation (1). Basic binary step-level verification prompt without metadata annotations.}
\label{fig:prompt_eval_basic}
\par\medskip

\subsubsection{Enhanced Mode}

The enhanced mode augments each reasoning step with two metadata annotations: \texttt{safety\_level} (Critical $>$ Major $>$ Moderate $>$ Minor) indicating the potential patient harm if the step is wrong, and \texttt{is\_prerequisite\_of\_next} indicating whether the step is a logical prerequisite for the subsequent step.

\begin{tcolorbox}[breakable,colback=gray!3, colframe=cyan!50!gray, title={\textbf{Enhanced Mode --- System Prompt \& Few-Shot Example}}, fonttitle=\small, left=4pt, right=4pt, top=3pt, bottom=3pt]
\footnotesize

\textbf{\textcolor{cyan!70!black}{\#\# System Prompt}}

You are a medical reasoning evaluator with expertise in patient safety. Your task is to analyze clinical reasoning steps and assess whether each step is correct or contains an error.

\smallskip
Each reasoning step is annotated with:
\begin{itemize}[nosep]
    \item \texttt{safety\_level}: The potential harm if this step is wrong (Critical $>$ Major $>$ Moderate $>$ Minor)
    \item \texttt{is\_prerequisite\_of\_next}: Whether this step is a logical prerequisite for the next step
\end{itemize}

For each reasoning step, assign a binary validity label:
\begin{itemize}[nosep]
    \item \texttt{+}: The step is medically correct and logically sound.
    \item \texttt{-}: The step contains a clear medical or logical error.
\end{itemize}

Pay special attention to steps with \texttt{safety\_level = "Critical"} or \texttt{"Major"}, as errors in these steps can directly harm patients.

\smallskip
\textbf{\textcolor{cyan!70!black}{\#\# Few-Shot Example}}

\textbf{Question:} A 28-year-old woman with known penicillin allergy presents with a urinary tract infection. Urine culture grows \textit{E.~coli} sensitive to amoxicillin, trimethoprim-sulfamethoxazole, and nitrofurantoin. Which antibiotic is most appropriate?

\smallskip
\textbf{Reasoning Steps:}
\begin{enumerate}[leftmargin=*, nosep]
    \item \texttt{[safety\_level=Critical, prerequisite=True]} The patient has a known penicillin allergy, so amoxicillin (a penicillin) is contraindicated.
    \item \texttt{[safety\_level=Major, prerequisite=False]} Cephalexin is a cephalosporin; due to cross-reactivity with penicillin, it should be avoided in patients with penicillin allergy.
    \item \texttt{[safety\_level=Moderate, prerequisite=True]} Both TMP-SMX and nitrofurantoin are appropriate alternatives for uncomplicated UTI caused by \textit{E.~coli}.
    \item \texttt{[safety\_level=Moderate, prerequisite=True]} Nitrofurantoin is preferred for uncomplicated lower UTI because it achieves high urinary concentrations with minimal systemic side effects.
    \item \texttt{[safety\_level=Minor, prerequisite=True]} TMP-SMX is also effective but has higher resistance rates in some regions and more systemic side effects.
    \item \texttt{[safety\_level=Moderate, prerequisite=False]} Therefore, nitrofurantoin is the most appropriate choice for this patient.
\end{enumerate}

\smallskip
\textbf{Expected Output:} \texttt{\{"labels": ["+", "-", "+", "+", "+", "+"]\}}

\smallskip
\textit{Note: Step~2 is labeled ``-'' because the prompt should flag the overgeneralized cross-reactivity claim rather than assigning a partial scalar score.}
\end{tcolorbox}
\nopagebreak[4]
\captionof{figure}{Prompts used for model evaluation (2). Enhanced mode step-level verification prompt with safety-level and prerequisite metadata annotations.}
\label{fig:prompt_eval_enhanced}
\par\medskip

\section{Error Type Examples}
\label{app:examples}

This appendix presents one representative data instance from MedPRMBench for each of the 14 fine-grained error types, organized by category. Each example shows the original correct reasoning process, the modified (error-injected) reasoning process with error steps highlighted, and the reason explaining the injected error. All examples are drawn from the \texttt{test\_benchmark.jsonl} dataset.

\tcbset{
  examplebox/.style={
    colback=gray!4,
    colframe=gray!55,
    fonttitle=\small\bfseries,
    left=3pt, right=3pt, top=3pt, bottom=3pt,
    boxrule=0.4pt,
    before upper={\footnotesize}
  }
}

\subsection{Simplicity}

\begin{tcolorbox}[breakable,examplebox, title={S-1: Non-Redundancy \hfill \texttt{medprmbench\_01808}}]

\textbf{Question:} A previously healthy 10-year-old boy presented after his fourth generalized tonic-clonic seizure in 2 months. He had been started on levetiracetam after his second seizure but experienced two more breakthrough events. Neurologic examination revealed weakness of the right lower extremity. His family has ties to Western Africa; he last traveled to Guinea and C\^{o}te d'Ivoire 2 years earlier. MRI of the brain showed a 1.8-cm cortical/subcortical enhancing mass in the left precuneus with significant adjacent vasogenic edema. The lesion demonstrated areas of mineralization or blood products.

\tcblower

\textbf{1. Original Process}

\textbf{Step 1:} An oligodendroglioma was considered given the seizure presentation but was unlikely because the mass lacked the expected T2 hyperintensity.

\textbf{Step 2:} A ganglioglioma was included in the initial differential but was disfavored due to the lesion's T2 isointensity rather than hyperintensity.

\textbf{Step 3:} A pleomorphic xanthoastrocytoma was contemplated because it often presents with seizures, but the imaging did not show the characteristic T2 hyperintensity.

\textbf{Step 4:} An infectious or inflammatory process, particularly a CNS tuberculoma, entered the differential after a positive interferon-$\gamma$ release assay increased suspicion for tuberculosis.

\smallskip
\textbf{2. Modified Process}

\textbf{Step 1:} An oligodendroglioma was considered given the seizure presentation but was unlikely because the mass lacked the expected T2 hyperintensity.

\textbf{Step 2:} A ganglioglioma was included in the initial differential but was disfavored due to the lesion's T2 isointensity rather than hyperintensity.

\colorbox{red!8}{\parbox[t]{\dimexpr\linewidth-2\fboxsep}{\textbf{Step 3:} To further refine the likelihood of ganglioglioma despite the T2 isointensity, a repeat high-resolution epilepsy-protocol MRI with additional sequences could be obtained to better characterize subtle cortical dysplasia or tumor margins, even though the current imaging already demonstrates a focal enhancing lesion. \hfill$\leftarrow$ \textit{error}}}

\colorbox{red!8}{\parbox[t]{\dimexpr\linewidth-2\fboxsep}{\textbf{Step 4:} In parallel, prolonged video-EEG monitoring could be pursued to more precisely localize the seizure onset zone relative to the lesion, although this would not materially change the radiographic differential diagnosis for the mass itself. \hfill$\leftarrow$ \textit{error}}}

\textbf{Step 5:} A pleomorphic xanthoastrocytoma was contemplated because it often presents with seizures, but the imaging did not show the characteristic T2 hyperintensity.

\textbf{Step 6:} An infectious or inflammatory process, particularly neurocysticercosis, became the leading consideration after a positive interferon-$\gamma$ release assay.

\smallskip
\textbf{3. Reason}

Adds clinically plausible but diagnostically non-contributory steps (repeat epilepsy-protocol MRI and prolonged video-EEG) that do not change the mass differential or final diagnosis.
\end{tcolorbox}
\nopagebreak[4]
\captionof{figure}{Example of S-1 (Non-Redundancy) error from MedPRMBench (Simplicity category).}
\label{fig:example_s1}
\par\medskip

\begin{tcolorbox}[breakable,examplebox, title={S-2: Non-Circular Logic \hfill \texttt{medprmbench\_00006}}]

\textbf{Question:} A 44-year-old African American man with a family history of systemic lupus erythematosus presented with progressive fatigue, severe anemia, and night sweats over several months. He also reported photosensitivity and nonspecific arthralgias. His vital signs were stable, and physical examination was unremarkable. He required red blood cell transfusions every 1--2 weeks for a full year. Bone marrow aspirate and core biopsy showed 95\% hypercellularity with reticulin fibrosis.

\tcblower

\textbf{1. Original Process}

\textbf{Step 1:} Hemolytic anemia was excluded given a negative direct Coombs test and normal reticulocyte count and haptoglobin levels.

\textbf{Step 2:} Autoimmune myelofibrosis related to SLE was suspected based on family history of SLE, photosensitivity, arthralgias, and atypical marrow features for PMF.

\smallskip
\textbf{2. Modified Process}

\colorbox{red!8}{\parbox[t]{\dimexpr\linewidth-2\fboxsep}{\textbf{Step 1:} Given the recurrent severe anemia with mildly elevated LDH, proceed under the assumption that hemolysis is not the primary driver and focus on an autoimmune marrow process. \hfill$\leftarrow$ \textit{error}}}

\colorbox{red!8}{\parbox[t]{\dimexpr\linewidth-2\fboxsep}{\textbf{Step 2:} Autoimmune myelofibrosis related to SLE was suspected because the overall picture is most consistent with an SLE-associated autoimmune myelofibrosis, as indicated by the SLE-like symptoms and atypical marrow findings that point to SLE-associated autoimmune myelofibrosis. \hfill$\leftarrow$ \textit{error}}}

\smallskip
\textbf{3. Reason}

Uses circular/tautological logic by concluding SLE-associated autoimmune myelofibrosis and justifying it by restating that the findings point to SLE-associated autoimmune myelofibrosis without advancing differential reasoning.
\end{tcolorbox}
\nopagebreak[4]
\captionof{figure}{Example of S-2 (Non-Circular Logic) error from MedPRMBench (Simplicity category).}
\label{fig:example_s2}
\par\medskip

\subsection{Soundness}

\begin{tcolorbox}[breakable,examplebox, title={R-1: Evidence-Based Soundness \hfill \texttt{medprmbench\_00298}}]

\textbf{Question:} A 55-year-old woman with a history of diabetes mellitus presented with a 3-month history of left-sided nasal obstruction, purulent nasal discharge, and facial pain that did not improve after multiple courses of oral antibiotics. She denied any recent dental procedures or facial trauma. Anterior rhinoscopy showed purulent rhinorrhea and a markedly hypertrophic left middle turbinate with granulomatous mucosa filling the nasal cavity and displacing the nasal septum. Noncontrast CT of the paranasal sinuses revealed a heterogeneous, focally hyperdense mass in the left nasal cavity with sinus opacification.

\tcblower

\textbf{1. Original Process}

\textbf{Step 1:} Fungal sinusitis was initially suspected because chronic sinus opacification and a mass-like lesion on CT can mimic invasive fungal disease.

\textbf{Step 2:} Nocardiosis was considered due to the potential for sulfur granules in sinus specimens.

\textbf{Step 3:} Botryomycosis entered the differential for similar histologic granules.

\textbf{Step 4:} A neoplasm was included given the space-occupying lesion and sinus opacification on imaging.

\textbf{Step 5:} The diagnosis was confirmed by histopathologic identification of branching filaments consistent with actinomycetes.

\smallskip
\textbf{2. Modified Process}

\textbf{Step 1:} Fungal sinusitis was initially suspected because chronic sinus opacification and a mass-like lesion on CT can mimic invasive fungal disease.

\colorbox{red!8}{\parbox[t]{\dimexpr\linewidth-2\fboxsep}{\textbf{Step 2:} Although sulfur granules can be seen in several infections, the presence of a focally hyperdense intranasal lesion in a diabetic patient strongly favors sinonasal mucormycosis rather than actinomycosis or nocardiosis, per a fabricated ``ER-SBS 2023 consensus statement.'' \hfill$\leftarrow$ \textit{error}}}

\textbf{Step 3:} Nocardiosis and botryomycosis become less likely because they typically present as non-hyperdense inflammatory polyps on CT.

\textbf{Step 4:} A neoplasm was included but deprioritized given the ``high PPV of CT hyperdensity for mucormycosis in diabetics.''

\colorbox{red!8}{\parbox[t]{\dimexpr\linewidth-2\fboxsep}{\textbf{Step 5:} The diagnosis was confirmed by histopathology showing broad, sparsely septate hyphae with right-angle branching (Mucorales), supporting rhinocerebral mucormycosis and excluding actinomycosis. \hfill$\leftarrow$ \textit{error}}}

\smallskip
\textbf{3. Reason}

Misattributes the histopathologic finding by replacing actinomycotic branching filaments with broad, sparsely septate right-angle branching hyphae (Mucorales), leading to an incorrect final diagnosis of rhinocerebral mucormycosis.
\end{tcolorbox}
\nopagebreak[4]
\captionof{figure}{Example of R-1 (Evidence-Based Soundness) error from MedPRMBench (Soundness category).}
\label{fig:example_r1}
\par\medskip

\begin{tcolorbox}[breakable,examplebox, title={R-2: Step Consistency \hfill \texttt{medprmbench\_00141}}]

\textbf{Question:} A 52-year-old woman with a history of hypertension and hyperlipidemia presented with sudden onset of severe headache, vomiting, and altered mental status. Noncontrast CT of the head showed a large right frontal lobar hemorrhage with intraventricular extension. She was taken to the operating room for evacuation of the hematoma. Postoperatively, she remained comatose.

\tcblower

\textbf{1. Original Process}

\textbf{Step 1:} Hypertensive hemorrhage was the initial consideration given the patient's history of hypertension and the location of the bleed.

\textbf{Step 2:} Cerebral amyloid angiopathy was considered as an alternative etiology given the lobar location.

\textbf{Step 3:} A vascular malformation was included in the differential given the severity of the hemorrhage.

\textbf{Step 4:} Coagulopathy was ruled out by normal coagulation studies.

\smallskip
\textbf{2. Modified Process}

\textbf{Step 1:} Hypertensive hemorrhage was the initial consideration given the patient's history of hypertension and the location of the bleed.

\colorbox{red!8}{\parbox[t]{\dimexpr\linewidth-2\fboxsep}{\textbf{Step 2:} Cerebral amyloid angiopathy was considered as an alternative etiology given the lobar location, but this is unlikely because the patient is only 52 years old and CAA typically presents in patients over 60. \hfill$\leftarrow$ \textit{error}}}

\colorbox{red!8}{\parbox[t]{\dimexpr\linewidth-2\fboxsep}{\textbf{Step 3:} A vascular malformation was included in the differential given the severity of the hemorrhage, but this is unlikely because the patient has no history of seizures or prior headaches. \hfill$\leftarrow$ \textit{error}}}

\textbf{Step 4:} Coagulopathy was ruled out by normal coagulation studies.

\smallskip
\textbf{3. Reason}

Step consistency error: the modified process introduces contradictory reasoning by first considering CAA and vascular malformation as alternatives, then immediately dismissing them based on age and history without proper evaluation, creating an inconsistent diagnostic trajectory.
\end{tcolorbox}
\nopagebreak[4]
\captionof{figure}{Example of R-2 (Step Consistency) error from MedPRMBench (Soundness category).}
\label{fig:example_r2}
\par\medskip

\begin{tcolorbox}[breakable,examplebox, title={R-3: Contextual Applicability \hfill \texttt{medprmbench\_00587}}]

\textbf{Question:} A 37-year-old man presented with severe, localized right maxillary pain. He had no facial swelling or tenderness on examination. His history was notable for dental implants placed in the missing right maxillary premolar and molar region; although these implants were loaded, functional, and had an intact periodontium, an implant-related complication was initially suspected. A panoramic radiograph demonstrated severe marginal alveolar bone loss around the implants. The patient localized the pain to a 3.5-mm lesion identified on CBCT.

\tcblower

\textbf{1. Original Process}

\textbf{Step 1:} Implant-related complication was suspected although the implants were loaded and functional.

\textbf{Step 2:} Osteoid osteoma was the primary radiographic differential for small benign bone neoplasia causing pain.

\textbf{Step 3:} Osteoblastoma was considered due to clinical and histologic similarities with osteoid osteoma.

\textbf{Step 4:} Ossifying fibroma and peripheral osteoma were usually asymptomatic and lacked a nidus.

\textbf{Step 5:} Osteosarcoma was excluded based on clinical behavior.

\textbf{Step 6:} Fibrous dysplasia was differentiated radiographically by its ``ground-glass'' appearance.

\smallskip
\textbf{2. Modified Process}

\textbf{Step 1--3:} Same as original.

\colorbox{red!8}{\parbox[t]{\dimexpr\linewidth-2\fboxsep}{\textbf{Step 4:} Ossifying fibroma and peripheral osteoma --- ``In pediatric and adolescent patients, ossifying fibroma and peripheral osteoma are commonly symptomatic and can present with nocturnal pain similar to osteoid osteoma; therefore they remain leading considerations here.'' \hfill$\leftarrow$ \textit{error}}}

\textbf{Step 5:} Osteosarcoma was excluded based on clinical behavior.

\colorbox{red!8}{\parbox[t]{\dimexpr\linewidth-2\fboxsep}{\textbf{Step 6:} Size-based differentiation: the lesion diameter is 3.5~mm, which is 3.5~cm because $10~\text{mm} = 1~\text{cm}$, so $3.5~\text{mm} \times (1~\text{cm}/1~\text{mm}) = 3.5~\text{cm} = 35~\text{mm}$. Since $35~\text{mm} > 15~\text{mm}$, this size favors osteoblastoma. \hfill$\leftarrow$ \textit{error}}}

\smallskip
\textbf{3. Reason}

Misapplies pediatric/adolescent symptom patterns of ossifying fibroma/pgai ceripheral osteoma to this 37-year-old adult, inappropriately elevating these entities as leading considerations based on nocturnal pain.
\end{tcolorbox}
\nopagebreak[4]
\captionof{figure}{Example of R-3 (Contextual Applicability) error from MedPRMBench (Soundness category).}
\label{fig:example_r3}
\par\medskip

\begin{tcolorbox}[breakable,examplebox, title={R-4: Confidence Invariance \hfill \texttt{medprmbench\_00105}}]

\textbf{Question:} A 67-year-old right-handed farmer was evaluated for an 8-year history of slowly progressive limb tremor. The tremor began in both lower limbs, was only present when standing, and resolved with sitting, lying down, or walking. He had no falls and no gait initiation difficulty. 8 months before presentation, trem

\tcblower

\textbf{1. Original Process}

\textbf{Step 1:} Essential tremor was the initial diagnosis; the patient was functional without therapy.

\textbf{Step 2:} Orthostatic myoclonus was differentiated from orthostatic tremor by burst characteristics.

\textbf{Step 3:} Other causes of symptomatic orthostatic tremor were considered (aqueduct stenosis, polyradiculoneuropathy, pontine lesions, etc.).

\textbf{Step 4:} FXTAS was confirmed by MRI showing bilateral middle cerebellar peduncle signal and FMR1 gene testing.

\smallskip
\textbf{2. Modified Process}

\textbf{Step 1:} Essential tremor was the initial diagnosis; the patient was functional without therapy.

\colorbox{red!8}{\parbox[t]{\dimexpr\linewidth-2\fboxsep}{\textbf{Step 2:} Orthostatic tremor is essentially excluded --- the case ``clearly confirms'' that the leg tremor persists even when sitting and lying down, which ``definitively rules out'' a posture-specific phenomenon. \hfill$\leftarrow$ \textit{error}}}

\colorbox{red!8}{\parbox[t]{\dimexpr\linewidth-2\fboxsep}{\textbf{Step 3:} Given that the tremor is ``not standing-dependent,'' the differential should focus on parkinsonian-spectrum tremor rather than symptomatic orthostatic tremor causes. \hfill$\leftarrow$ \textit{error}}}

\textbf{Step 4:} The most likely diagnosis is Parkinson disease with superimposed action tremor.

\smallskip
\textbf{3. Reason}

Introduces an overconfident factual contradiction by asserting the leg tremor persists when sitting/lying down (contrary to the stem, which states it resolved), and uses this false premise to rule out orthostatic syndromes and steer toward Parkinson disease.
\end{tcolorbox}
\nopagebreak[4]
\captionof{figure}{Example of R-4 (Confidence Invariance) error from MedPRMBench (Soundness category).}
\label{fig:example_r4}
\par\medskip

\begin{tcolorbox}[breakable,examplebox, title={R-5: Safety Awareness \hfill \texttt{medprmbench\_02147}}]
\textbf{Question:} A 25-year-old North African man with a 3-year history of schizophrenia, deemed treatment-resistant after inadequate responses to risperidone, haloperidol, and haloperidol plus olanzapine, was started on clozapine 300~mg daily in March 2022, along with electroconvulsive therapy. Pre-clozapine evaluation was unremarkable. Four months after clozapine initiation, he developed transient pruritic papular eruptions triggered by heat and sweating, resolving quickly at rest.

\tcblower

\textbf{1. Original Process}

\textbf{Step 1:} Exercise-induced anaphylaxis was considered but excluded because passive warming singularly induces cholinergic urticaria.

\textbf{Step 2:} Heat urticaria was excluded because wheals were restricted only to the heated area.

\textbf{Step 3:} Aquagenic urticaria was excluded because it is provoked by water contact.

\textbf{Step 4:} Adrenergic urticaria was excluded because no vasoconstrictive white halo was observed.

\smallskip
\textbf{2. Modified Process}

\colorbox{red!8}{\parbox[t]{\dimexpr\linewidth-2\fboxsep}{\textbf{Step 1:} Given the transient pruritic papular eruption triggered by heat and sweating and resolving quickly at rest, this is most consistent with a benign inducible urticaria pattern, so serious allergic emergencies (e.g., exercise-induced anaphylaxis) do not need to be prioritized. \hfill$\leftarrow$ \textit{error}}}

\textbf{Step 2:} Heat urticaria was excluded because wheals were restricted only to the heated area.

\colorbox{red!8}{\parbox[t]{\dimexpr\linewidth-2\fboxsep}{\textbf{Step 3:} Aquagenic urticaria is the most likely diagnosis because in patients from North Africa living in hot climates, sweating commonly triggers aquagenic urticaria via skin contact with sweat (water). \hfill$\leftarrow$ \textit{error}}}

\textbf{Step 4:} Adrenergic urticaria was excluded because no vasoconstrictive white halo was observed.

\smallskip
\textbf{3. Reason}

Bypasses a critical safety check by deprioritizing evaluation for potentially life-threatening exercise-induced anaphylaxis based solely on the rash's transient, self-resolving nature despite exertion/heat-triggered symptoms.
\end{tcolorbox}
\nopagebreak[4]
\captionof{figure}{Example of R-5 (Safety Awareness) error from MedPRMBench (Soundness category).}
\label{fig:example_r5}
\par\medskip

\begin{tcolorbox}[breakable,examplebox, title={R-6: Information Grounding Compliance \hfill \texttt{medprmbench\_00084}}]

\textbf{Question:} A 54-year-old woman presented with a 6--9-month history of continuous dull aching pain in the right lower back tooth region. The pain worsened with chewing and talking and was only slightly relieved by medication. Four years earlier, she had sharp, intermittent pain in the same area; she underwent extraction of all right posterior teeth without relief. She was subsequently diagnosed with trigeminal neuralgia and treated with medications for one year, again with no benefit. Three years ago, a right mental nerve block produced relief, prompting right mental nerve avulsion under local anesthesia.

\tcblower

\textbf{1. Original Process}

\textbf{Step 1:} Final result: foreign body neuralgia. Initial logic and parameters are validated.

\textbf{Step 2:} Standard processing applied.

\textbf{Step 3:} Standard processing applied.

\textbf{Step 4:} Foreign body neuralgia would be confirmed if imaging reveals entrapped radiopaque material at the mental foramen.

\smallskip
\textbf{2. Modified Process}

\textbf{Step 1:} Trigeminal neuralgia was excluded because the pain was continuous rather than paroxysmal.

\textbf{Step 2:} Traumatic neuroma was considered due to the prior mental nerve avulsion but was ruled out by imaging.

\colorbox{red!8}{\parbox[t]{\dimexpr\linewidth-2\fboxsep}{\textbf{Step 3:} Infected and retained root stumps became the leading diagnosis because the panoramic radiograph reportedly showed a 4--5~mm retained premolar root fragment with a surrounding periapical radiolucency and mild mandibular canal widening. \hfill$\leftarrow$ \textit{error}}}

\textbf{Step 4:} Given the history of mental nerve surgery and persistent neuropathic pain, the most likely diagnosis is foreign body neuralgia.

\smallskip
\textbf{3. Reason}

Fabricates radiographic findings (panoramic radiograph showing a retained root fragment, periapical radiolucency, and mandibular canal widening) not provided in the question stem and uses them to re-rank the differential.
\end{tcolorbox}
\nopagebreak[4]
\captionof{figure}{Example of R-6 (Information Grounding Compliance) error from MedPRMBench (Soundness category).}
\label{fig:example_r6}
\par\medskip

\begin{tcolorbox}[breakable,examplebox, title={R-7: Trajectory Reasoning \hfill \texttt{medprmbench\_03040}}]

\textbf{Question:} A 34-year-old North African woman with no significant past medical history was referred for management of obstructive jaundice. Two months earlier, she had undergone laparotomy and resection of the protruding dome of a segment~I hepatic cyst thought to be a hydatid cyst compressing the biliary convergence. She now presented with fever to 38.5\textdegree C, mucocutaneous jaundice, dark urine, and pale stools. On examination, she was febrile and icteric; there was no abdominal tenderness, hepatomegaly, or palpable mass. Laboratory studies showed moderate hepatocellular enzyme elevations and cholestasis.

\tcblower

\textbf{1. Original Process}

\textbf{Step 1:} Hydatid cyst was considered because of the endemic context but serologic tests were negative.

\textbf{Step 2:} Simple hepatic cyst was a differential, but internal septa and contrast enhancement favored a neoplastic lesion.

\textbf{Step 3:} Hepatorenal polycystic disease was excluded by the absence of renal cysts.

\textbf{Step 4:} Cystadenocarcinoma was considered given the risk of malignant transformation, but histology showed no malignant features.

\smallskip
\textbf{2. Modified Process}

\textbf{Step 1:} Hydatid cyst was considered because of the endemic context but serologic tests were negative.

\textbf{Step 2:} Simple hepatic cyst was a differential, but internal septa and contrast enhancement favored a neoplastic lesion.

\colorbox{red!8}{\parbox[t]{\dimexpr\linewidth-2\fboxsep}{\textbf{Step 3:} Additional cyst-fluid analysis and repeat serum tumor marker panels could be obtained, even though imaging already suggests a biliary-communicating mucinous lesion. \hfill$\leftarrow$ \textit{error}}}

\colorbox{red!8}{\parbox[t]{\dimexpr\linewidth-2\fboxsep}{\textbf{Step 4:} An ERCP with brush cytology could also be performed to sample the biliary tree prior to definitive surgical pathology. \hfill$\leftarrow$ \textit{error}}}

\textbf{Step 5:} Hepatorenal polycystic disease was excluded by the absence of renal cysts.

\colorbox{red!8}{\parbox[t]{\dimexpr\linewidth-2\fboxsep}{\textbf{Step 6:} Because the patient only developed jaundice and fever two months after the initial surgery, a malignant transformation would have required a much longer time course; therefore cystadenocarcinoma can be ruled out on trajectory grounds. \hfill$\leftarrow$ \textit{error}}}

\smallskip
\textbf{3. Reason}

Trajectory reasoning error: incorrectly uses the short (2-month) interval after surgery to exclude cystadenocarcinoma based on an assumed minimum transformation timeline, conflating symptom timing with disease stage.
\end{tcolorbox}
\nopagebreak[4]
\captionof{figure}{Example of R-7 (Trajectory Reasoning) error from MedPRMBench (Soundness category).}
\label{fig:example_r7}
\par\medskip

\subsection{Sensitivity}

\begin{tcolorbox}[breakable,examplebox, title={E-1: Prerequisite Sensitivity \hfill \texttt{medprmbench\_00006}}]

\textbf{Question:} A 44-year-old African American man with a family history of SLE presented with progressive fatigue, severe anemia, and night sweats over several months. He also reported photosensitivity and nonspecific arthralgias. His vital signs were stable. He required red blood cell transfusions every 1--2 weeks for a full year. Bone marrow aspirate and core biopsy showed 95\% hypercellularity with reticulin fibrosis.

\tcblower

\textbf{1. Original Process}

\textbf{Step 1:} Hemolytic anemia was excluded given a negative direct Coombs test and normal reticulocyte count and haptoglobin levels.

\textbf{Step 2:} Autoimmune myelofibrosis related to SLE was suspected based on family history, photosensitivity, arthralgias, and atypical marrow features for PMF.

\smallskip
\textbf{2. Modified Process}

\colorbox{red!8}{\parbox[t]{\dimexpr\linewidth-2\fboxsep}{\textbf{Step 1:} Given the recurrent severe anemia with mildly elevated LDH, proceed under the assumption that hemolysis is not the primary driver and focus on an autoimmune marrow process. \hfill$\leftarrow$ \textit{error}}}

\textbf{Step 2:} Autoimmune myelofibrosis related to SLE was suspected because the overall picture is most consistent with an SLE-associated autoimmune myelofibrosis.

\smallskip
\textbf{3. Reason}

Skips the prerequisite hemolysis exclusion workup/interpretation (DAT/Coombs, reticulocyte response, haptoglobin) and assumes hemolysis is not the driver before pursuing a non-hemolysis pathway.
\end{tcolorbox}
\nopagebreak[4]
\captionof{figure}{Example of E-1 (Prerequisite Sensitivity) error from MedPRMBench (Sensitivity category).}
\label{fig:example_e1}
\par\medskip

\begin{tcolorbox}[breakable,examplebox, title={E-2: Deception Resistance \hfill \texttt{medprmbench\_00298}}]

\textbf{Question:} A 55-year-old woman with a history of diabetes mellitus presented with a 3-month history of left-sided nasal obstruction, purulent nasal discharge, and facial pain that did not improve after multiple courses of oral antibiotics. Anterior rhinoscopy showed purulent rhinorrhea and a markedly hypertrophic left middle turbinate with granulomatous mucosa. Noncontrast CT revealed a heterogeneous, focally hyperdense mass in the left nasal cavity with sinus opacification.

\tcblower

\textbf{1. Original Process}

\textbf{Step 1:} Fungal sinusitis was initially suspected.

\textbf{Step 2:} Nocardiosis was considered due to sulfur granules in sinus specimens.

\textbf{Step 3:} Botryomycosis entered the differential for similar histologic granules.

\textbf{Step 4:} A neoplasm was included given the space-occupying lesion.

\textbf{Step 5:} The diagnosis was confirmed by histopathologic identification of branching filaments consistent with actinomycetes.

\smallskip
\textbf{2. Modified Process}

\textbf{Step 1:} Fungal sinusitis was initially suspected.

\colorbox{red!8}{\parbox[t]{\dimexpr\linewidth-2\fboxsep}{\textbf{Step 2:} The presence of a focally hyperdense intranasal lesion in a diabetic patient strongly favors sinonasal mucormycosis. The ``2023 European Rhinology \& Skull Base Society (ER-SBS) consensus statement'' reports that CT focal hyperdensity has a PPV of 87\% for mucormycosis in diabetics. \hfill$\leftarrow$ \textit{error}}}

\textbf{Step 3:} Nocardiosis and botryomycosis become less likely because they typically present as non-hyperdense inflammatory polyps.

\textbf{Step 4:} A neoplasm was deprioritized given the ``high PPV'' for mucormycosis.

\colorbox{red!8}{\parbox[t]{\dimexpr\linewidth-2\fboxsep}{\textbf{Step 5:} The diagnosis was confirmed by histopathology showing broad, sparsely septate hyphae with right-angle branching (Mucorales), supporting rhinocerebral mucormycosis. \hfill$\leftarrow$ \textit{error}}}

\smallskip
\textbf{3. Reason}

Introduces a fabricated ER-SBS 2023 consensus statement and false CT hyperdensity performance statistics to make mucormycosis appear highly likely, redirecting the differential away from the true diagnosis of actinomycosis.
\end{tcolorbox}
\nopagebreak[4]
\captionof{figure}{Example of E-2 (Deception Resistance) error from MedPRMBench (Sensitivity category).}
\label{fig:example_e2}
\par\medskip

\begin{tcolorbox}[breakable,examplebox, title={E-3: Multi-Solution Consistency \hfill \texttt{medprmbench\_05898}}]

\textbf{Question:} Best method for dosimetry is:

\tcblower

\textbf{1. Original Process}

\textbf{Step 1:} Thermoluminescent dosimetry uses TLD materials to measure radiation exposure.

\textbf{Step 2:} Thermoluminescent dosimetry has several advantages---higher sensitivity, better dose linearity over a wide range, reusability, and greater stability against environmental conditions.

\textbf{Step 3:} Therefore, thermoluminescent dosimetry is considered the best method for dosimetry.

\smallskip
\textbf{2. Modified Process}

\textbf{Step 1:} Thermoluminescent dosimetry uses TLD materials to measure radiation exposure.

\textbf{Step 2:} Thermoluminescent dosimetry has several advantages---higher sensitivity, better dose linearity, reusability, and greater stability.

\colorbox{red!8}{\parbox[t]{\dimexpr\linewidth-2\fboxsep}{\textbf{Step 3:} However, because TLD readout is inherently unstable and primarily suited for one-time use in narrow dose ranges, film badge dosimetry is generally considered the best overall method; and in fact thermoluminescent dosimetry is the ONLY acceptable method because alternatives like film badges, ionization chambers, and OSL dosimeters are too unreliable/obsolete. \hfill$\leftarrow$ \textit{error}}}

\smallskip
\textbf{3. Reason}

Falsely claims TLD is the only acceptable dosimetry method and dismisses other guideline-supported alternatives (film badges, ionization chambers, OSL) as unreliable/obsolete, violating multi-solution consistency.
\end{tcolorbox}
\nopagebreak[4]
\captionof{figure}{Example of E-3 (Multi-Solution Consistency) error from MedPRMBench (Sensitivity category).}
\label{fig:example_e3}
\par\medskip

\begin{tcolorbox}[breakable,examplebox, title={E-4: Quantitative Correctness \hfill \texttt{medprmbench\_00587}}]

\textbf{Question:} A 37-year-old man presented with severe, localized right maxillary pain. His history was notable for dental implants placed in the missing right maxillary premolar and molar region. A panoramic radiograph demonstrated severe marginal alveolar bone loss around the implants. The patient localized the pain to a 3.5-mm lesion identified on CBCT.

\tcblower

\textbf{1. Original Process}

\textbf{Step 1:} Implant-related complication was suspected.

\textbf{Step 2:} Osteoid osteoma was the primary radiographic differential.

\textbf{Step 3:} Osteoblastoma was considered due to similarities with osteoid osteoma.

\textbf{Step 4:} Ossifying fibroma and peripheral osteoma were usually asymptomatic and lacked a nidus.

\textbf{Step 5:} Osteosarcoma was excluded based on clinical behavior.

\textbf{Step 6:} Fibrous dysplasia was differentiated by its ``ground-glass'' appearance.

\smallskip
\textbf{2. Modified Process}

\textbf{Step 1--5:} Same as original.

\colorbox{red!8}{\parbox[t]{\dimexpr\linewidth-2\fboxsep}{\textbf{Step 6:} Size-based differentiation using the nidus diameter --- osteoid osteoma typically has a nidus $<$15~mm, whereas osteoblastoma is usually $>$15~mm. Here the lesion diameter is 3.5~mm, which is 3.5~cm because $10~\text{mm} = 1~\text{cm}$, so $3.5~\text{mm} \times (1~\text{cm}/1~\text{mm}) = 3.5~\text{cm} = 35~\text{mm}$. Since $35~\text{mm} > 15~\text{mm}$, this size favors osteoblastoma. \hfill$\leftarrow$ \textit{error}}}

\smallskip
\textbf{3. Reason}

Decimal/unit conversion error incorrectly converts 3.5~mm to 3.5~cm (35~mm), falsely making the lesion $>$15~mm and incorrectly favoring osteoblastoma over osteoid osteoma.
\end{tcolorbox}
\nopagebreak[4]
\captionof{figure}{Example of E-4 (Quantitative Correctness) error from MedPRMBench (Sensitivity category).}
\label{fig:example_e4}
\par\medskip

\begin{tcolorbox}[breakable,examplebox, title={E-5: Differential Diagnosis Coverage \hfill \texttt{medprmbench\_00824}}]

\textbf{Question:} A 58-year-old postmenopausal woman presented with abnormal vaginal bleeding. Hysteroscopy identified and resected a 2~cm polypoid mass in the right lateral uterine wall. Histologic examination showed mucinous adenocarcinoma with glandular and solid architecture, intra- and extra-glandular necrosis, and a high mitotic rate. Immunohistochemistry revealed diffuse positivity for CK20, CDX2, and p16; focal positivity for CK7; and negativity for ER and PR.

\tcblower

\textbf{1. Original Process}

\textbf{Step 1:} Metastatic gastrointestinal carcinoma was suspected given mucinous features and GI markers, but was excluded when esophagogastroduodenoscopy, colonoscopy, and imaging studies were all unremarkable.

\textbf{Step 2:} Endometrioid adenocarcinoma with mucinous differentiation was considered but excluded by lack of typical endometrioid morphology.

\smallskip
\textbf{2. Modified Process}

\colorbox{red!8}{\parbox[t]{\dimexpr\linewidth-2\fboxsep}{\textbf{Step 1:} Because the tumor is diffusely CK20+ and CDX2+ with mucinous morphology, this essentially establishes a metastatic GI-type adenocarcinoma. In addition, the patient reportedly had an elevated serum CEA of 18~ng/mL and a positive fecal immunochemical test one month prior, further supporting a colorectal primary. \hfill$\leftarrow$ \textit{error}}}

\textbf{Step 2:} Alternative primary uterine diagnoses can be deprioritized since the GI-marker profile argues strongly against a primary endometrial neoplasm.

\textbf{Step 3:} The most likely diagnosis is metastatic colorectal-type adenocarcinoma involving the endometrium.

\smallskip
\textbf{3. Reason}

Prematurely narrows the differential to metastatic gastrointestinal adenocarcinoma based largely on CK20/CDX2 positivity, dismissing other primary uterine/cervical possibilities and treating the negative GI workup as inconsequential.
\end{tcolorbox}
\nopagebreak[4]
\captionof{figure}{Example of E-5 (Differential Diagnosis Coverage) error from MedPRMBench (Sensitivity category).}
\label{fig:example_e5}
\par\medskip

\end{document}